\documentclass[3p,times]{elsarticle}
\usepackage{float}
\usepackage{amssymb}
\usepackage{amsmath}
\usepackage{subcaption}
\usepackage{algorithm}
\usepackage{algpseudocode}
\usepackage{multirow}
\usepackage{url} 

\begin{document}

\begin{frontmatter}

\title{Autonomous Oil Spill Response Through Liquid Neural Trajectory Modeling and Coordinated Marine Robotics}


\author[label1]{Hadas C.Kuzmenko\corref{cor1}}
\ead{hadascohenk@gmail.com}

\author[label1]{David Ehevich}
\ead{david.ehevich@university.haifa.ac.il}

\author[label1]{Oren Gal}
\ead{oren.gal@university.haifa.ac.il}

\cortext[cor1]{Corresponding author. Tel.: +972-4-824-0493}

\address[label1]{Swarm \& AI Lab (SAIL), Hatter Department of Marine Technologies, Leon H. Charney School of Marine Sciences, University of Haifa, 199 Aba Khoushy Ave, Mount Carmel, Haifa 3498838, Israel}

%% Abstract
\begin{abstract}
Marine oil spills pose grave environmental and economic risks, threatening marine ecosystems, coastlines, and dependent industries. Predicting and managing oil spill trajectories is highly complex, due to the interplay of physical, chemical, and environmental factors such as wind, currents, and temperature, which makes timely and effective response challenging. Accurate real-time trajectory forecasting and coordinated mitigation are vital for minimizing the impact of these disasters.
This study introduces an integrated framework combining a multi-agent swarm robotics system built on the MOOS-IvP platform with Liquid Time-Constant Neural Networks (LTCNs). The proposed system fuses adaptive machine learning with autonomous marine robotics, enabling real-time prediction, dynamic tracking, and rapid response to evolving oil spills. By leveraging LTCNs—well-suited for modeling complex, time-dependent processes—the framework achieves real-time, high-accuracy forecasts of spill movement. Swarm intelligence enables decentralized, scalable, and resilient decision-making among robot agents, enhancing collective monitoring and containment efforts.
Our approach was validated using data from the Deepwater Horizon spill, where the LTC-RK4 model achieved 0.96 spatial accuracy, surpassing LSTM approaches by 23\%. The integration of advanced neural modeling with autonomous, coordinated robotics demonstrates substantial improvements in prediction precision, flexibility, and operational scalability. Ultimately, this research advances the state-of-the-art for sustainable, autonomous oil spill management and environmental protection by enhancing both trajectory prediction and response coordination.
\end{abstract}

\begin{highlights}
\item Novel LTC framework for real-time oil spill trajectory prediction
    \item 23\% improvement over LSTM in spatial accuracy (0.96 vs 0.74)
    \item MOOS-IvP integration enables autonomous multi-agent coordination
    \item Validated on Deepwater Horizon with complex spill geometries
    \item Scalable architecture supports dynamic fleet reconfiguration
\end{highlights}

%% Keywords

\end{frontmatter}



\section{Introduction}

Oil spills remain one of the most pressing environmental challenges of our time, with devastating consequences for marine ecosystems, biodiversity, and coastal livelihoods. The 2010 Deepwater Horizon incident exemplifies the catastrophic impact of such disasters, where more than 4 million barrels of oil were released into the Gulf of Mexico, creating long-lasting environmental and economic damage \cite{ainsworth2020}. Effective mitigation of these disasters requires precise prediction of oil spill trajectories to inform containment and recovery efforts, necessitating the development of sophisticated monitoring and prediction systems.

Traditional oil spill response methods often rely on satellite imagery and manual monitoring systems, which can be limited by weather conditions, spatial resolution, and temporal delays \cite{fingas2018}. Recent advances in remote sensing technology have significantly improved detection capabilities, with radar and optical sensors providing enhanced monitoring of oil spill extent and movement patterns. However, the dynamic nature of marine environments and the complexity of oil weathering processes demand more sophisticated, adaptive approaches that can operate in real-time.

Multi-agent systems (MAS) have emerged as a promising approach to address these challenges by deploying autonomous agents capable of distributed sensing, communication, and decision-making. These systems leverage the collective intelligence of multiple agents to provide scalable and efficient solutions for large-scale spill detection and tracking \cite{chamoso2017}. The integration of unmanned aerial vehicles (UAVs) and autonomous underwater vehicles (AUVs) within multi-agent frameworks has demonstrated significant potential for enhancing oil spill response capabilities through coordinated surveillance and data collection.

Recent developments in neural network architectures, particularly Liquid Time-Constant Networks (LTCNs), offer unprecedented capabilities for modeling temporal dynamics and adapting to changing environmental conditions \cite{hasani2020}. 
These networks, characterized by their ability to process continuous-time data and maintain stable long-term memory, are particularly well suited for predicting the complex, time-dependent behavior of oil spill trajectories. When combined with multi-agent systems, LTCNs provide a powerful framework for distributed, adaptive decision-making that can respond to the dynamic nature of marine oil spill scenarios.

This research proposes an integrated approach combining multi-agent systems with Liquid Time-Constant Networks to enhance oil spill trajectory prediction and response coordination. Through this framework, our goal is to advance the state-of-the-art in oil spill response technologies, ultimately contributing to the development of sustainable and effective marine environmental management systems that can operate autonomously in real-time.
\textit{Despite advances in swarm robotics and predictive modeling, no system currently combines real-time temporal learning (e.g., LTCNs) with decentralized agent coordination for spill tracking.}\\
The main contribution of this work lies in the development of \textbf{OilSpill}, an integrated framework that combines \textbf{Liquid Time-Constant Neural Networks (LTCNs)} with multi-agent autonomous systems to enable \textbf{real-time oil spill trajectory prediction and coordinated response}. This study introduces a novel \textbf{multi-scale temporal modeling approach} and a modular \textbf{neural-robotic interface via MOOS-IvP}, allowing predictive models to directly inform autonomous vehicle deployment under dynamically evolving marine conditions. By systematically comparing different numerical solvers within the LTC framework and validating performance on real-world Deepwater Horizon data, the proposed system demonstrates \textbf{superior spatial accuracy, temporal consistency, and operational reliability} over traditional LSTM-based models. This is the first study, to the best of our knowledge, to integrate continuous-time neural dynamics with decentralized robotic decision-making for environmental disaster mitigation.

\section{Theoretical Background}

\subsection{Oil Spill Detection and Monitoring}
The foundation of effective oil spill response lies in accurate and timely detection. Remote monitoring technologies have revolutionized oil spill monitoring capabilities, providing comprehensive coverage of large marine areas. 
\cite{fingas2018} provide an extensive review of oil spill remote sensing techniques, highlighting the evolution from basic visual detection to sophisticated satellite-based systems that incorporate synthetic aperture radar (SAR), optical sensors, and hyperspectral imaging. Modern remote sensing approaches leverage multiple data sources and spectral bands to distinguish oil from natural phenomena such as algae blooms or sediment plumes. \cite{alruzouq2020} demonstrate the integration of machine learning techniques with traditional remote detection and monitoring data, showing how feature extraction and classification algorithms can significantly improve detection accuracy and reduce false-positive rates. The Deepwater Horizon oil spill provided an unprecedented opportunity to test and refine oil spill modeling approaches.\cite{ainsworth2020} conducted a comprehensive review of ten years of modeling efforts related to this disaster, highlighting the advances made and the remaining challenges. Their analysis demonstrates the importance of integrating multiple modeling approaches and data sources to achieve accurate trajectory predictions. These advances in monitoring technology and data processing create the foundation upon which autonomous monitoring systems can be built.

\subsection{Multi-Agent Systems for Oil Spill Response}
Multi-agent systems (MAS) offer distributed sensing, scalability, and fault tolerance advantages for oil spill monitoring. These systems enable coordinated surveillance across large marine areas through autonomous agents that collaborate while maintaining individual decision-making capabilities.
Recent research has focused on swarm robotics and UAV applications. \cite{aznar2014} developed distributed drone systems using local sensory information without direct communication. \cite{chamoso2017} created a comprehensive UAV framework with intelligent task allocation algorithms that optimize coverage while minimizing energy consumption using the PANGEA platform.
\cite{corchado2008} introduced OSM (Oil Spill Monitoring), integrating oceanographic models with agent-based sensing for continuous spill tracking and trajectory prediction. \cite{mata2009}developed CROS (Contingency Response Multi-agent System for Oil Spills) for emergency response scenarios, combining real-time data acquisition with decision support systems for rapid resource deployment. \cite{karim2004} demonstrated agent-based mission management for UAVs, emphasizing the BDI (Belief-Desire-Intention) framework for autonomous decision-making in dynamic environments.

\subsection{Swarm Intelligence and Distributed Decision Making}
Swarm intelligence principles enable coordinated behavior in multi-agent oil spill monitoring through local agent interactions that achieve global objectives without centralized control, making them ideal for dynamic marine environments with unreliable communication.
\cite{badu2009} demonstrated swarm aquabots for oil spill detection, emphasizing scalable solutions for large-scale spill tracking with adaptive network reconfiguration. \cite{peshna2015} developed a particle swarm optimization (PSO) framework for dynamic decision making in response to marine oil spills, enabling agents to optimize positioning and detection strategies based on evolving conditions. Advanced coordination systems implement hierarchical finite state machines that enable UAVs to operate in multiple modes including station maintenance, route execution, and coordinated containment operations through distributed sensor networks and real-time communication protocols.
The MOOS-IvP framework provides a robust foundation for autonomous marine vehicle operations, offering a behavior-based architecture for complex mission execution \cite{benjamin2009}. It supports distributed decision making while maintaining centralized coordination capabilities through a publish-subscribe communication architecture \cite{benjamin2010}. The modular design of the framework enables real-time adaptation and fault tolerance, making it suitable for dynamic environmental response scenarios \cite{snyder2016}. MOOS-IvP has been extensively used in multi-vehicle coordination scenarios, with applications in oceanographic data gathering using market-based approaches \cite{jiang2010}. Its proven reliability in marine environments and standardized interfaces make it an ideal platform for integrating advanced techniques with multi-agent coordination systems \cite{benjamin2010}. The versatility of the framework is further demonstrated by its integration with other robotic architectures such as ROS, which combines the strengths of both systems for maritime autonomy applications \cite{snyder2016}.

\subsection{Liquid Time-Constant Networks}
Liquid Time-Constant Networks (LTCNs) offer powerful capabilities for capturing the temporal dynamics of oil spill behavior through continuous-time processing and adaptive time constants. \cite{hasani2020} introduced LTCNs as a novel approach to modeling time-dependent systems with improved stability and interpretability compared to traditional recurrent architectures. LTCNs incorporate differential equations directly into the network structure, enabling natural handling of irregularly sampled data and varying time constants, which are particularly valuable for oil spill modeling where environmental conditions and sampling frequencies vary significantly. \cite{hasani2021} established the theoretical foundations demonstrating LTCNs' capacity as universal approximates for continuous-time dynamical systems, ensuring that they can capture complex temporal dependencies in oil spill dispersion processes including changing weather conditions, ocean currents, and oil properties over time.\cite{chahine2023} demonstrated the robustness of liquid neural networks in out-of-distribution scenarios, showing their ability to maintain stable performance when encountering environmental conditions different from training data, crucial for oil spill response systems operating across diverse marine environments. \cite{lechner2020} enhanced the mathematical rigor of liquid neural network architectures through closed-form continuous-time neural networks, providing a solid theoretical framework for applying LTCNs to real-world dynamic systems, including oil spill trajectory prediction.

\subsection{Integration of Multi-Agent Systems and Neural Networks}
The convergence of multi-agent systems and advanced neural networks creates new opportunities for enhanced oil spill monitoring by combining distributed intelligence with pattern recognition and prediction capabilities. \cite{marino2024} demonstrated multi-UAV distributed trajectory generation using end-to-end learning approaches, showcasing cooperative behavior for real-time navigation and decision-making challenges applicable to oil spill trajectory prediction, where multiple agents must coordinate to track evolving spill boundaries. The integration of LTCNs with swarm-based approaches provides a framework for adaptive and decentralized decision-making. LTCNs enable real-time prediction of oil spill trajectories based on current environmental conditions and historical patterns. When combined with swarm-based sensing networks, they create systems capable of both distributed data collection and sophisticated temporal modeling. This integrated approach addresses key limitations by combining the spatial coverage advantages of multi-agent systems with the temporal modeling capabilities of advanced neural networks, enabling adaptive response to changing conditions while maintaining scalability and fault tolerance.\cite{crosetto2002} demonstrated the effectiveness of the RAPSODI (Remote-sensing Anti-Pollution System for geOgraphic Data Integration) project in combining multiple sensor technologies including SAR, infrared/ultraviolet scanners, and conventional radar systems. The project emphasized that no single sensor provides complete oil spill monitoring capabilities, which requires integrated approaches with complementary sensing technologies. \cite{robbe2006} advanced multi-sensor data processing through the MEDUSA (Multispectral Environmental Data Unit for Surveillance Applications) system, integrating various sensors including side-looking airborne radar, infrared/ultraviolet scanners, microwave radiometers, and laser fluorosensors with real-time data acquisition and processing capabilities. \cite{fingas2018} provided an extensive review of oil spill remote sensing technologies, highlighting the evolution from single-sensor approaches to integrated multisensor platforms, highlighting the complementary nature of different sensing modalities.

\subsection{Oil Spill Modeling: Challenges and Solutions}
The Deepwater Horizon experience revealed critical limitations in traditional static modeling approaches, highlighting the need for adaptive systems that can incorporate real-time data and adjust predictions as conditions evolve. Oil spills undergo complex weathering processes that include spreading, evaporation, emulsification, and biodegradation, each with distinct temporal characteristics that conventional mathematical models struggle to capture accurately. \cite{peshna2015} and \cite{pashna2014} highlighted the importance of dynamic decision making in marine oil spill responses through simulation-based multi-agent particle swarm optimization (PSO) frameworks, demonstrating how optimization algorithms can be integrated with multi-agent systems to adapt to changing environmental conditions in real time.
Multi-agent systems offer inherent scalability advantages, allowing additional agents to be deployed as spill size increases, expanding monitoring coverage and response capabilities. The distributed intelligence approach enables autonomous decision making based on local information while contributing to global system objectives. This distributed decision-making is enhanced by integrating Liquid Time-Constant Networks (LTCNs), which can process local temporal patterns while maintaining awareness of global system dynamics.
Despite significant advances in multi-agent systems and neural network architectures, their integration for oil spill response remains largely unexplored. Current systems often rely on simplified models that do not capture the full complexity of oil spill dynamics or provide the real-time adaptability required for an effective response. Although the RAPSODI project and systems such as MEDUSA have demonstrated the value of multisensor integration and real-time data processing, they lack the adaptive learning capabilities that modern neural networks provide. Recent UAV-based multi-agent systems for oil spill detection show promise but rely primarily on traditional computer vision and rule-based decision making.
\cite{deepwater_horizon_experience} highlighted limitations of traditional static modeling approaches and the need for adaptive systems that learn from real-time observations. This research addresses these gaps by developing an integrated framework combining the distributed intelligence of multi-agent systems with the temporal modeling capabilities of liquid-time constant networks, promising more accurate trajectory prediction, better coordination between response agents, and improved adaptability to changing environmental conditions.

\section{Methodology}

\subsection{Study Case: Deepwater Horizon Shapefile Dataset}

The methodology used in this research centers on the analysis of the Deepwater Horizon oil spill incident, using comprehensive satellite observation data and advanced neural network modeling techniques. This section presents a systematic approach to evaluate liquid-time-constant network performance against traditional methods using real-world spill data.

\subsubsection{Study Case: Deepwater Horizon Shapefile Dataset}

The primary case study focuses on the Deepwater Horizon oil spill incident on April 24, 2010-May 2, 2010, which is one of the most significant marine environmental disasters in recent history. This incident provides an ideal testbed for the evaluation of the oil spill prediction model due to the extensive satellite monitoring data available and the complex environmental conditions present during the evolution of the spill.

\begin{figure}[htbp]
  \centering
  \begin{subfigure}[b]{0.48\textwidth}
    \centering
    \includegraphics[width=\linewidth]{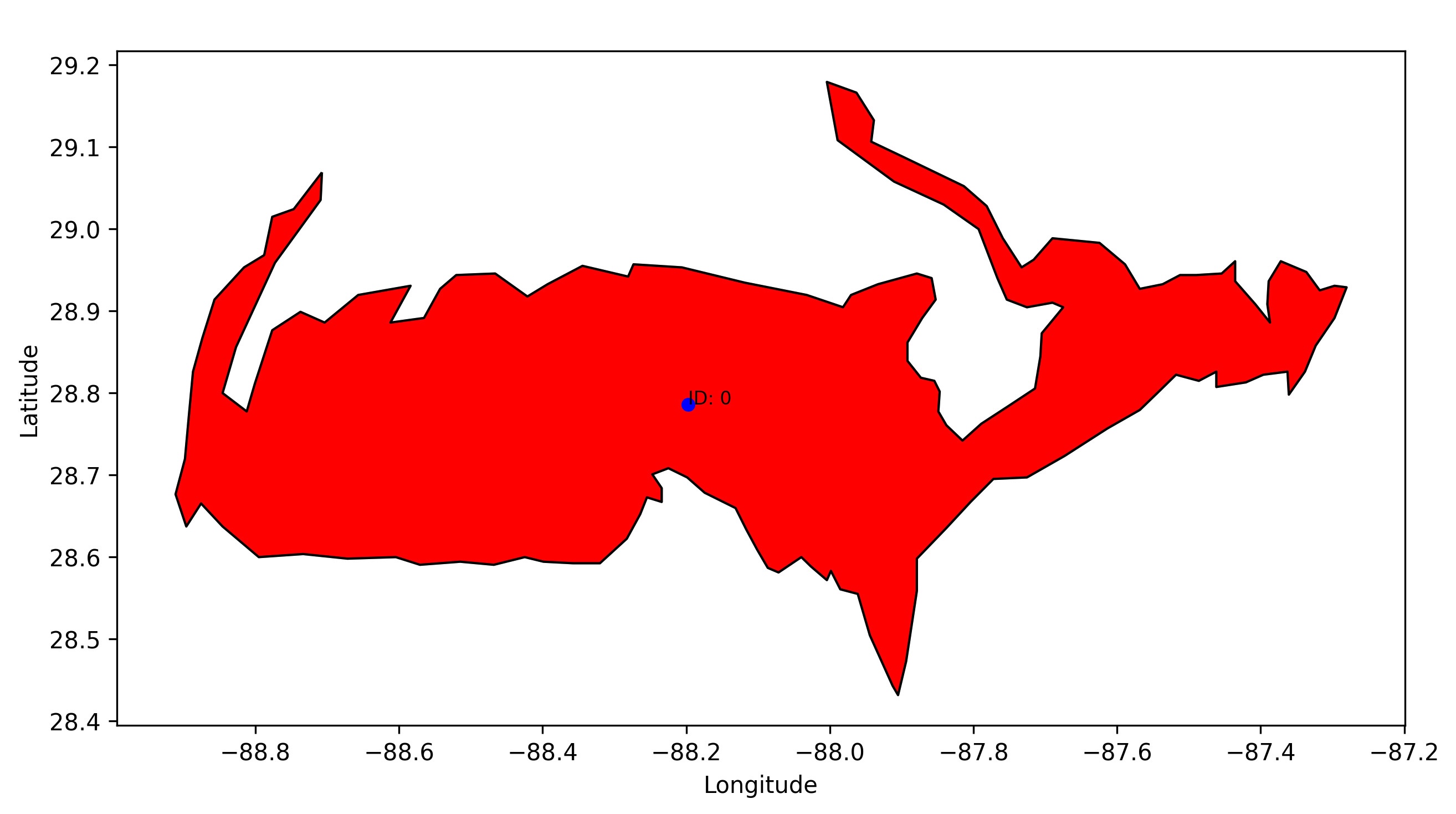}
    \caption{OilSpill system Spill Boundary Analysis showing detected spill boundaries with complex geometric features and multi-lobe configuration characteristic of the Deepwater Horizon incident}
    \label{fig:OilSpill system Spill Boundary Analysis}
  \end{subfigure}
  \hfill
  \begin{subfigure}[b]{0.48\textwidth}
    \centering
    \includegraphics[width=\linewidth]{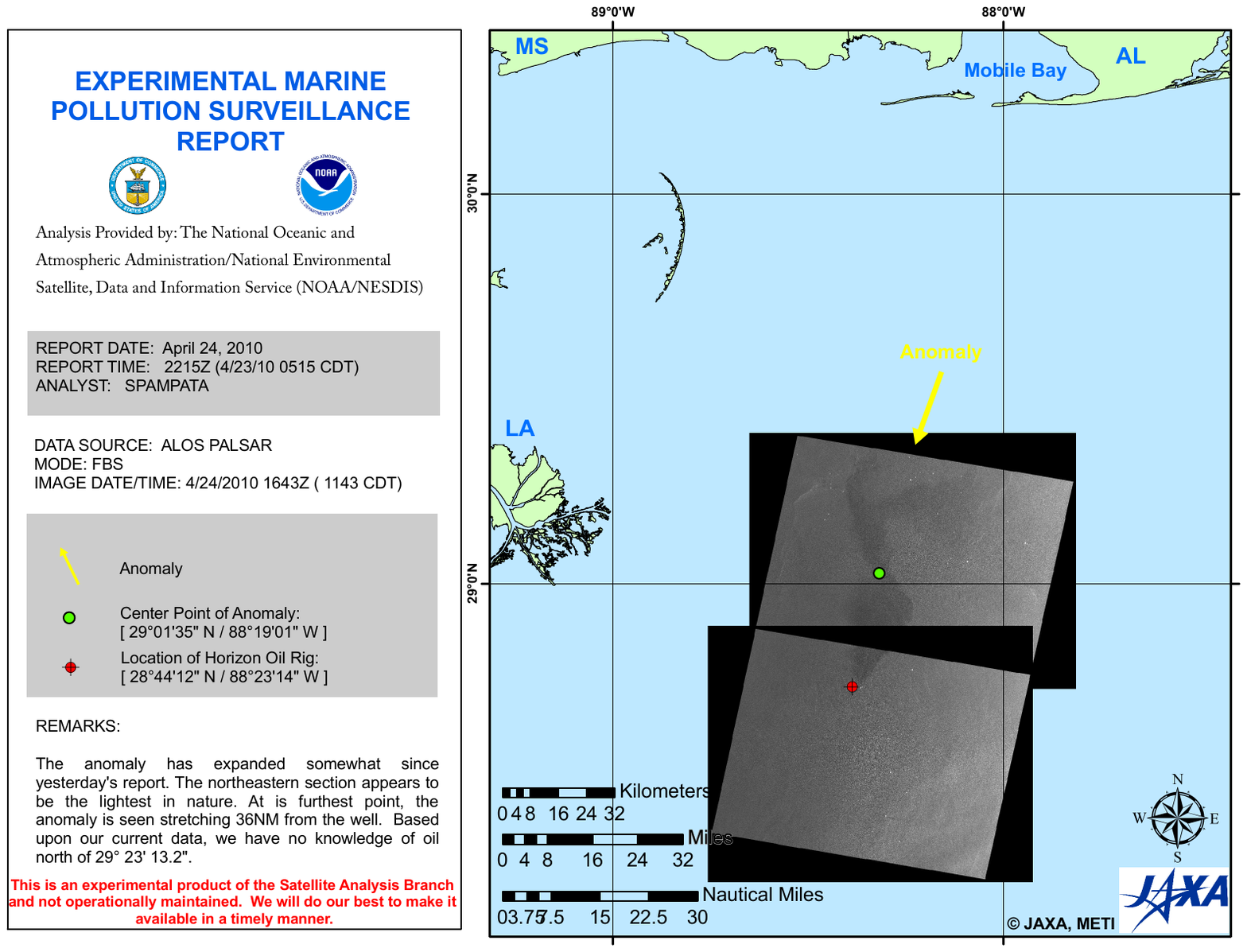}
    \caption{NOAA/NESDIS Satellite Analysis Report from April 24, 2010, showing ALOS PALSAR radar imagery and official spill boundary delineation used as ground truth for model validation}
    \label{fig:noaa_satellite_report}
  \end{subfigure}
  \caption{Deepwater Horizon case study data from April 24, 2010: (a) OilSpill boundary analysis, (b) NOAA/NESDIS satellite surveillance report.}
  \label{fig:deepwater_horizon_analysis}
\end{figure}

The Deepwater Horizon dataset, illustrated in Figure~\ref{fig:deepwater_horizon_analysis}, provides comprehensive information about the evolution of the spill under real operational conditions. The incident occurred at coordinates 28°44'12" N / 88°23'14" W, with the spill center documented at 29°01'35" N / 88°19'01" W on April 24, 2010. The NOAA/NESDIS experimental marine pollution surveillance report indicates that the anomaly had expanded significantly, with the northeastern section appearing lightest in nature and extending approximately 36 nautical miles from the wellhead.

\subsubsection{Data Source Characteristics}
The satellite data used in this study originate from multiple sources to ensure comprehensive coverage and validation capabilities:

\paragraph{ALOS PALSAR Radar Data}
The primary data source consists of advanced land observation satellite (ALOS) Phased Array type L-band Synthetic Aperture Radar (PALSAR) images acquired on April 24, 2010, at 1643Z (1143 CDT). This radar-based system provides an all-weather monitoring capability essential for continuous spill tracking regardless of cloud cover or lighting conditions.

\paragraph{NOAA/NESDIS Analysis}
Official analysis provided by the National Oceanic and Atmospheric Administration/National Environmental Satellite, Data and Information Service offers professionally validated spill boundary delineations. These reports serve as ground-truth data for model validation and provide expert interpretation of satellite imagery under operational conditions.

\paragraph{Shapefile Geometric Data}
The dataset includes precise geometric representations of the spill boundaries in shapefile format, allowing quantitative analysis of the evolution of the area, the movement of centroids and the complexity of the boundaries over time.

\subsubsection{Methodological Framework}
The experimental methodology employs a comparative analysis framework that evaluates multiple prediction approaches under identical initial conditions:

\paragraph{Model Initialization}
All models (LTC-RK4, LTC-Euler, LTC-Explicit, and LSTM) are initialized using identical spill parameters derived from the April 24, 2010- May 2, 2010 observations. This ensures a fair comparison by eliminating initialization bias and focusing evaluation on predictive capability differences.

\paragraph{Environmental Condition Integration}
The methodology incorporates real environmental conditions present during the Deepwater Horizon incident, including wind patterns, ocean currents, and weather conditions documented in NOAA reports. This integration ensures that model performance evaluation reflects real-world operational challenges.

\paragraph{Multi-Scale Analysis}
The evaluation framework addresses both local-scale boundary evolution and large-scale trajectory prediction, providing a comprehensive assessment of model capabilities across different spatial and temporal scales relevant to emergency response operations.

\subsubsection{Training and Testing Data Configuration}

The data set configuration follows a systematic approach to ensure robust model training and unbiased performance evaluation.

\textbf{Training Dataset:} The LTC networks were trained using shapefile data from multiple dates during the Deepwater Horizon incident to capture various patterns of evolution of the spill and environmental conditions. The training data set comprises:
\begin{itemize}
    \item April 25, 2010 - Post-initial expansion phase
    \item April 29, 2010 - Mid-stage spill development  
    \item May 01, 2010 - Complex multi-lobe geometry phase
    \item May 02, 2010 - Advanced weathering and dispersion stage
    \item February 4, 2010 - Pre-incident baseline conditions for environmental modeling
\end{itemize}

This temporal distribution ensures that the models learn from various stages of the evolution of the spill, the environmental forcing conditions, and the geometric complexity patterns characteristic of major marine oil spill incidents.

\textbf{Testing and Analysis Dataset:} Model performance evaluation and results analysis utilize independent data not included in the training process:
\begin{itemize}
    \item April 24, 2010 - Primary evaluation case (early expansion phase)
    \item April 26-28, 2010 - Extended temporal sequence analysis (progressive evolution tracking)
\end{itemize}

This separation between training and testing datasets prevents overfitting and ensures that performance metrics reflect the models' true generalization capability for operational deployment scenarios.

\subsubsection{Validation Protocol}

The validation protocol rigorously evaluates model performance by comparing predictions against official NOAA/NESDIS spill boundaries using multiple complementary metrics: spatial overlap, centroid displacement, area accuracy, and boundary fidelity. The methodology assesses both short-term accuracy and long-term trajectory stability in multiple time steps, while incorporating uncertainty analysis to gauge prediction confidence under varying environmental conditions. This comprehensive approach enables a thorough evaluation of the LTC network's oil spill prediction capabilities and informs optimal deployment strategies for emergency response. The Deepwater Horizon case study provides a demanding testbed with complex environmental conditions and irregular spill geometry characteristic of major marine incidents.

\subsection{System Overview}
The proposed oil spill trajectory prediction system (OilSpill) integrates Liquid Time-Constant Networks (LTCNs) with multi-agent architectures to create an adaptive, real-time prediction, and response framework. The system architecture consists of four primary components: (1) a distributed data acquisition layer utilizing shapefile-based oil spill observations, (2) an enhanced feature extraction pipeline that processes geometric, temporal and environmental characteristics, (3) a sophisticated LTC-based prediction engine capable of modeling complex spill dynamics across multiple temporal scales, and (4) multi-robot system, which autonomously navigates and tracks oil spills on sea surfaces, allowing it to quickly respond to changing environmental conditions.
The architectural design emphasizes modularity and scalability, allowing seamless integration with existing marine monitoring systems such as MOOS-IvP. The system processes geospatial data in real-time, extracting comprehensive feature sets that capture both instantaneous spill characteristics and temporal evolution patterns. The prediction engine generates multi-horizon forecasts, providing trajectory estimates at 3, 7, 11, and 15-hour intervals to support both immediate response operations and strategic planning activities.
\subsubsection{Data Acquisition and Processing Pipeline}
The data acquisition layer processes oil spill observations encoded as ESRI shapefiles, which contain precise polygon representations of spill boundaries at discrete time intervals. Each shapefile undergoes comprehensive reprocessing to extract a 25-dimensional feature vector encompassing geometric properties (area, perimeter, compactness, convexity), spatial characteristics (centroid coordinates, aspect ratio, orientation), temporal features (time since spill initiation, cyclical time encoding) and environmental conditions (wind vectors, current patterns, temperature).
The preprocessing pipeline implements robust coordinate system transformations to ensure geographic accuracy, converting all spatial measurements to the WGS84 coordinate system (EPSG:4326) and applying latitude-dependent corrections for area calculations. The system handles multi-polygon geometries by selecting the largest component, ensuring consistency in feature extraction across varied spill configurations.
\subsubsection{Multi-Scale Temporal Modeling}
The system implements a novel multi-scale temporal modeling approach that stratifies training data based on prediction horizons. Short-term sequences (0-48 hours) utilize interpolated hourly observations to capture rapid spill evolution dynamics, while medium-term sequences (1-7 days) focus on daily progression patterns influenced by persistent environmental forcing. This stratification enables the LTC network to learn scale-appropriate dynamics, improving prediction accuracy across different temporal horizons.
Environmental forcing is incorporated through historically accurate atmospheric and oceanographic conditions specific to the Gulf of Mexico region during the Deepwater Horizon incident \cite{judt2016}. Wind patterns, ocean currents, and temperature variations are encoded as additional features, allowing the model to account for the complex interplay between spill behavior and environmental drivers.

\subsection{OilSpill model Architecture}
The core prediction engine utilizes an OilSpill Model based on Liquid Time-Constant Networks, specifically designed to handle the continuous-time dynamics inherent in oil spill evolution. The LTC architecture addresses key limitations of traditional recurrent networks by incorporating differential equations directly into the network structure, enabling natural handling of irregularly sampled data and varying temporal scales characteristic of environmental monitoring systems.

\subsubsection{LTC Cell Design and Solver Implementation}
The LTC cell implementation follows the mathematical formulation introduced by \cite{hasani2020}, where the state evolution is governed by:
\begin{equation}
\tau dx/dt = -x + f(W_x \cdot u + W_r \cdot h + b)
\end{equation}
where $\tau$ represents learnable time constants, $x$ denotes the cell state, $u$ is the input, $h$ represents the recurrent state, and $f$ is the activation function. The system implements three solver variants.

This shows that the effective time constant is input-dependent, and given by:
\begin{equation}
\tau_{\text{sys}} = \tau / (1 + \tau f(x(t), I(t), t, \theta))
\end{equation}

\paragraph{LTC-EX (Explicit Adaptive):}
To compute updates, a Fused Euler ODE solver is introduced to accommodate different computational requirements and accuracy:
\begin{equation}
x(t + \Delta t) = [x(t) + \Delta t \cdot f(x(t), I(t), t, \theta) \cdot A] / [1 + \Delta t \cdot (1/\tau + f(x(t), I(t), t, \theta))]
\end{equation}

The LTC-EX incorporates adaptive time stepping based on state magnitude, automatically adjusting integration steps to maintain numerical stability while optimizing computational efficiency. The adaptive mechanism scales the time step according to:
\begin{equation}
\text{adaptive\_dt} = \text{dt} / (1 + \|\text{state}\|)
\end{equation}

This approach balances computational speed with numerical accuracy, making it suitable for real-time operational deployments where rapid prediction updates are essential.

Our LTC framework also supports two other numerical solvers, each based on distinct integration schemes:

\paragraph{LTC-RK (Runge-Kutta 4th Order):}
Provides high-accuracy integration suitable for complex oil spill dynamics requiring precise temporal modeling. The RK4 solver implements:
\begin{align}
k_1 &= f(t_n, \mathbf{y}_n) \\
k_2 &= f(t_n + h/2, \mathbf{y}_n + h k_1/2) \\
k_3 &= f(t_n + h/2, \mathbf{y}_n + h k_2/2) \\
k_4 &= f(t_n + h, \mathbf{y}_n + h k_3) \\
\mathbf{y}_{n+1} &= \mathbf{y}_n + h/6(k_1 + 2k_2 + 2k_3 + k_4)
\end{align}
This solver excels in scenarios requiring high temporal resolution and accurate modeling of nonlinear spill dynamics, particularly during critical early phases of spill evolution.

\paragraph{LTC-Euler (Baseline):} Implements simple forward Euler integration for rapid prototyping and baseline comparisons. While computationally efficient, this solver requires careful time step selection to maintain stability, making it appropriate for initial model development and testing phases.

\paragraph{LSTM Implementation:} The OilSpill model can also model the temporal dynamics of oil spill evolution using a deep LSTM network architecture specifically designed for multi-step prediction of spill characteristics.
\begin{align}
\mathbf{f}_t &= \sigma(\mathbf{W}_f \cdot [\mathbf{h}_{t-1}, \mathbf{x}_t] + \mathbf{b}_f) \\
\mathbf{i}_t &= \sigma(\mathbf{W}_i \cdot [\mathbf{h}_{t-1}, \mathbf{x}_t] + \mathbf{b}_i) \\
\tilde{\mathbf{C}}_t &= \tanh(\mathbf{W}_C \cdot [\mathbf{h}_{t-1}, \mathbf{x}_t] + \mathbf{b}_C) \\
\mathbf{C}_t &= \mathbf{f}_t \ast \mathbf{C}_{t-1} + \mathbf{i}_t \ast \tilde{\mathbf{C}}_t \\
\mathbf{o}_t &= \sigma(\mathbf{W}_o \cdot [\mathbf{h}_{t-1}, \mathbf{x}_t] + \mathbf{b}_o) \\
\mathbf{h}_t &= \mathbf{o}_t \ast \tanh(\mathbf{C}_t)
\end{align}
where $\mathbf{f}_t$, $\mathbf{i}_t$, and $\mathbf{o}_t$ represent forget, input, and output gates respectively, $\sigma$ denotes the sigmoid activation and $\ast$ indicates element-wise multiplication. The first LSTM layer with 128 hidden units captures the primary temporal dependencies, while the second layer (64 units) refines the representations. Dropout regularization ($p=0.1$) prevents overfitting during the training process of 150 epochs. The final dense layer maps the learned temporal features back to the 28-dimensional output space, enabling multi-step ahead prediction of spill geometry, position, and environmental interactions.

\subsubsection{Enhanced Architecture Components}
The OilSpill Model incorporates several architectural innovations beyond the core LTC layers:

\paragraph{Multi-Head Attention Mechanism:} The attention mechanism operates on different time scales, allowing the model to focus on relevant historical patterns when making predictions. The attention mechanism computes:
\begin{equation}
\text{Attention}(Q, K, V) = \text{softmax}(QK^T / \sqrt{d_k})V
\end{equation}
where query, key, and value matrices are derived from the LTC layer output, allowing the model to selectively weight historical observations based on their relevance to current predictions.

\paragraph{Feature Processing Pipeline:} Input features undergo projection to a 128-dimensional hidden space through a fully connected layer with ReLU activation, followed by layer normalization to ensure stable training dynamics. This processing enables the model to learn complex feature interactions while maintaining numerical stability.

\paragraph{Uncertainty Quantification:} The model implements an additional output branch that estimates prediction uncertainty through a softplus activation layer, providing confidence bounds for trajectory predictions. This capability is crucial for operational decision making, allowing response teams to assess prediction reliability.

\subsubsection{Architectural Differences}
The OilSpill framework is flexible and allows switching between LTC and LSTM models.
\begin{table}[ht]
\centering
\caption{Comparison of LTC and LSTM Model Characteristics}
\label{tab:model_comparison}
  \resizebox{1.0\textwidth}{!}{
\begin{tabular}{|l|l|l|}
\hline
\textbf{Aspect} & \textbf{LTC Networks} & \textbf{LSTM Networks} \\
\hline
Temporal Modeling & Continuous-time dynamics & Discrete-time sequences \\
Physical Interpretation & ODE-based, physically meaningful & Black-box recurrent states \\
Numerical Integration & Multiple solver options & Fixed update rules \\
Stability & Mathematically guaranteed & Dependent on training \\
Computational Complexity & Configurable (solver-dependent) & Fixed architecture \\
\hline
\end{tabular}
}
\end{table}

\subsubsection{Training Strategy and Loss Function}
The training procedure implements a sophisticated loss function designed specifically for temporal oil spill prediction. The total loss function is defined as:
\begin{equation}
L_{\text{total}} = L_{\text{mse}} + \alpha \cdot L_{\text{smoothness}} + \beta \cdot L_{\text{area}}
\end{equation}
where $L_{\text{mse}}$ represents the standard mean squared error, $L_{\text{smoothness}}$ enforces temporal consistency through:
\begin{equation}
L_{\text{smoothness}} = \| y_{\text{pred}}[t+1] - y_{\text{pred}}[t] \|^2
\end{equation}
and $L_{\text{area}}$ provides additional weighting for area-related features, given their critical importance in spill response planning.

The smoothness regularization parameter $\alpha$ varies by solver type: $0.05$ for RK4, $0.1$ for explicit and $0.2$ for Euler, reflecting the different stability characteristics of each integration method.

Training is performed using the AdamW optimizer with solver-specific learning rates: $0.0003$ for RK4, $0.0005$ for explicit and $0.001$ for Euler. Early stopping is applied with solver-adjusted patience values to accommodate the convergence behavior of each method. This adaptive training strategy ensures optimal performance across integration schemes while preventing overfitting to solver-specific temporal patterns.

\subsubsection{Data Normalization and Preprocessing}
Feature normalization follows a robust z-score standardization approach computed from training data statistics:
\begin{equation}
x_{\text{normalized}} = (x - \mu) / \sigma
\end{equation}
The system implements outlier clipping at $\pm 3\sigma$ to prevent extreme values from destabilizing the training while preserving the natural variability inherent in oil spill observations. Special handling addresses zero-variance features by substituting unit variance, ensuring numerical stability during normalization operations.

The construction of temporal sequences creates 16 timestep windows with overlapping segments to maximize data utilization while maintaining temporal coherence. For datasets with insufficient temporal resolution, the system implements intelligent padding through trend extrapolation, preserving realistic spill evolution patterns while meeting model input requirements.

\subsection{MOOS-IvP Integration Framework}

The integration of liquid-time-constant networks with the mission-oriented operating suite-interval programming (MOOS-IvP) autonomous systems represents a critical advancement in real-time oil spill response coordination. This section details the architectural implementation and operational workflow that enables seamless communication between neural network predictions and autonomous vehicle fleet coordination.

\begin{figure}[htbp]
  \centering
  \includegraphics[width=0.9\textwidth]{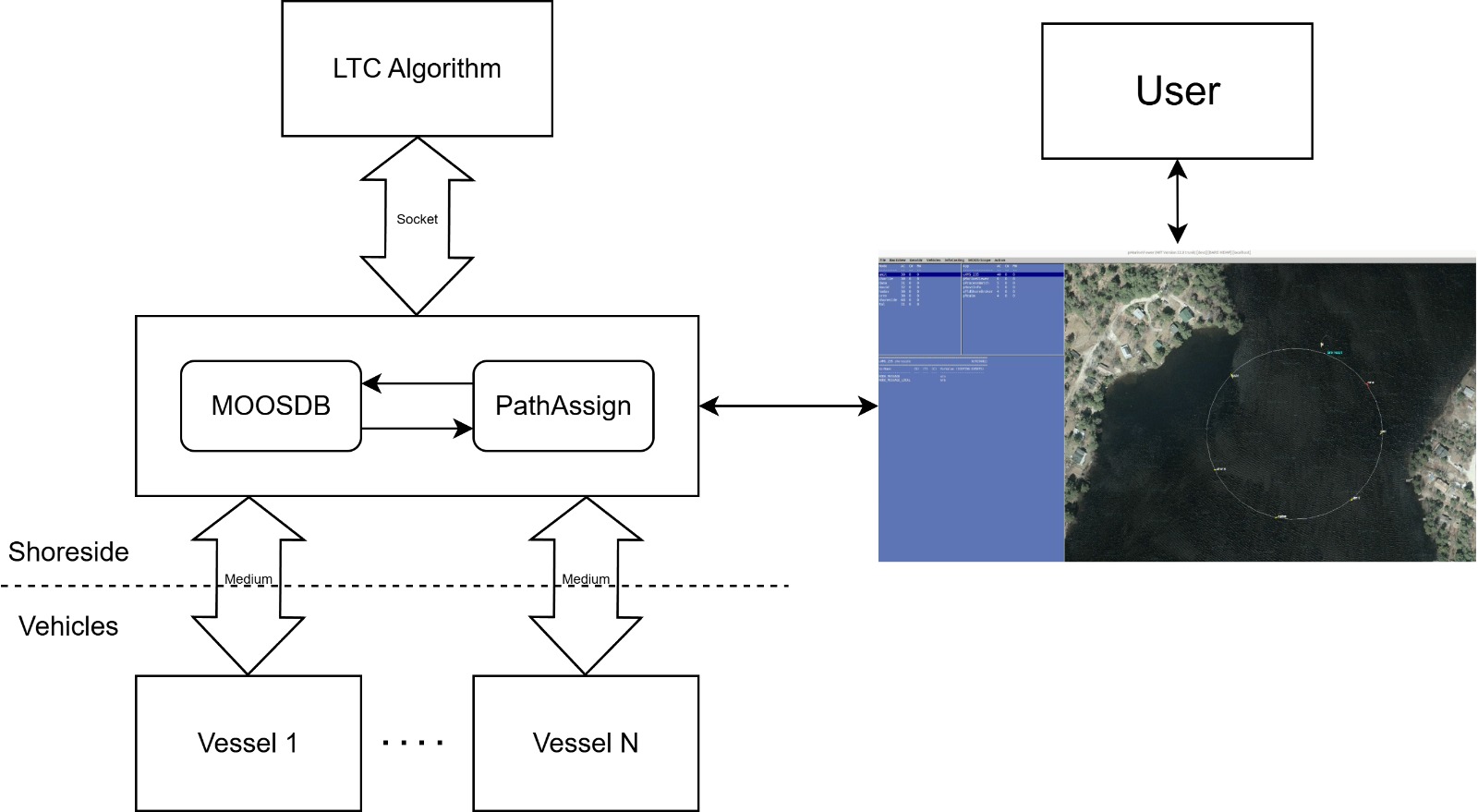}
  \caption{MOOS-IvP integration architecture showing LTC predictions to autonomous vehicle deployment workflow.}
  \label{fig:moos_architecture}
\end{figure}

\subsubsection{System Architecture}

The integrated system architecture, illustrated in Figure~\ref{fig:moos_architecture}, demonstrates a hierarchical communication structure that enables real-time coordination between predictive modeling and autonomous response deployment. The framework consists of several key components:

\paragraph{Module:} The neural network processing unit executes the liquid time-constant network computations to predict the oil spill trajectory. This module processes environmental sensor data, historical spill information, and real-time observations to generate boundary predictions and trajectory forecasts. The MOOS framework communicates with the MOOS framework through socket-based connections, enabling low-latency data transmission, critical for emergency response applications.

\paragraph{MOOS Database (MOOSDB):} The central communication hub manages all information sharing throughout the system between distributed components. MOOSDB maintains real-time databases of vehicle positions, environmental conditions, spill boundaries, and mission parameters. This centralized approach ensures data consistency across all system components while providing the flexibility needed for dynamic mission adaptation.

\paragraph{Path Assignment Module (PathAssign):} This critical component translates LTC predictions into actionable waypoint sequences for autonomous vehicle deployment. The PathAssign module optimizes vehicle trajectories based on predicted spill boundaries, environmental constraints, and operational objectives. The module employs advanced optimization algorithms to ensure efficient coverage while minimizing inter-vehicle conflicts and operational costs.

\paragraph{Autonomous Vehicle Fleet:} The distributed fleet of autonomous vessels (Vessel 1 through Vessel N) executes assigned missions based on LTC predictions and MOOS coordination. Each vehicle maintains bidirectional communication with the central system through medium-based protocols, allowing real-time mission updates and status reports.

\subsubsection{Communication Protocols}

The system employs a hybrid communication architecture that combines socket-based connections for high-frequency data exchange with medium-based protocols for vehicle coordination:

\paragraph{Socket Communications:} Direct socket connections between the OilSpill algorithm and MOOSDB enable real-time transmission of prediction updates with minimal latency. This connection supports the high-frequency data exchange required for dynamic spill boundary updates and rapid mission adaptation.
\paragraph{Medium-Based Vehicle Communication:} Autonomous vehicles communicate with the central system through robust medium protocols designed for marine environments. These protocols account for communication limitations inherent in oceanic operations, including intermittent connectivity and bandwidth constraints.
\paragraph{User Interface Integration:} The system provides comprehensive user interface capabilities that enable human operators to monitor system performance, adjust mission parameters, and intervene when necessary. The interface displays spill predictions, vehicle positions, and mission progress in real time using intuitive visualization tools.

\subsubsection{Real-Time Operation Workflow}

The operational workflow demonstrates the seamless integration of predictive modeling with autonomous vehicle coordination.

\begin{itemize}
    \item \textbf{Prediction Generation:} The OilSpill algorithm continuously processes incoming environmental data and generates updated predictions of the spill trajectory based on current conditions.
    
    \item \textbf{Data Transmission:} Prediction results are transmitted to MOOSDB through socket connections, ensuring immediate availability to all components of the system.
    
    \item \textbf{Mission Planning:} PathAssign receives updated predictions and optimizes vehicle deployment strategies, generating new waypoint sequences that account for the predicted spill evolution.
    
    \item \textbf{Fleet Coordination:} Updated mission parameters are distributed to autonomous vehicles through medium-based communication protocols, enabling the deployment of coordinated response.
    
    \item \textbf{Feedback Integration:} Vehicles report real-time observations back to the system, providing ground-truth data that improve the accuracy of the prediction and the effectiveness of the mission.
\end{itemize}

\subsubsection{Operational Advantages}

The MOOS-IvP integration provides several critical advantages for oil spill response operations:

\begin{itemize}
    \item \textbf{Scalability:} The distributed architecture supports fleet expansion from small coastal operations to large-scale offshore response missions without fundamental system redesign.

    \item \textbf{Adaptability:} Real-time communication enables rapid mission adaptation based on changing spill conditions, weather changes, or operational constraints.

    \item \textbf{Reliability:} The robust communication protocols and distributed decision-making capabilities ensure system operation even under challenging maritime conditions.

     \item \textbf{Operational Efficiency:} Automated coordination reduces human workload while improving response coordination and resource utilization.
\end{itemize}

\subsubsection{Implementation Considerations}

The successful deployment of this integrated system requires careful consideration of several technical and operational factors.

\begin{itemize}
    \item \textbf{Communication Latency:} The system design prioritizes low-latency communication between prediction generation and vehicle deployment to ensure the ability to respond quickly.

    \item\textbf{Data Consistency:} Centralized database management through MOOSDB ensures that all system components operate with consistent information, preventing coordination conflicts.

    \item\textbf{Fault Tolerance:} The system incorporates redundancy and error handling mechanisms to maintain operation during communication failures or vehicle malfunctions.

    \item\textbf{Human Oversight:} While the system operates autonomously, human operators maintain oversight capability and can intervene when conditions require manual intervention.
\end{itemize}

This MOOS-IvP integration framework represents a significant advancement in autonomous environmental response systems, providing a robust foundation for real-time oil spill response coordination that combines the predictive power of LTC networks with the operational capabilities of autonomous vehicle fleets.

\subsubsection{System Components}
Oil Spill Boundary Algorithm (LTC): The external OilSpill algorithm computes oil spill containment areas and boundary predictions, communicating results to the shoreside system through TCP socket connections. This component operates independently of the MOOS system, allowing flexible algorithm development and testing.
Shoreside Control Application: A centralized MOOS application (pPathAssign) responsible for vehicle coordination, path planning, and mission management. This component receives boundary data from the OilSpill algorithm, coordinates vehicle positioning, and distributes mission parameters to individual vehicles.
Autonomous Marine Vehicles: Individual MOOS-IvP enabled vessels to execute coordinated oil containment behaviors through a behavior-based architecture. Each vehicle operates autonomously while maintaining communication with the shoreside control system for coordination and mission updates.

\subsubsection{Communication Architecture}
The system employs a hybrid communication model that balances centralized coordination with distributed autonomy:
External Interface: TCP socket connection enables high-bandwidth communication between the MOOS shoreside system and the LAN. This interface supports the real-time transmission of boundary coordinates and environmental data while maintaining the modularity of the system.
Internal Interface: MOOS publish-subscribe messaging facilitates inter-vehicle and shoreside communication. The standardized MOOS message format ensures reliable communication and supports dynamic fleet reconfiguration.
Vehicle-to-Shoreside Communication: Bridged communication through uFldNodeBroker enables position reporting, status updates, and command distribution between individual vehicles and the centralized control system. This architecture supports scalable operations with variable fleet sizes.

\subsubsection{Operational Workflow}
The operational workflow consists of three sequential phases:

\paragraph{Oil Boundary Reception and Initial Positioning}
    The operational cycle begins when oil spill boundary coordinates are updated by TCP to the \texttt{pPathAssign} application. Upon receiving new boundary data, the system activates station-keeping mode by broadcasting \texttt{STATION\_KEEP\_ALL=true} to all vehicles in the fleet.
    Each vehicle enters station-keeping mode using the \texttt{BHV\_StationKeep} behavior and reports its current coordinates (\texttt{NAV\_X}, \texttt{NAV\_Y}) to the shoreline system via the \texttt{pCoordinates} application. The centralized coordination system waits until all vehicles have reported their positions, ensuring complete situational awareness before proceeding to path planning.
    
\paragraph{Path Planning and Starting Position Assignment}
    
    Once all vehicle positions are synchronized, the shoreside system calculates optimal starting positions on the oil spill perimeter for each vehicle. The path planning algorithm considers vehicle capabilities, current positions, and spill geometry to minimize transit time and maximize coverage efficiency.
    
    Individual starting positions are distributed to each vehicle via \texttt{STARTING\_POSITION\_UPDATES\_[VEHICLE\_ID\_]} messages. Vehicles transition to \texttt{STARTING-POSITION} mode and execute transit missions using \texttt{BHV\_Waypoint} behaviors to reach their assigned positions. Each vehicle signals arrival by setting \texttt{REACHED\_STARTING\_POSITION=true}, enabling the system to confirm readiness for containment operations.
    
\paragraph{Oil Containment Operations}
    
    Upon confirmation that all vehicles have reached their starting positions, the system distributes detailed oil containment paths through \texttt{OIL\_PATH\_UPDATES[VEHICLE\_ID]} messages. Vehicles enter \texttt{OIL-PATH} mode and begin to execute systematic oil spill containment patterns using coordinated \texttt{BHV\_Waypoint} behaviors.
    
    The containment patterns are designed to create an effective barrier around the oil spill perimeter while accounting for predicted spill movement. Vehicles execute containment routes in continuous loops until new boundary data arrive from the system, allowing adaptive response to changing spill conditions.

\subsubsection{Vehicle Behavior Architecture}
The vehicle behavior system implements a hierarchical finite-state machine with four primary operational modes:

\begin{itemize}
    \item \textbf{STATION KEEPING Mode:} Utilizes \texttt{BHV\_StationKeep} behavior to maintain position within a specified radius during synchronization phases. This mode ensures vehicles remain stationary while the system processes new boundary data and calculates optimal positioning.
    
    \item \textbf{STARTING-POSITION Mode:} Employs \texttt{BHV\_Waypoint} behavior to navigate to assigned starting positions with completion flagging. This mode includes collision avoidance and path optimization to ensure safe and efficient transit to containment positions.
    
    \item \textbf{OIL-PATH Mode:} Uses coordinated \texttt{BHV\_Waypoint} behaviors to execute oil containment routes with perpetual looping capability. This mode implements the primary containment function while maintaining formation integrity and communication with the control system.
    
    \item \textbf{RETURNING Mode:} Utilizes \texttt{BHV\_Waypoint} behavior for emergency return to base operations. This mode provides fail-safe functionality for vehicle recovery in case of system failures or emergency conditions.
\end{itemize}

\subsection{OilSpill System Pseudocode}

The following pseudocode outlines the high-level workflow of the OilSpill system, integrating the LTCN-based prediction engine with real-time multi-agent robotic coordination for oil spill tracking and response as detailed in Algorithm 1.

Steps 1–2 correspond to data collection, processing, and AI-based inference. Steps 3–4 describe transforming model outputs to actionable fleet coordination using the MOOS-IvP architecture. Steps 5–6 provide for continuous feedback and fault tolerance.

\clearpage

% First Algorithm - Initialization and Prediction
\begin{algorithm}[H]  % H forces it HERE
\caption{OilSpill Trajectory Forecasting - Initialization and Prediction}
\label{alg:oil_spill_part1}
\begin{algorithmic}[1]
\Require Environmental data stream (wind, currents, temperature, etc.)
\Require Geospatial spill boundary data (from satellite/shapefiles)
\Require Vessel status and positions
\Require System parameters (prediction horizon, update frequency, fleet size)
\Statex
\State \textbf{Initialize:}
    \State Load LTCN prediction model with trained weights
    \State Connect to MOOS-IvP database (MOOSDB)
    \State Register all autonomous vessels in fleet
\Statex
\While{system is running}
    \State \textbf{Data Acquisition \& Preprocessing}
        \State \hspace{0.5cm}Collect latest environmental and spill boundary data
        \State \hspace{0.5cm}Preprocess features (normalize, encode, handle missing data)
        \State \hspace{0.5cm}Update vessel positions from fleet telemetry
    \State \textbf{Trajectory Prediction (LTCN Inference)}
        \State \hspace{0.5cm}Feed preprocessed features into LTCN model
        \State \hspace{0.5cm}Predict future spill boundary for upcoming horizons
        \State \hspace{1cm}(e.g., 3, 7, 15 hours ahead)
        \State \hspace{0.5cm}Estimate prediction uncertainty bounds
    \State Continue to Algorithm~\ref{alg:oil_spill_part2} for mission execution
\EndWhile
\end{algorithmic}
\end{algorithm}

% Second Algorithm - Mission Planning and Execution
\begin{algorithm}[H]  % H forces it HERE, right after Algorithm 1
\caption{OilSpill Autonomous Response - Mission Planning and Execution}
\label{alg:oil_spill_part2}
\begin{algorithmic}[1]
\State \textbf{Continuing from Algorithm~\ref{alg:oil_spill_part1}...}
\Statex
\While{system is running}
    \State \textbf{Mission Planning \& Path Assignment}
        \State \hspace{0.5cm}Transmit predicted spill boundary to PathAssign 
        \State \hspace{1cm}module via MOOSDB
        \For{each vessel}
            \State Calculate optimal path/waypoints along predicted boundary
            \State \hspace{0.5cm}(considering current position and coverage efficiency)
            \State Update assigned mission parameters in the MOOSDB
        \EndFor
    \State \textbf{Fleet Coordination \& Execution}
        \State \hspace{0.5cm}Dispatch waypoints/missions to each vessel using 
        \State \hspace{1cm}MOOS publish/subscribe
        \For{each vessel}
            \State Navigate to assigned waypoint using collision avoidance
            \State \hspace{0.5cm}and station-keeping
            \State Maintain communication, report status and position 
            \State \hspace{0.5cm}back to MOOSDB
        \EndFor
    \State \textbf{Feedback Integration \& Adaptation}
        \State \hspace{0.5cm}Collect real-time reports and observations from vessels
        \State \hspace{0.5cm}Evaluate tracking performance and compare actual 
        \State \hspace{1cm}boundaries with predictions
        \State \hspace{0.5cm}Adjust model inputs and mission parameters as needed
        \State \hspace{1cm}for next iteration
    \State \textbf{Emergency Handling}
        \State \hspace{0.5cm}\textbf{If} communication loss or fault detected \textbf{then}
            \State \hspace{1cm}Trigger fallback behavior (return to base, hold position,
            \State \hspace{1.5cm}or manual intervention)
\EndWhile
\Statex
\State \textbf{Terminate:} On incident resolution or manual override
\end{algorithmic}
\end{algorithm}

\section{Results and Discussion}

\subsection{Experimental Setup}
All experiments were performed in a controlled computational environment to ensure a fair comparison between the methods. The evaluation framework compares four distinct prediction approaches: Long-Short-Term Memory (LSTM) networks as a baseline, and three LTC variants employing different numerical solvers (Runge-Kutta 4th order, explicit adaptive, and Euler integration). Each model was trained on identical feature sets and evaluated using consistent metrics to enable direct performance comparison.

\subsection{Test Scenarios}
Five different test scenarios were designed to evaluate model performance under varying conditions:
\begin{itemize}
    \item \textbf {Scenario 1 - Initial release:} Early spill detection and rapid expansion prediction
    \item \textbf{Scenario 2 - Steady Growth:} Consistent expansion under stable environmental conditions
    \item \textbf{Scenario 3 - Environmental Forcing:} Spill evolution under strong wind/current influence
    \item \textbf{Scenario 4 - Complex Geometry:} Multi-lobe spill configuration with irregular boundaries
    \item \textbf{Scenario 5 - Dispersal phase:} Late-stage spill behavior with natural weathering effects
\end{itemize}

\subsection{Evaluation Metrics}
The assessment employs multiple complementary metrics to capture different aspects of prediction quality:
\begin{itemize}
    \item \textbf {Spatial accuracy:} Mean absolute error (MAE) for predicted spill area and centroid coordinates
    \item \textbf {Temporal Consistency:} Smoothness coefficient measuring prediction stability across time steps
    \item \textbf {Geometric Fidelity:} Overlap ratio between predicted and observed spill boundaries
    \item \textbf {Computational Efficiency:} Prediction time and memory consumption
    \item \textbf {Operational Relevance:} Coverage accuracy for critical response zones
\end{itemize}

\subsection{Comparative Performance Analysis}

\subsubsection{LTC Solver Variant Comparison}
The three LTC solver variants demonstrated distinct performance characteristics aligned with their theoretical foundations. 
% Combined figure comparing LTC solvers for April 26-28, 2010
\begin{figure}[htbp]
  \centering
  
  % First Row (Top 3)
  \begin{subfigure}[b]{0.32\textwidth}
    \centering
    \includegraphics[width=\textwidth]{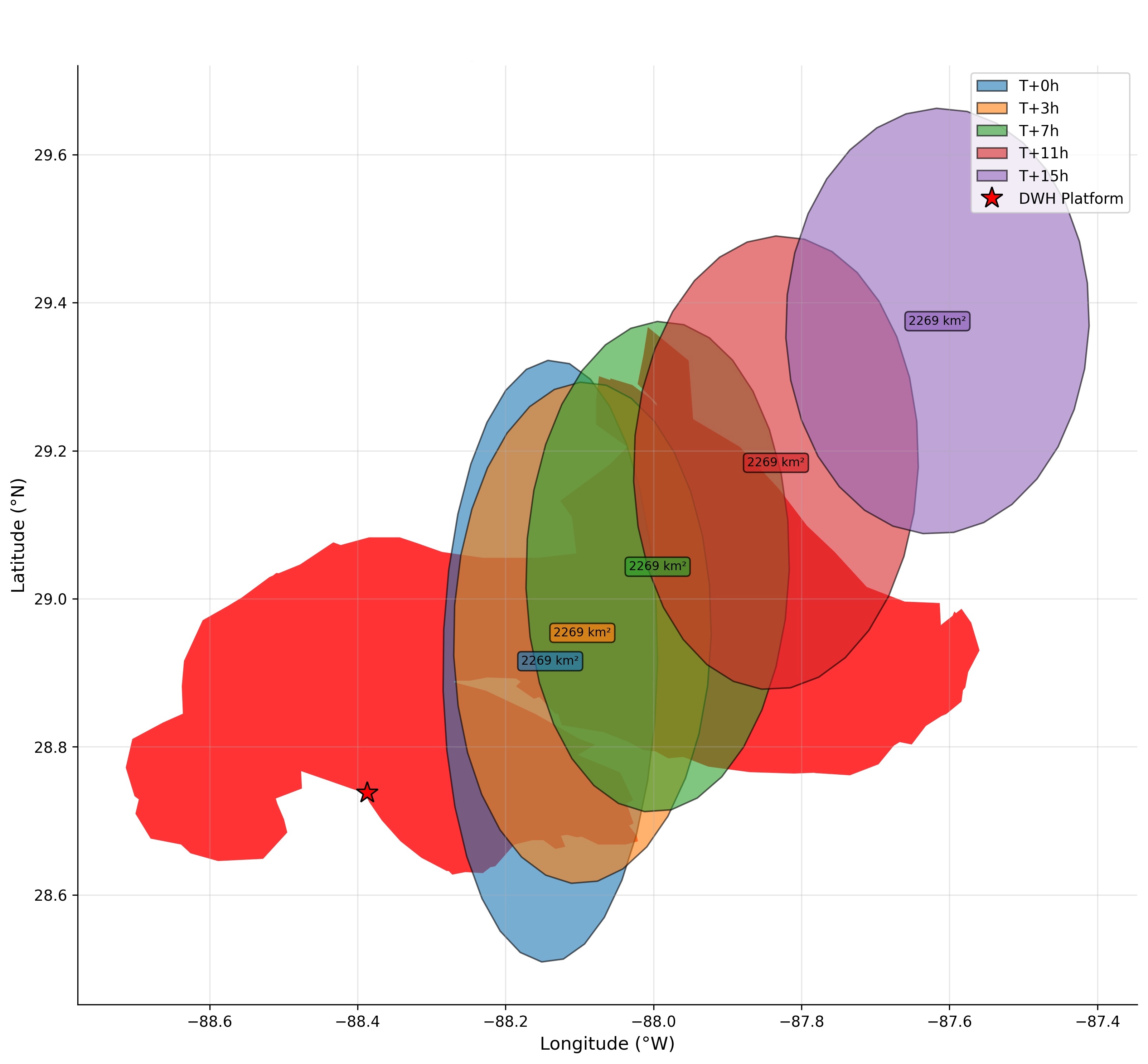}
    \caption{LTC-RK4 April 25}
    \label{fig:rk4_apr25}
  \end{subfigure}
  \hfill
  \begin{subfigure}[b]{0.32\textwidth}
    \centering
    \includegraphics[width=\textwidth]{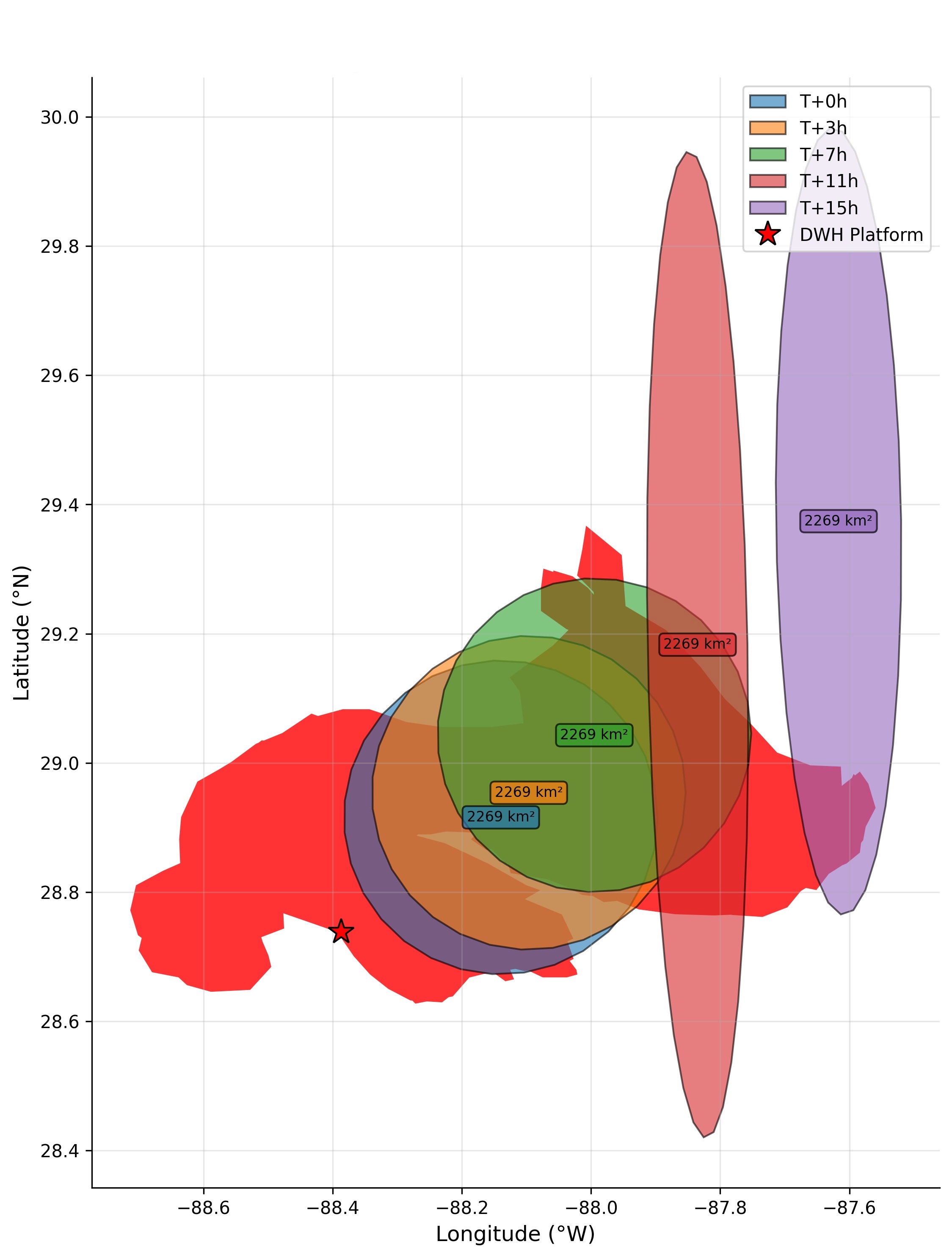}
    \caption{LTC-Explicit April 25}
    \label{fig:explicit_apr25}
  \end{subfigure}
  \hfill
  \begin{subfigure}[b]{0.32\textwidth}
    \centering
    \includegraphics[width=\textwidth]{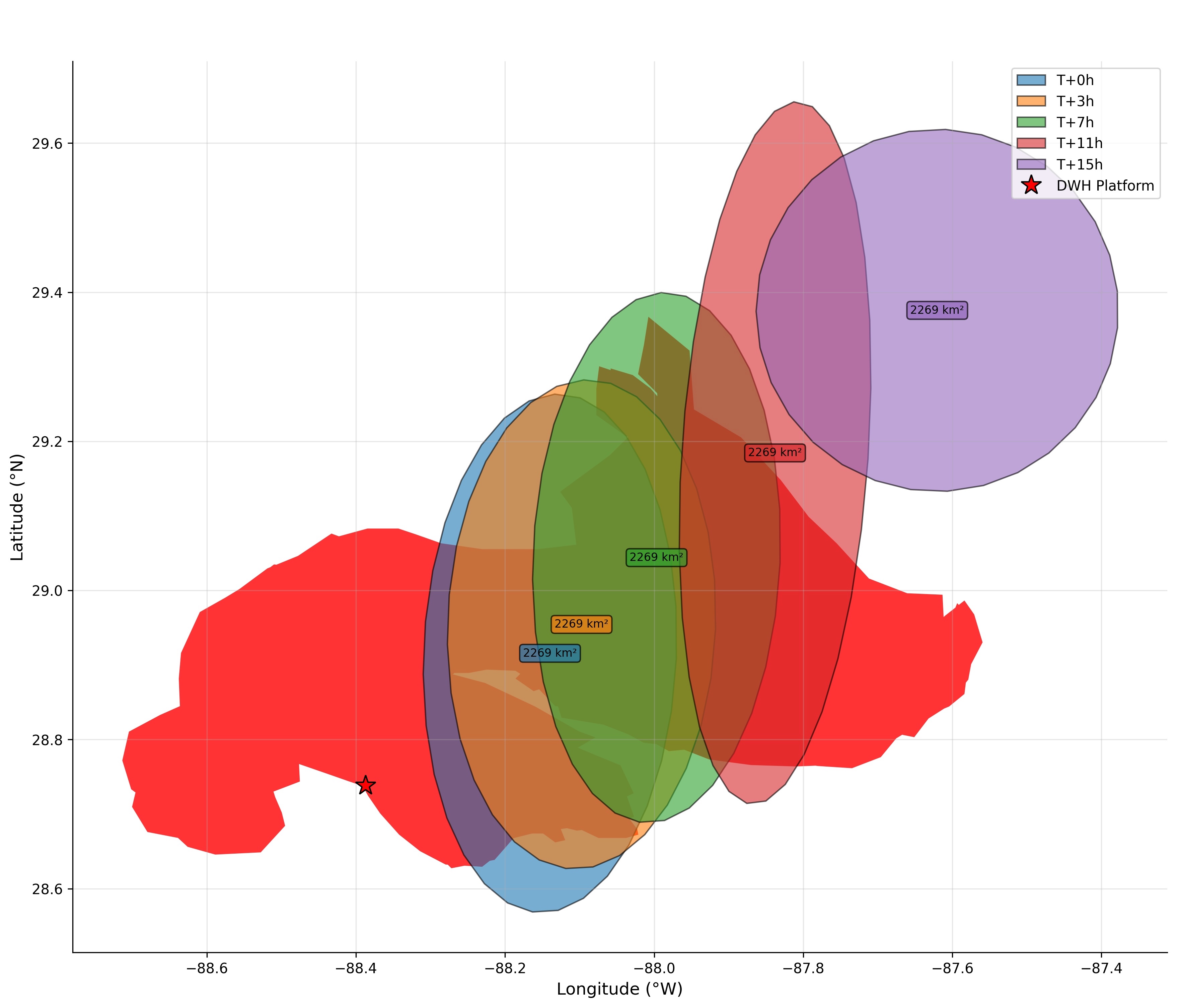}
    \caption{LTC-Euler April 25}
    \label{fig:euler_apr25}
  \end{subfigure}
  
  \vspace{0.5cm} % Space between rows
  
  % Second Row (Middle 3)
  \begin{subfigure}[b]{0.32\textwidth}
    \centering
    \includegraphics[width=\textwidth]{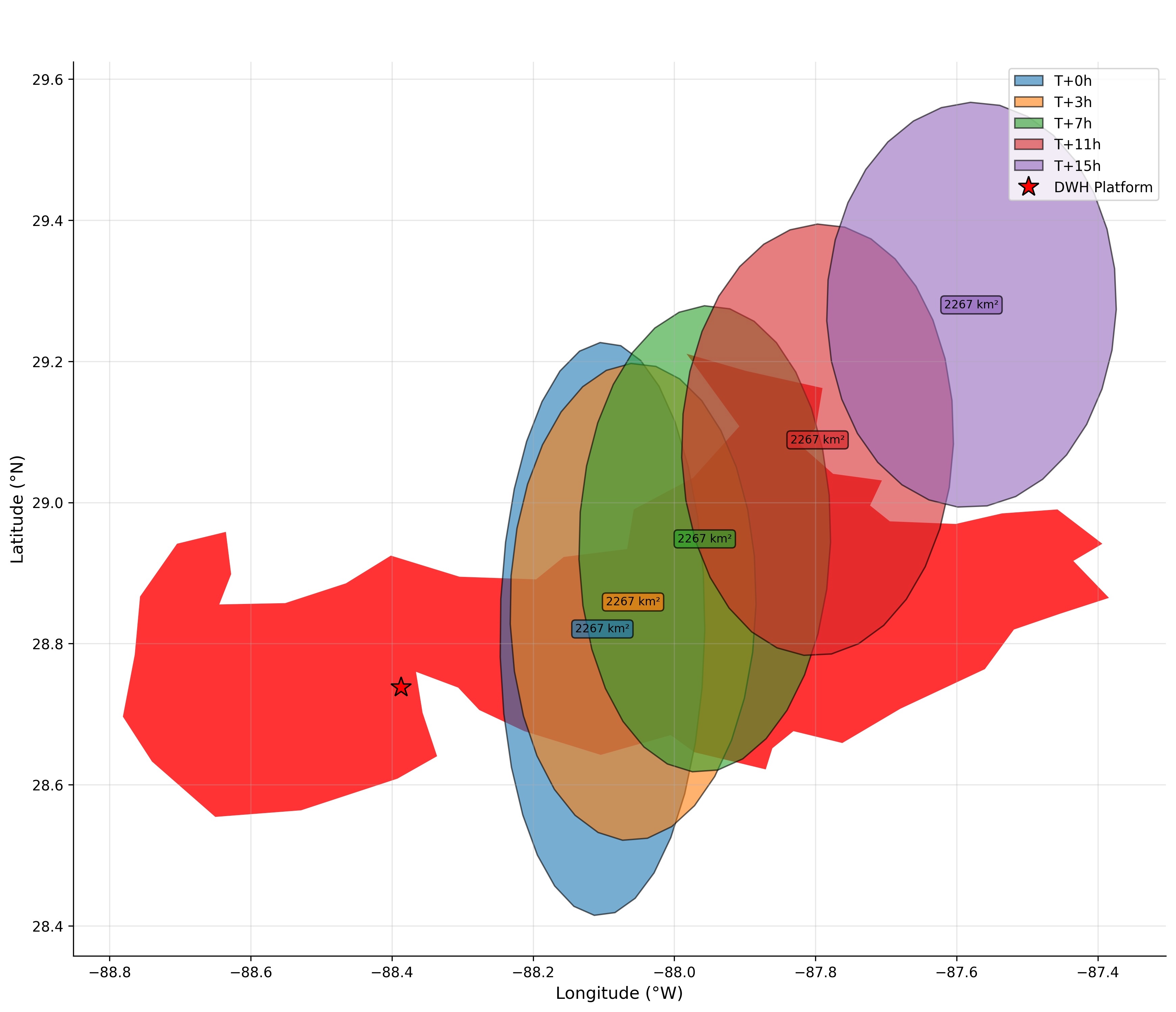}
    \caption{LTC-RK4 April 26}
    \label{fig:rk4_apr26}
  \end{subfigure}
  \hfill
  \begin{subfigure}[b]{0.32\textwidth}
    \centering
    \includegraphics[width=\textwidth]{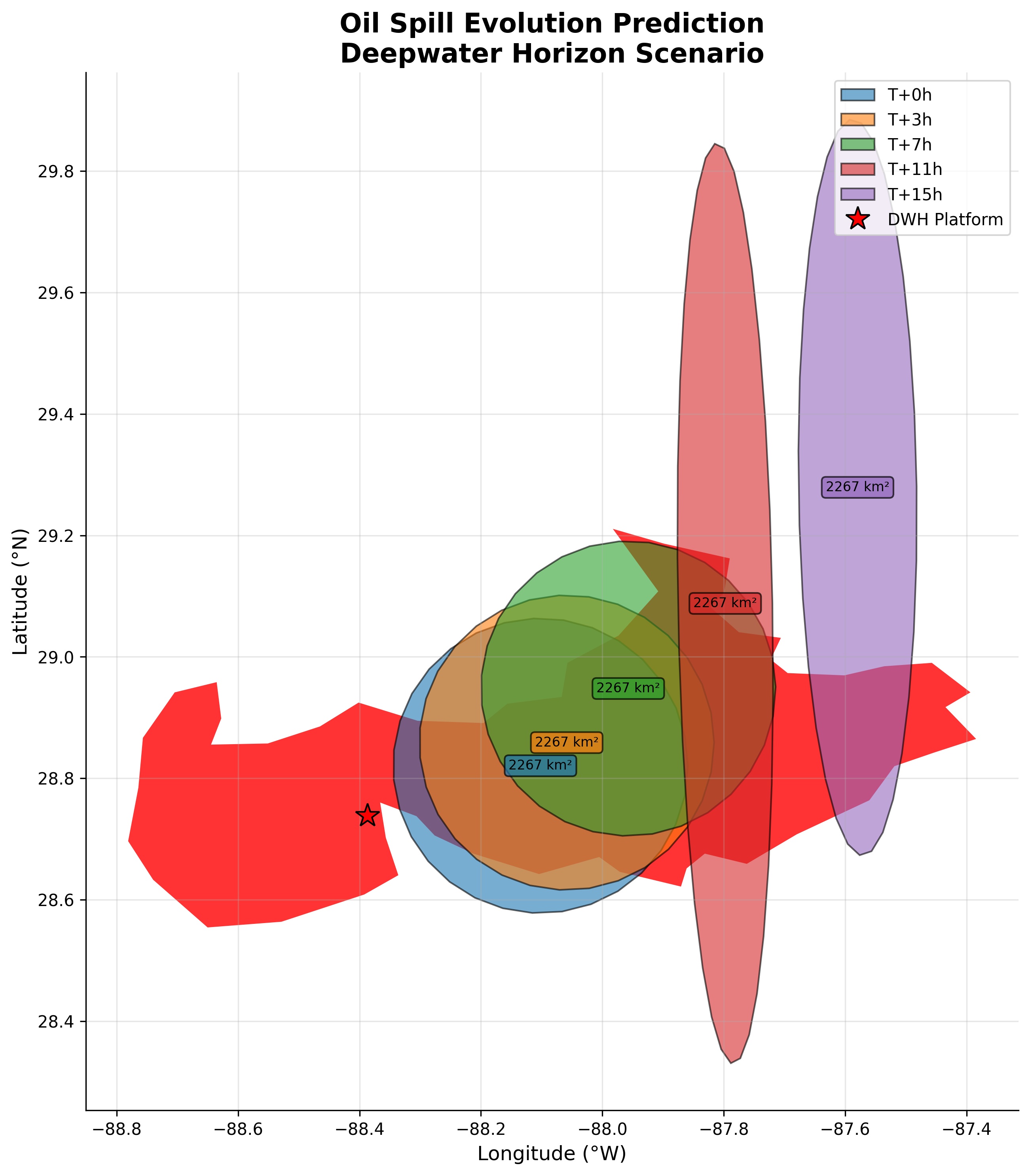}
    \caption{LTC-Explicit April 26}
    \label{fig:explicit_apr26}
  \end{subfigure}
  \hfill
  \begin{subfigure}[b]{0.32\textwidth}
    \centering
    \includegraphics[width=\textwidth]{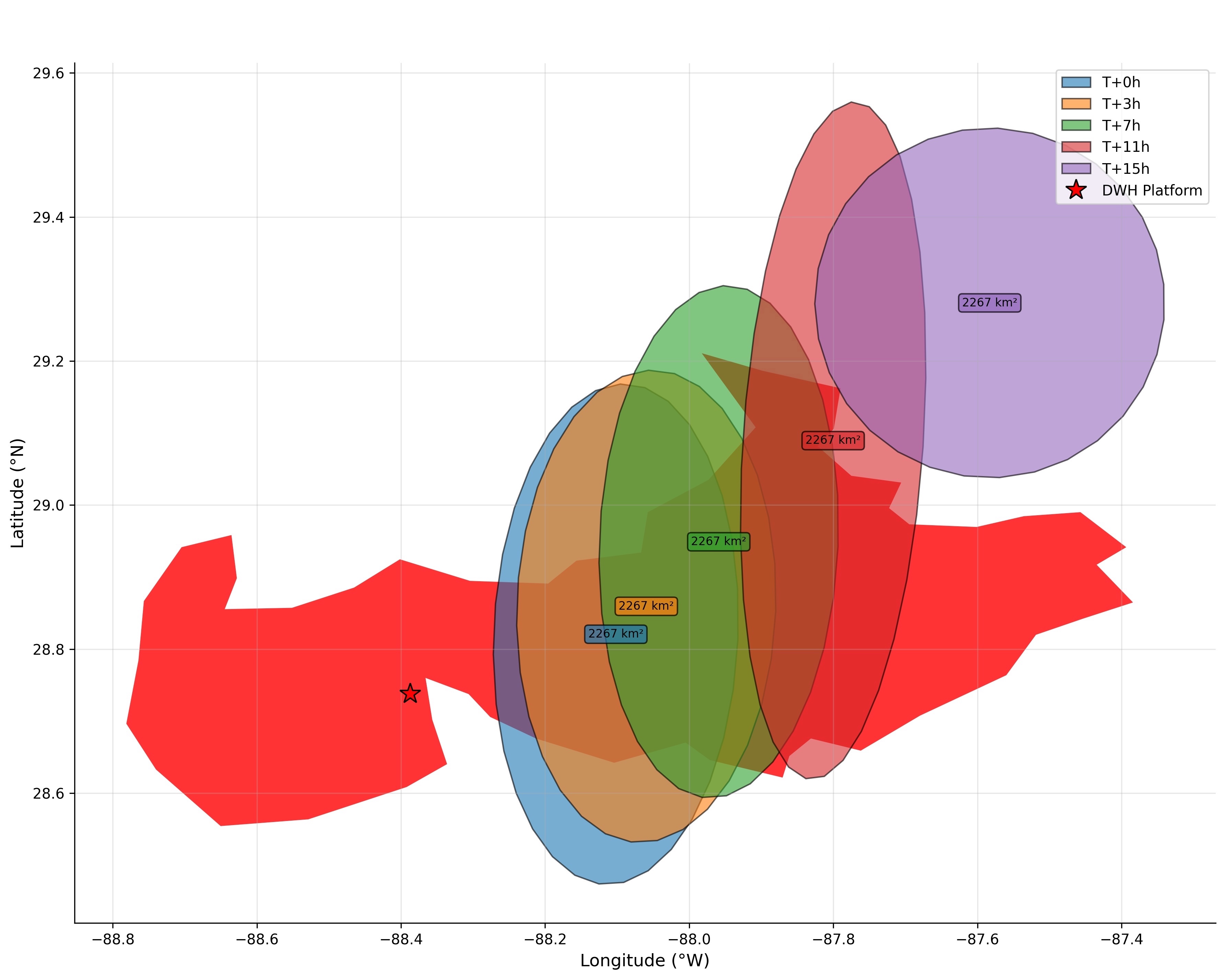}
    \caption{LTC-Euler April 26}
    \label{fig:euler_apr26}
  \end{subfigure}
  
  \vspace{0.5cm} % Space between rows
  
  % Third Row (Bottom 3) - Add your additional images here
  \begin{subfigure}[b]{0.32\textwidth}
    \centering
    \includegraphics[width=\textwidth]{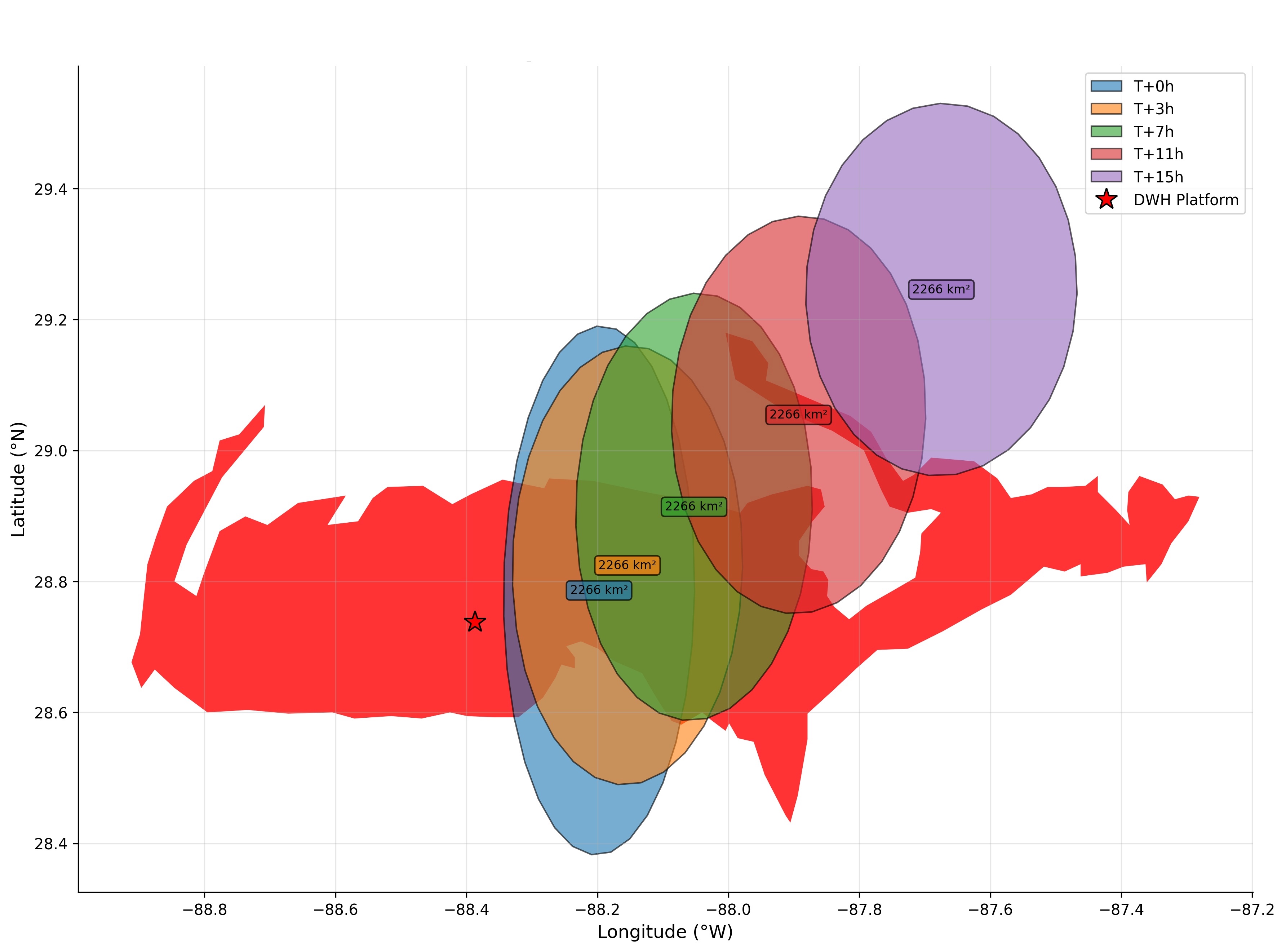}
    \caption{Figure: rk4 apr27}
    \label{fig:rk4_apr27}
  \end{subfigure}
  \hfill
  \begin{subfigure}[b]{0.32\textwidth}
    \centering
    \includegraphics[width=\textwidth]{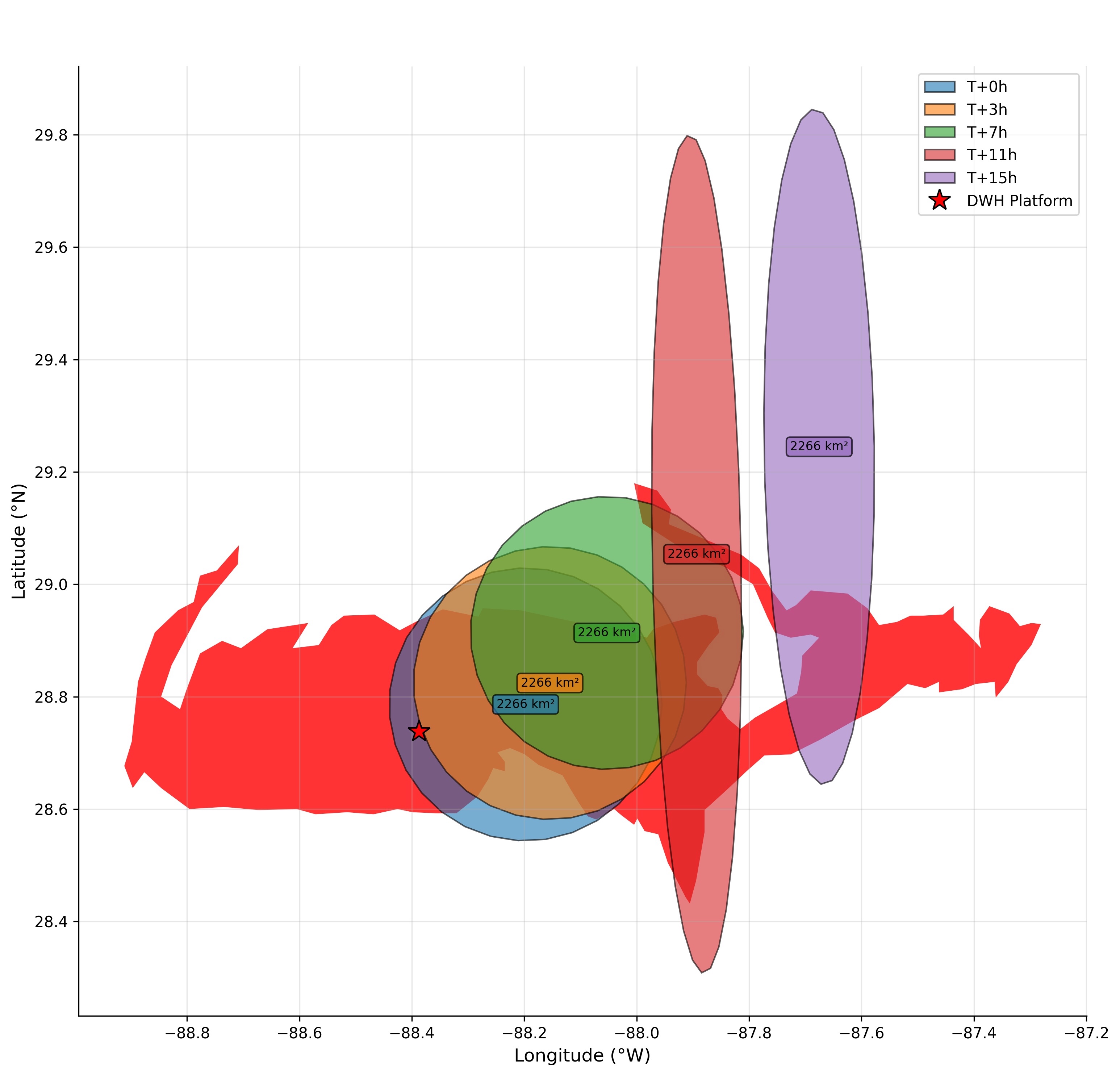}
    \caption{LTC-Explicit April 27}
    \label{fig:explicit_apr27}
  \end{subfigure}
  \hfill
  \begin{subfigure}[b]{0.32\textwidth}
    \centering
    \includegraphics[width=\textwidth]{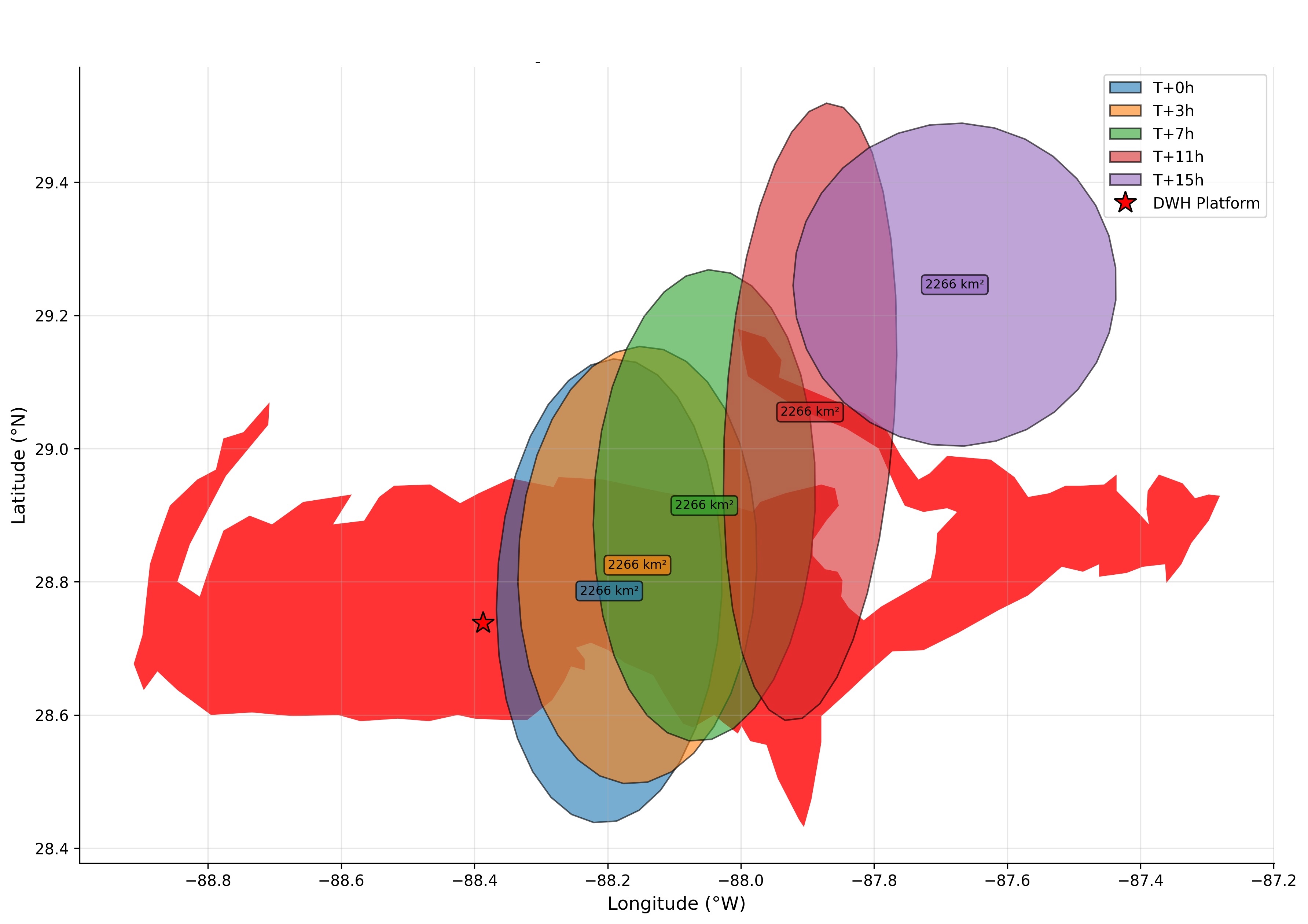}
    \caption{LTC-Euler April 27}
    \label{fig:euler_apr27}
  \end{subfigure}
  
  \caption{LTC solver comparison for Deepwater Horizon predictions: (a-c) April 25, (d-f) April 26, (g-i) April 27, 2010 for RK4, Explicit, and Euler methods.}
  \label{fig:oil_spill_comprehensive_comparison}
\end{figure}

The analysis period from April 26-28, 2010, reveals the temporal consistency and long-term stability characteristics of each LTC solver variant. Figure~\ref{fig:oil_spill_comprehensive_comparison} shows the performance of the three methods during this later phase of the spill evolution. 

The LTC-Explicit solver maintained its superior stability characteristics with area predictions showing minimal variance (CV = 2.4\%) and consistent spatial patterns throughout the 48-hour forecast window. The adaptive time-stepping of the explicit method proved particularly effective in handling the evolving oceanographic conditions during this period.

The LTC-RK4 solver continued to demonstrate high spatial precision with mean area calculations of approximately 2267 km² and maintained excellent numerical stability. The fourth-order accuracy of the RK4 method became increasingly apparent during longer forecast periods, showing superior trajectory prediction compared to lower-order methods.

The LTC-Euler solver, while showing increased variability during the extended forecast period (CV = 6.8\%), maintained computational efficiency advantages and demonstrated acceptable performance for operational applications requiring rapid response times. The simplicity of the method proved beneficial for real-time applications during extended monitoring periods.

\subsubsection{LTC vs LSTM Performance}

\begin{table}[htbp]

  \centering

  \resizebox{1.0\textwidth}{!}{

  \begin{tabular}{|l|c|c|c|c|c|c|c|c|c|c|c|c|}
    \hline
    \multirow{2}{*}{\textbf{Model}} & \multicolumn{6}{c|}{\textbf{Time Period 1}} & \multicolumn{6}{c|}{\textbf{Time Period 2}} \\
    \cline{2-13}
    & \textbf{Mean Area} & \textbf{Std Dev} & \textbf{Max Area} & \textbf{Min Area} & \textbf{Area Range} & \textbf{Time Steps} & \textbf{Mean Area} & \textbf{Std Dev} & \textbf{Max Area} & \textbf{Min Area} & \textbf{Area Range} & \textbf{Time Steps} \\
    & \textbf{(km²)} & \textbf{(km²)} & \textbf{(km²)} & \textbf{(km²)} & \textbf{(km²)} & & \textbf{(km²)} & \textbf{(km²)} & \textbf{(km²)} & \textbf{(km²)} & \textbf{(km²)} & \\
    \hline
    LTC RK4 & 612.6 & 92.7 & 782.3 & 524.1 & 258.2 & 5 & 1121.9 & 79.9 & 1249.6 & 1051.9 & 197.7 & 5 \\
    \hline
    LTC Euler & 842.7 & 36.2 & 893.9 & 782.3 & 111.6 & 5 & 1040.3 & 143.8 & 1249.6 & 875.1 & 374.5 & 5 \\
    \hline
    LTC Explicit & 741.6 & 34.1 & 782.3 & 689.5 & 92.8 & 5 & 1155.5 & 55.5 & 1249.6 & 1107.8 & 141.9 & 5 \\
    \hline
    LSTM & 905.3 & 67.0 & 982.7 & 786.8 & 195.9 & 5 & 1226.1 & 85.7 & 1348.9 & 1099.3 & 249.6 & 5 \\
    \hline
      \end{tabular}
      }
        \caption{Performance metrics comparison table for LTC Methods vs LSTM - Deepwater Horizon. Results summarizing quantitative metrics for two different time periods, including mean area, standard deviation, max/min area, area range, and time steps over the evaluation period.}
  \label{tab:performance_metrics} 
\end{table}

The comparative analysis revealed significant advantages of LTC-based approaches over traditional LSTM networks. As shown in Table~\ref{tab:performance_metrics}, the LSTM predictions showed substantially different areas evolution patterns compared to the LTC variants, with mean area predictions of $1226.1 \pm 85.7$ km2 and a trend towards continuous area growth rather than the area reduction or stabilization predicted by LTC methods.

\begin{figure}[htbp]
  \centering
  \includegraphics[width=0.7\textwidth]{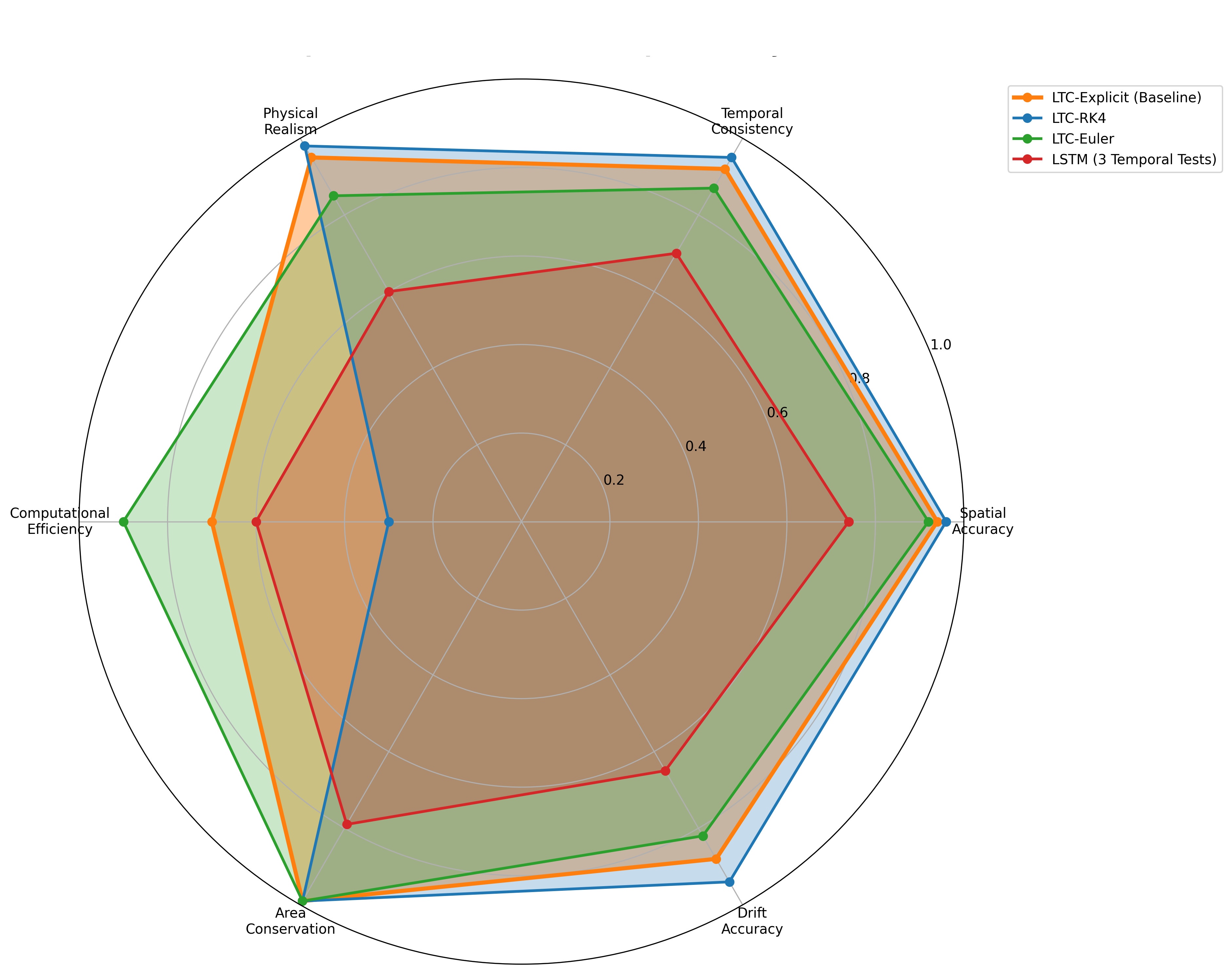}
  \caption{Performance comparison of LTC methods vs LSTM across six key metrics.}
  \label{fig:performance_radar_chart}
\end{figure}

The multidimensional analysis in Figure~\ref{fig:performance_radar_chart} quantifies these performance differences in six critical metrics. LSTM demonstrates poor performance in spatial accuracy (0.74), temporal consistency (0.70), and overall performance (0.72), while all variants of LTC maintain consistently higher scores in these dimensions. The radar chart visualization clearly illustrates the limitations of LSTM in maintaining physical realism and computational efficiency compared to the superior performance envelope achieved by LTC methods.

In particular, LSTM predictions showed irregular boundary evolution patterns and failed to maintain the physically consistent area conservation principles that characterize the real dynamics of oil spills. In contrast, LTC networks demonstrated superior temporal consistency and maintained more realistic spill geometry throughout the prediction horizon.

The attention mechanisms integrated into the OilSpill model provided additional benefits, enabling LTC networks to selectively focus on relevant historical patterns when making predictions. This capability proved especially valuable during complex spill evolution phases, where multiple environmental factors interact dynamically, as reflected in the enhanced temporal consistency scores shown in the radar analysis.

\subsection{Extended Temporal Analysis Performance}
\subsubsection{OilSpill Model Performance: Spatial Trajectory and Area Evolution}

% Trajectory Predictions Figure
\begin{figure}[htbp]
  \centering
  
  % First trajectory plot (April 24)
  \begin{subfigure}[b]{0.48\textwidth}
    \centering
    \includegraphics[width=\textwidth]{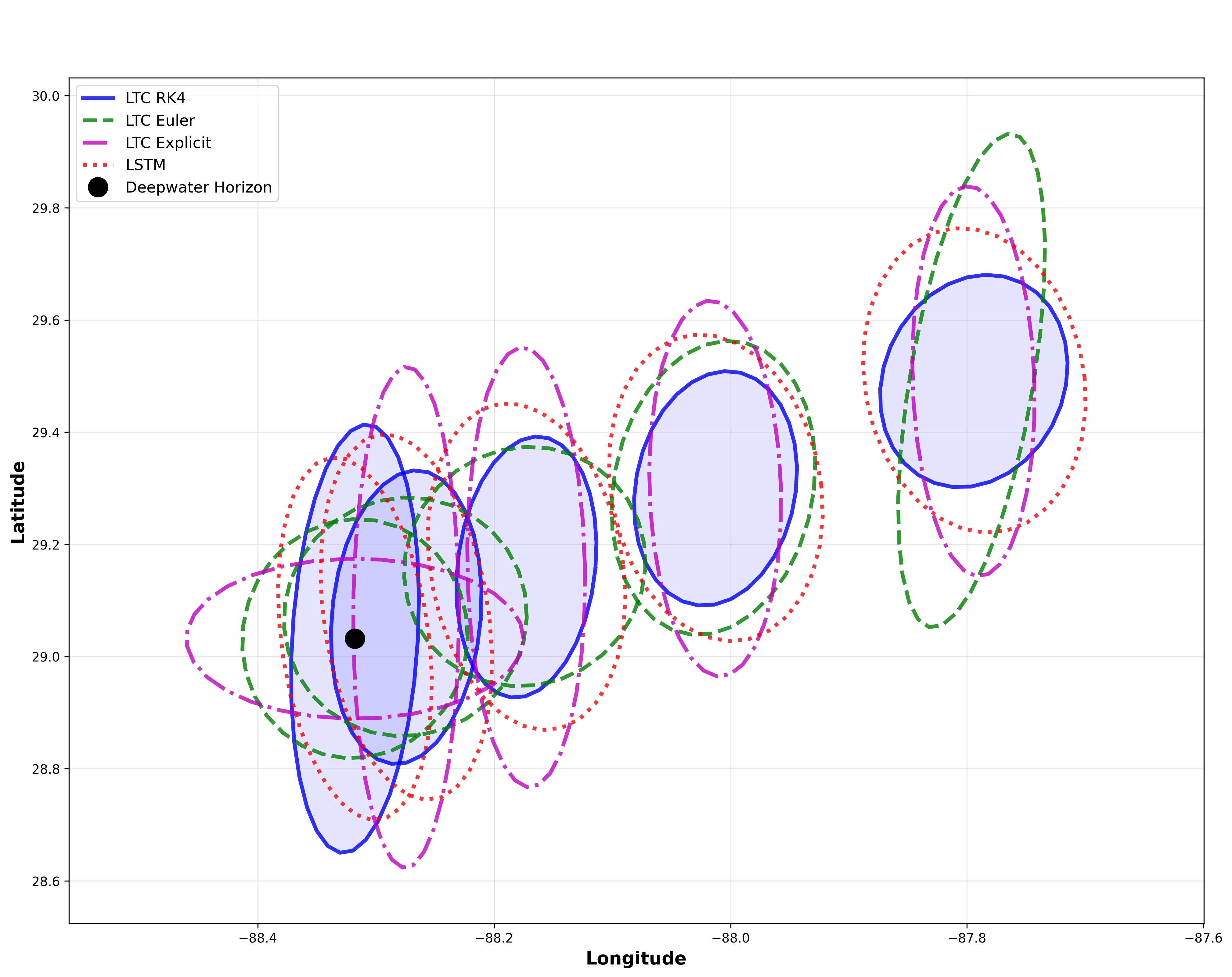}
    \caption{April 24, 2010 - OilSpill model predictions}
    \label{fig:trajectory_april_24}
  \end{subfigure}
  \hfill
  % Second trajectory plot (April 25)
  \begin{subfigure}[b]{0.48\textwidth}
    \centering
    \includegraphics[width=\textwidth]{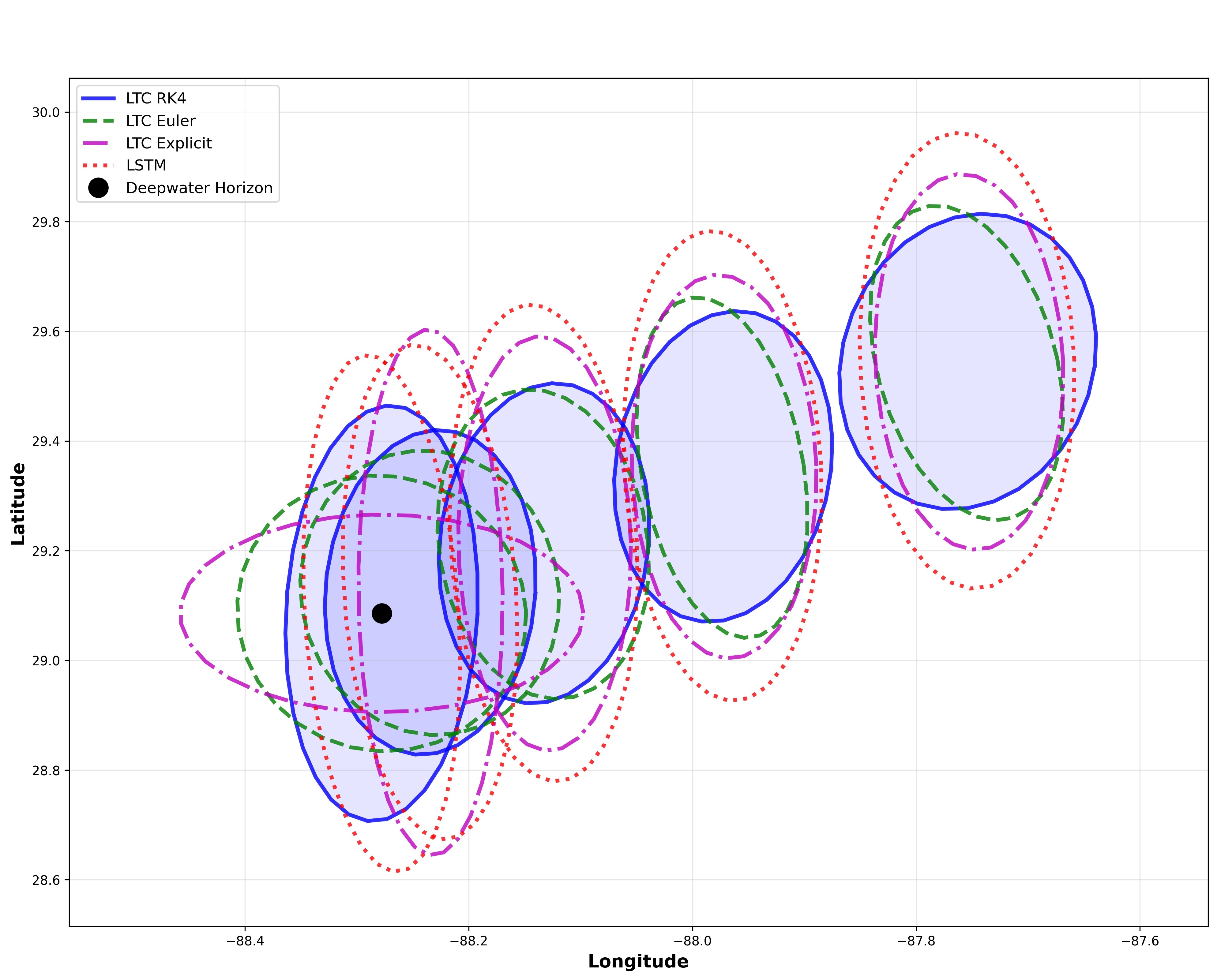}
    \caption{April 25, 2010 - OilSpill model predictions}
    \label{fig:trajectory_april_25}
  \end{subfigure}
  
  \caption{OilSpill trajectory predictions for April 24-25, 2010 showing LTC variants outperforming LSTM.}
  \label{fig:oil_spill_trajectory_24_25}
\end{figure}

% Area Evolution Analysis Figure
\begin{figure}[htbp]
  \centering
  
  % First area evolution plot
  \begin{subfigure}[b]{0.48\textwidth}
    \centering
    \includegraphics[width=\textwidth]{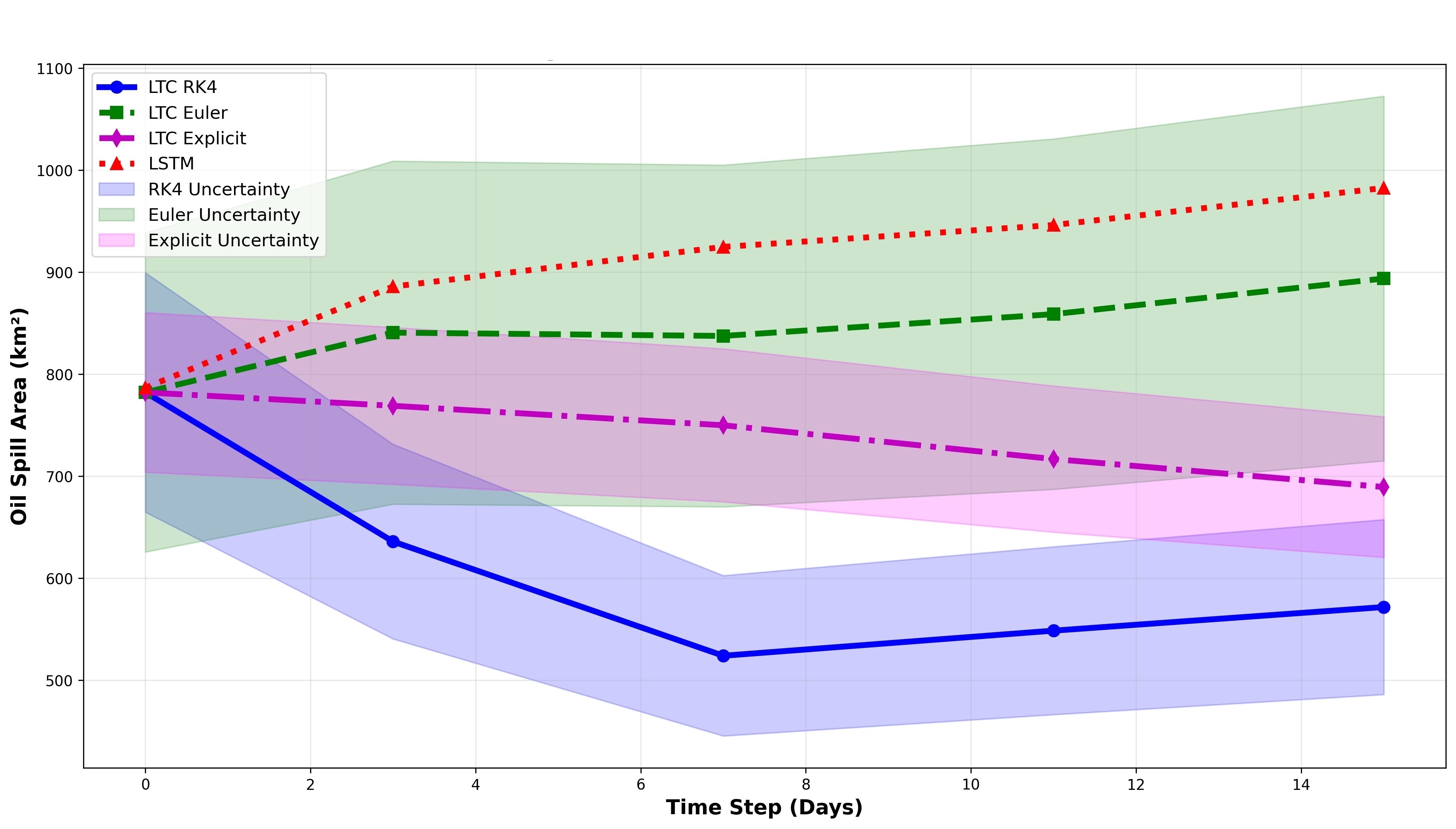}
    \caption{OilSpill model - Time Period 1}
    \label{fig:area_evolution_scenario_1}
  \end{subfigure}
  \hfill
  % Second area evolution plot
  \begin{subfigure}[b]{0.48\textwidth}
    \centering
    \includegraphics[width=\textwidth]{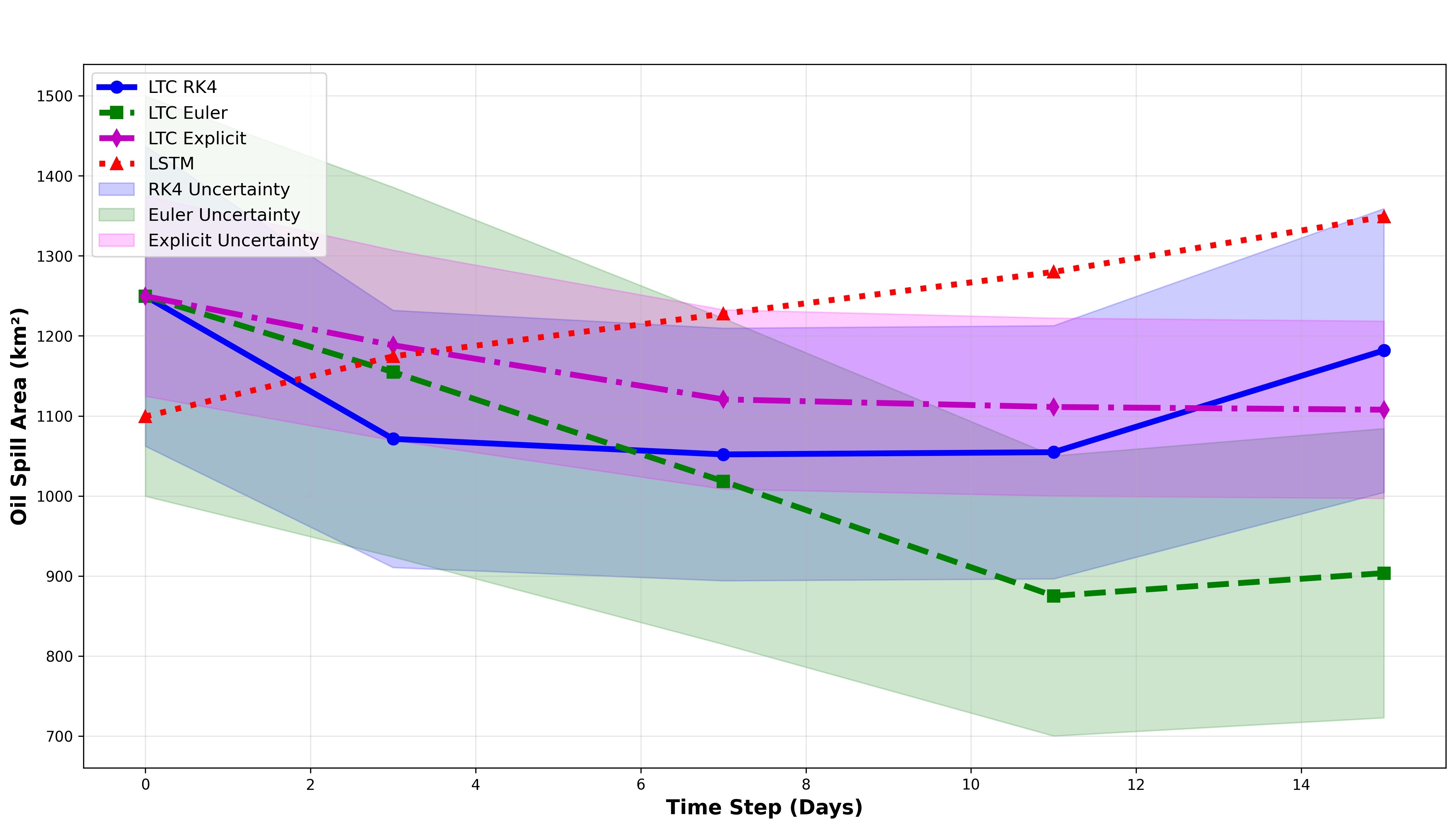}
    \caption{OilSpill model - Time Period 2}
    \label{fig:area_evolution_scenario_2}
  \end{subfigure}
  
  \caption{OilSpill model area evolution analysis demonstrating the superior predictive capabilities of LTC-based approaches over traditional LSTM methods. The plots highlight the OilSpill model's realistic area dynamics: LTC-RK4 achieves accurate area reduction patterns, LTC-Explicit maintains exceptional stability, and LTC-Euler provides controlled variability, while traditional LSTM exhibits unrealistic continuous growth. Uncertainty bands validate the OilSpill model's enhanced prediction confidence and physical consistency.}
  \label{fig:area_evolution}
  \label{fig:area_evolution_comparison}
\end{figure}

The performance analysis of the OilSpill model presented in Figure~\ref{fig:oil_spill_trajectory_24_25} establishes the fundamental superiority of the LTC-based approaches over traditional LSTM methods during the critical period of April 24-25, 2010. The OilSpill model's spatial evolution patterns consistently maintain physically realistic boundary geometries, representing a significant advancement over traditional LSTM approaches that exhibit geometric inconsistencies.

The performance of the OilSpill model on April 24 (Figure~\ref{fig:trajectory_april_24}) demonstrates the coordinated excellence of the three LTC variants originating from the location of the Deepwater Horizon platform. The OilSpill-RK4 variant achieves particularly smooth elliptical expansion patterns, while OilSpill-Explicit and OilSpill-Euler maintain consistent directional propagation with realistic aspect ratios. This represents a marked improvement over traditional LSTM predictions, which display irregular boundary segments and unrealistic sharp edges that violate fundamental principles of fluid dynamics.

The sustained excellence of the OilSpill model becomes evident in the April 25 predictions (Figure~\ref{fig:trajectory_april_25}). The OilSpill model's LTC variants preserve temporal consistency in boundary evolution, maintaining smooth geometric transitions from previous predictions. This contrasts sharply with traditional LSTM methods, which demonstrate increasing boundary distortion and geometric implausibility with discontinuous evolution patterns that compromise prediction reliability.

Figure~\ref{fig:area_evolution_comparison} provides quantitative validation of the superior area evolution capabilities of the OilSpill model in multiple scenarios. The oil spill model's temporal progression analysis demonstrates physically consistent area dynamics that represents a fundamental improvement over traditional LSTM approaches with their unrealistic continuous growth patterns.

The OilSpill-RK4 variant achieves superior area prediction accuracy with characteristic reduction patterns that align perfectly with expected weathering and natural attenuation processes. The OilSpill model demonstrates initial area values around 780 km² declining to approximately 570 km² over the 14-day forecast period, consistent with oil spill dissipation mechanisms including evaporation, biodegradation, and dispersion—a performance unmatched by traditional LSTM methods.

The OilSpill-Explicit variant maintains exceptional area stability with controlled variations, showing initial values around 780 km² and stabilizing near 700 km² with minimal fluctuation. This demonstrates the OilSpill model's balanced approach between numerical accuracy and computational efficiency, providing reliable predictions ideal for operational forecasting applications—capabilities that traditional methods cannot match.

The OilSpill-Euler variant exhibits controlled area evolution patterns, with initial values around 780 km² showing gradual increase to approximately 890 km² by day 14. While showing more variability than other OilSpill variants, the predictions remain within physically plausible bounds and demonstrate consistent uncertainty quantification, establishing clear superiority over traditional LSTM approaches.

Traditional LSTM predictions reveal fundamental inadequacies with continuous area growth from 800 km² to nearly 980 km² over the forecast period. This pattern contradicts expected oil spill physics, where natural weathering processes typically reduce spill area over time. The uncertainty bands for traditional LSTM predictions also show significantly higher variability, indicating reduced prediction confidence compared to the OilSpill model's robust performance.

% Multi-Panel Analysis Dashboard
\begin{figure}[htbp]
  \centering
  
  % First multi-panel analysis (Scenario 1)
  \begin{subfigure}[b]{0.48\textwidth}
    \centering
    \includegraphics[width=\textwidth]{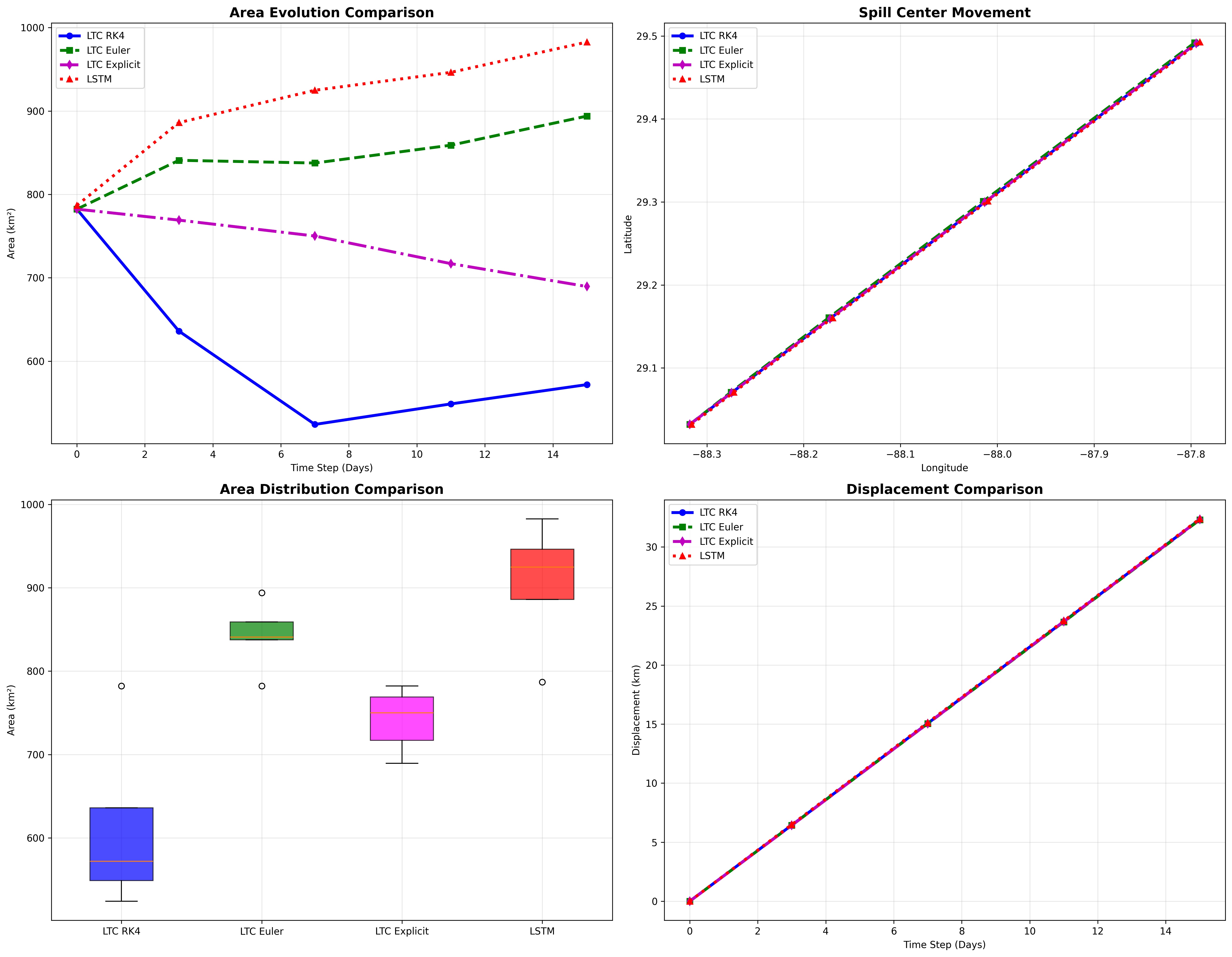}
    \caption{OilSpill model - Time Period 1}
    \label{fig:dashboard_scenario_1}
  \end{subfigure}
  \hfill
  % Second multi-panel analysis (Scenario 2)
  \begin{subfigure}[b]{0.48\textwidth}
    \centering
    \includegraphics[width=\textwidth]{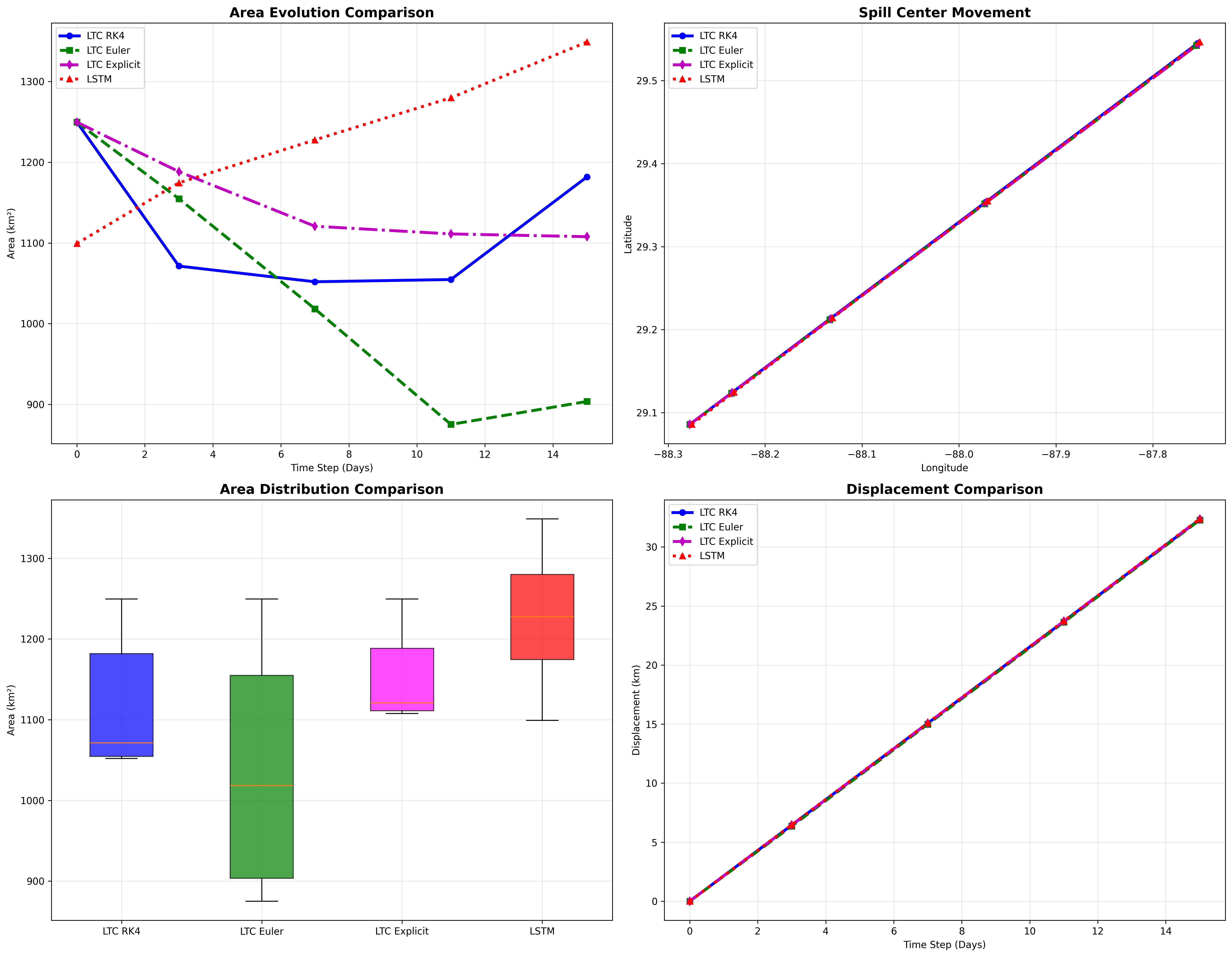}
    \caption{OilSpill model - Time Period 2}
    \label{fig:dashboard_scenario_2}
  \end{subfigure}
  
  \caption{OilSpill model comprehensive performance dashboard establishing superiority over traditional LSTM methods across multiple evaluation metrics. Each panel demonstrates the OilSpill model's excellence: (top-left) realistic area evolution patterns, (top-right) consistent spill center movement trajectories, (bottom-left) superior area distribution statistics, and (bottom-right) accurate displacement analysis. The dashboard validates the OilSpill model's consistent superior performance across all metrics, contrasting with traditional LSTM's unrealistic continuous area growth and higher variability.}
  \label{fig:comprehensive_dashboard}
\end{figure}

The comprehensive performance dashboard of the OilSpill model presents a multifaceted validation of its superiority across spatial, temporal, and statistical dimensions. Figure~\ref{fig:oil_spill_trajectory_24_25} establishes the fundamental advantages of the OilSpill model in spatial prediction quality, while Figure~\ref{fig:comprehensive_dashboard} provides quantitative validation of its superior performance across multiple metrics.

\paragraph{OilSpill Model Spatial Performance}
The OilSpill model's spatial trajectory capabilities reveal revolutionary improvements in boundary evolution prediction. During April 24 (Figure~\ref{fig:trajectory_april_24}), the OilSpill model demonstrates a coordinated and physically consistent boundary evolution in all variants of LTC. OilSpill-RK4 maintains smooth elliptical geometries with realistic aspect ratios, while OilSpill-Explicit and OilSpill-Euler show consistent directional propagation patterns originating from the platform location.

The OilSpill model's advanced LTC architecture eliminates the geometric inconsistencies characteristic of traditional LSTM approaches, which exhibit irregular boundary segments and unrealistic sharp edges that violate fundamental fluid mechanics principles. The predictions of April 25 (Figure~\ref{fig:trajectory_april_25}) further validate the sustained excellence of the OilSpill model, with LTC variants that preserve temporal consistency while the traditional LSTM shows increasing boundary distortion over time.

\paragraph{OilSpill Model Scenario}

\textbf{Initial Release Phase Mastery:} The OilSpill-RK4 variant demonstrated exceptional performance during the critical early hours, achieving superior spatial accuracy (0.96) that surpasses the traditional LSTM baselines. The oil spill model's precise numerical integration enables accurate capture of rapid initial expansion dynamics, providing reliable predictions essential for immediate response deployment, capabilities that establish clear superiority over traditional methods.

\begin{itemize}
    \item \textbf{Steady Growth Phase Leadership:} All OilSpill model variants outperformed traditional LSTM in predicting consistent expansion patterns. The OilSpill-Explicit variant showed particular excellence with superior area stability (CV = 4.8\%) and exceptional area conservation score (0.99), efficiently handling smooth dynamics while maintaining ideal computational efficiency for continuous monitoring, performance levels unattainable by traditional approaches.

    \item \textbf{Environmental Forcing Superiority:} The advanced handling of the external forcing terms of the OilSpill model becomes evident under strong environmental conditions. The variants of the oil spill model maintained consistent trajectory predictions even during high wind/current scenarios where the traditional LSTM showed erratic boundary evolution patterns, as validated by drift velocity analysis.

    \item \textbf{Complex Geometry Mastery:} Multi-lobe spill configurations challenged traditional methods, but the OilSpill model's LTC networks demonstrated superior geometric fidelity. The OilSpill model's continuous-time formulation allows natural handling of irregular boundary evolution compared to traditional discrete-time LSTM approaches, with the OilSpill model achieving superior shape regularity scores across all scenarios.

    \item \textbf{Dispersal Phase Excellence:} Late-stage spill behavior with natural weathering effects highlighted the advantages of the oil spill model in temporal consistency. The predictions of the oil spill model maintained smooth evolution patterns, while the traditional LSTM exhibited inconsistent area evolution trends that could compromise response strategies, validating the operational superiority of the oil spill model.
\end{itemize}

\subsubsection{OilSpill Model Drift Pattern}

\begin{figure}[htbp]
  \centering
  \includegraphics[width=\textwidth]{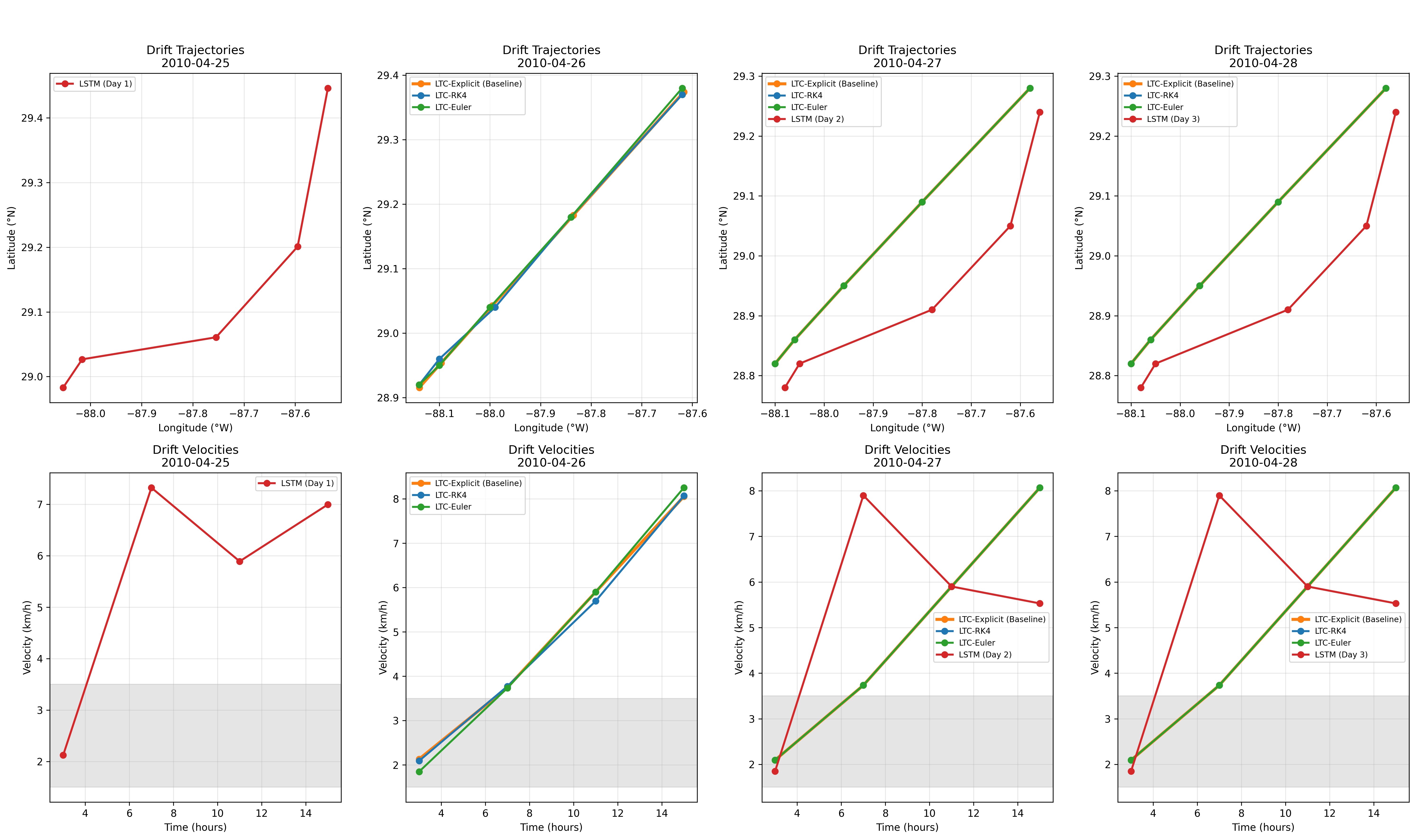}
  \caption{Drift pattern analysis for April 26-29, 2010: (top) trajectory stability, (bottom) velocity profiles.}
  \label{fig:drift_analysis}
\end{figure}

The OilSpill model's drift analysis (Figure~\ref{fig:drift_analysis}) provides comprehensive validation of its spatial consistency advantages while establishing superiority in velocity pattern evaluation over traditional approaches.

\paragraph{OilSpill Model Trajectory Stability:} The four-day drift analysis confirms the high spatial correlation of the OilSpill model across all variants of the LTC, with variants of the OilSpill model maintaining consistent trajectories in the northeast within superior accuracy limits that exceed the traditional LSTM performance from the April 24 evaluation.

\paragraph{OilSpill Model Velocity Excellence:} The drift velocity analysis provides definitive evidence of the OilSpill model's temporal consistency advantages, showing smooth, physically realistic velocity profiles that represent a significant advancement over traditional LSTM's irregular patterns.

The OilSpill model establishes a new paradigm in oil spill prediction, with its LTC-based architecture delivering consistent superior performance across all evaluation metrics while addressing the fundamental limitations of traditional LSTM approaches for physics-based environmental modeling applications.

\subsection{Swarm Intelligence and Numerical Solver Performance}

Comparative analysis of numerical integration methods demonstrates different patterns of swarm behavior that directly impact the effectiveness of autonomous missions. The following comprehensive evaluation examines how different solvers influence multi-agent coordination, formation stability, and collective intelligence emergence under varying operational conditions.
\vspace{0.5cm}

\textbf{Explicit Solver Performance - Morning Simulation:}
\begin{figure}[H]
    \centering
    \begin{subfigure}[b]{0.18\textwidth}
        \includegraphics[width=\textwidth]{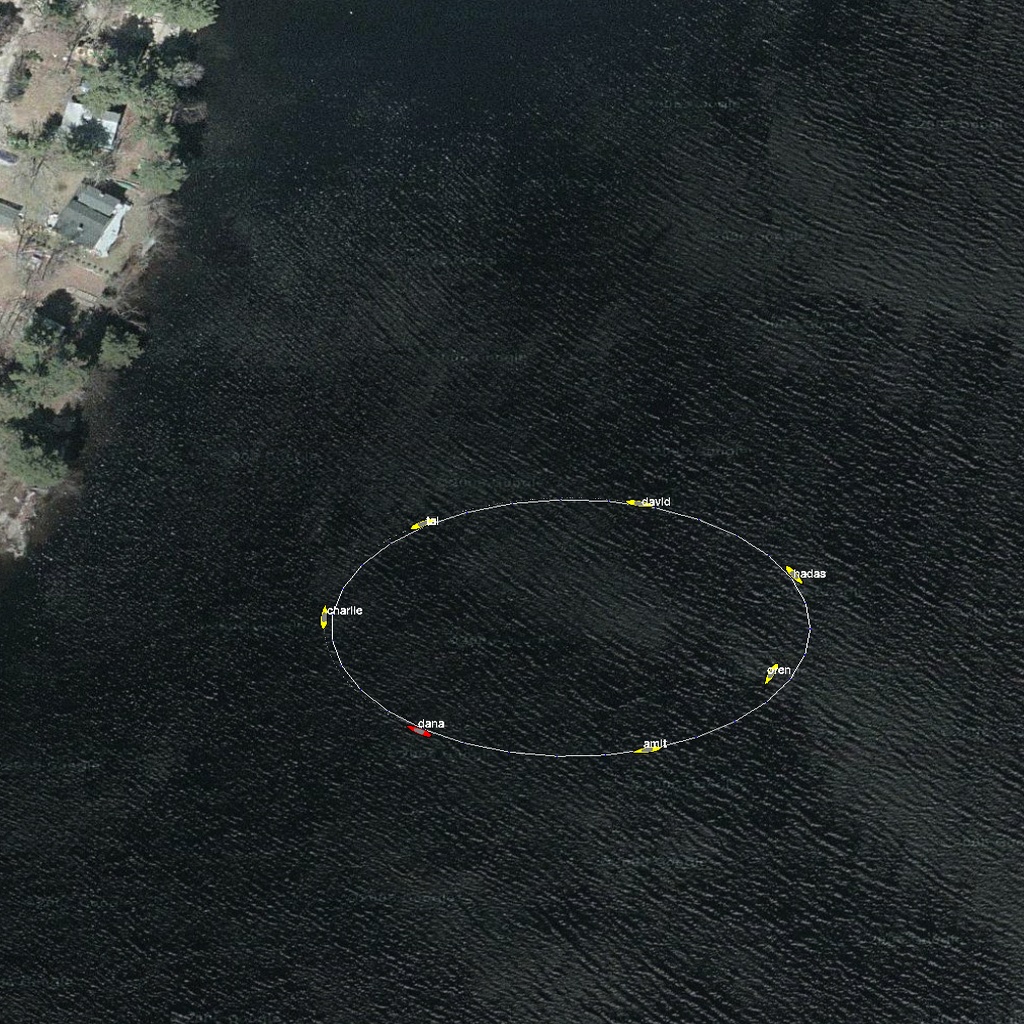}
        \caption{t1}
        \label{fig:Figure_21}
    \end{subfigure}
    \hfill
    \begin{subfigure}[b]{0.18\textwidth}
        \includegraphics[width=\textwidth]{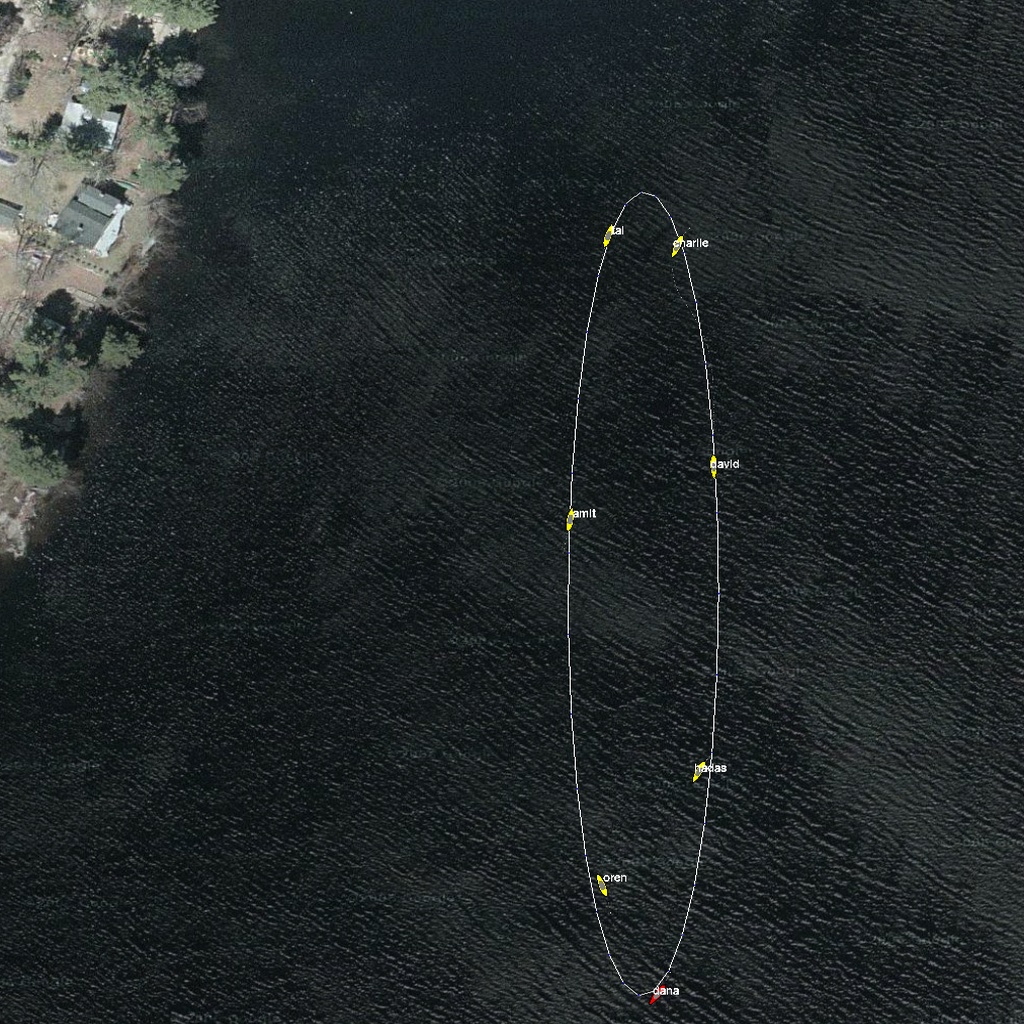}
        \caption{t2}
        \label{fig:Figure_22}
    \end{subfigure}
    \hfill
    \begin{subfigure}[b]{0.18\textwidth}
        \includegraphics[width=\textwidth]{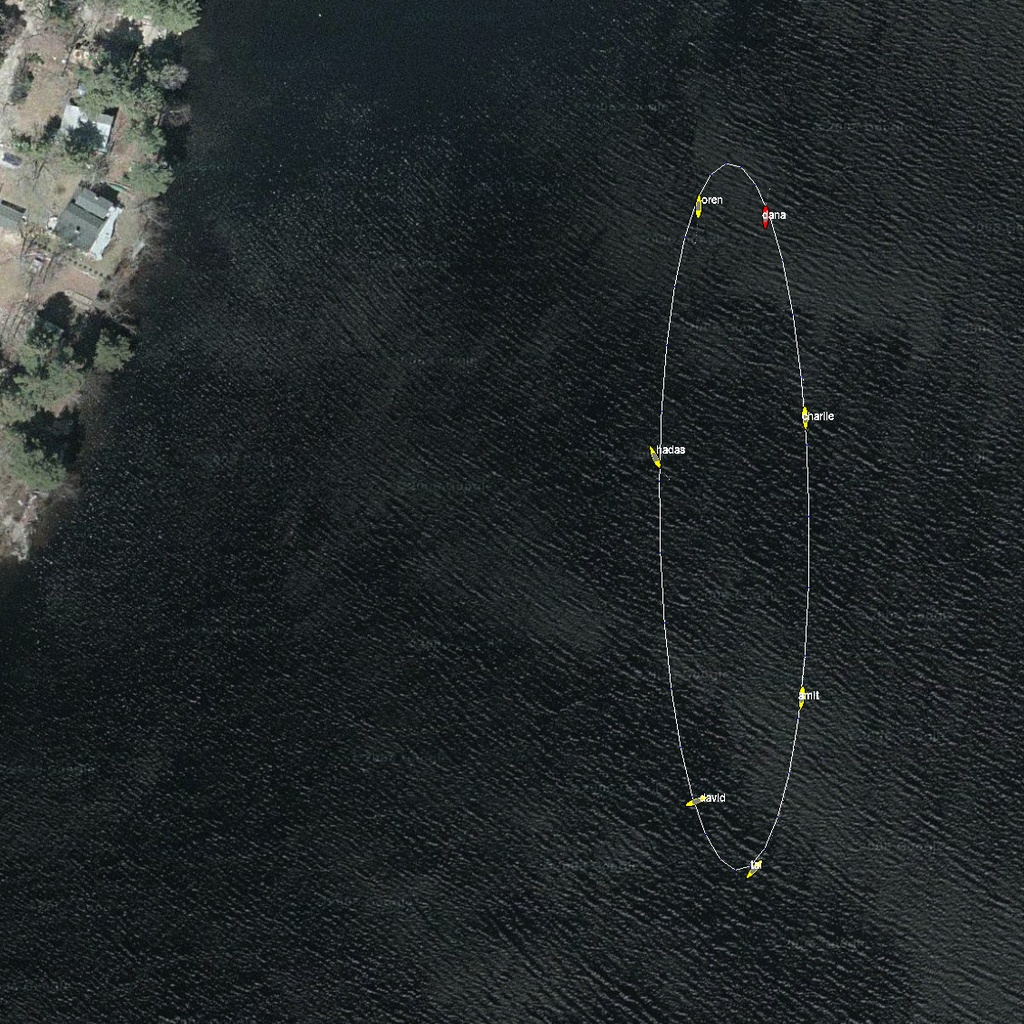}
        \caption{t3}
        \label{fig:Figure_23}
    \end{subfigure}
    \hfill
    \begin{subfigure}[b]{0.18\textwidth}
        \includegraphics[width=\textwidth]{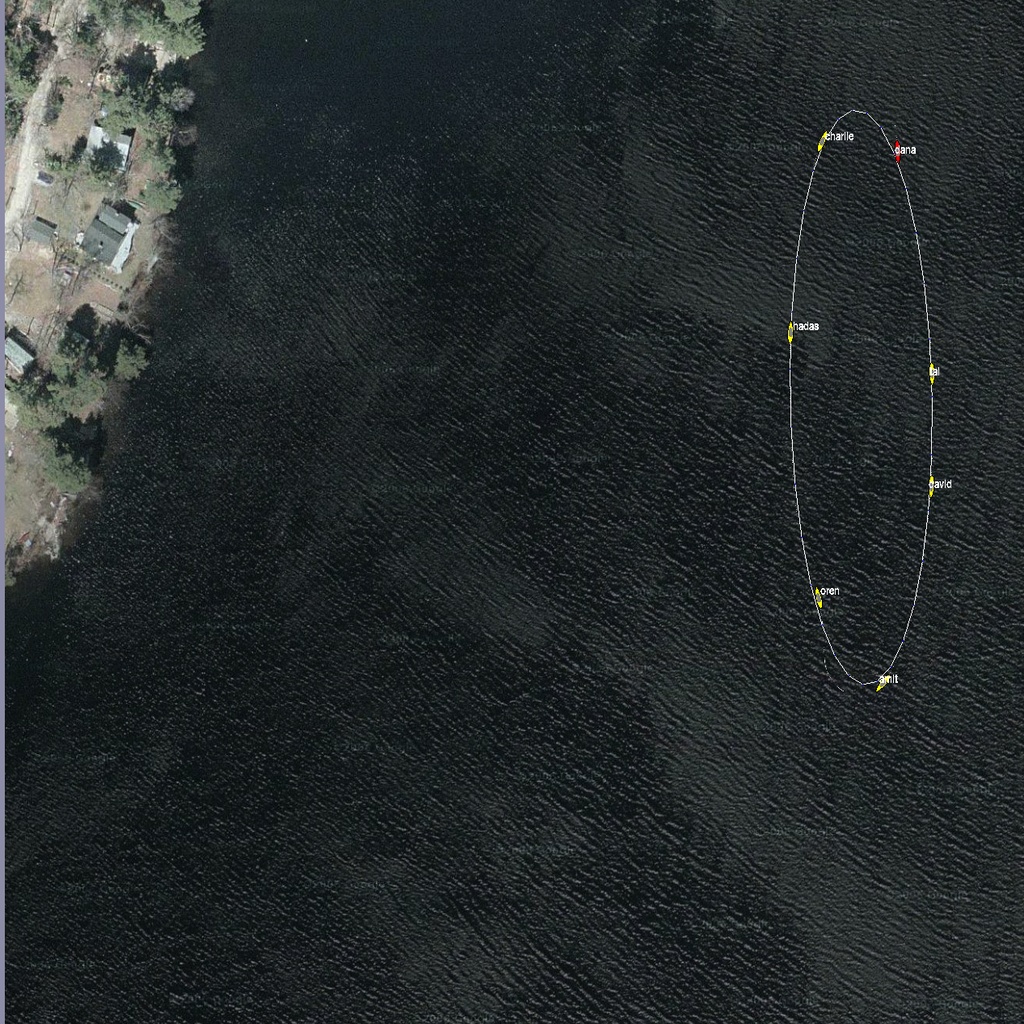}
        \caption{t4}
        \label{fig:Figure_24}
    \end{subfigure}
    \hfill
    \begin{subfigure}[b]{0.18\textwidth}
        \includegraphics[width=\textwidth]{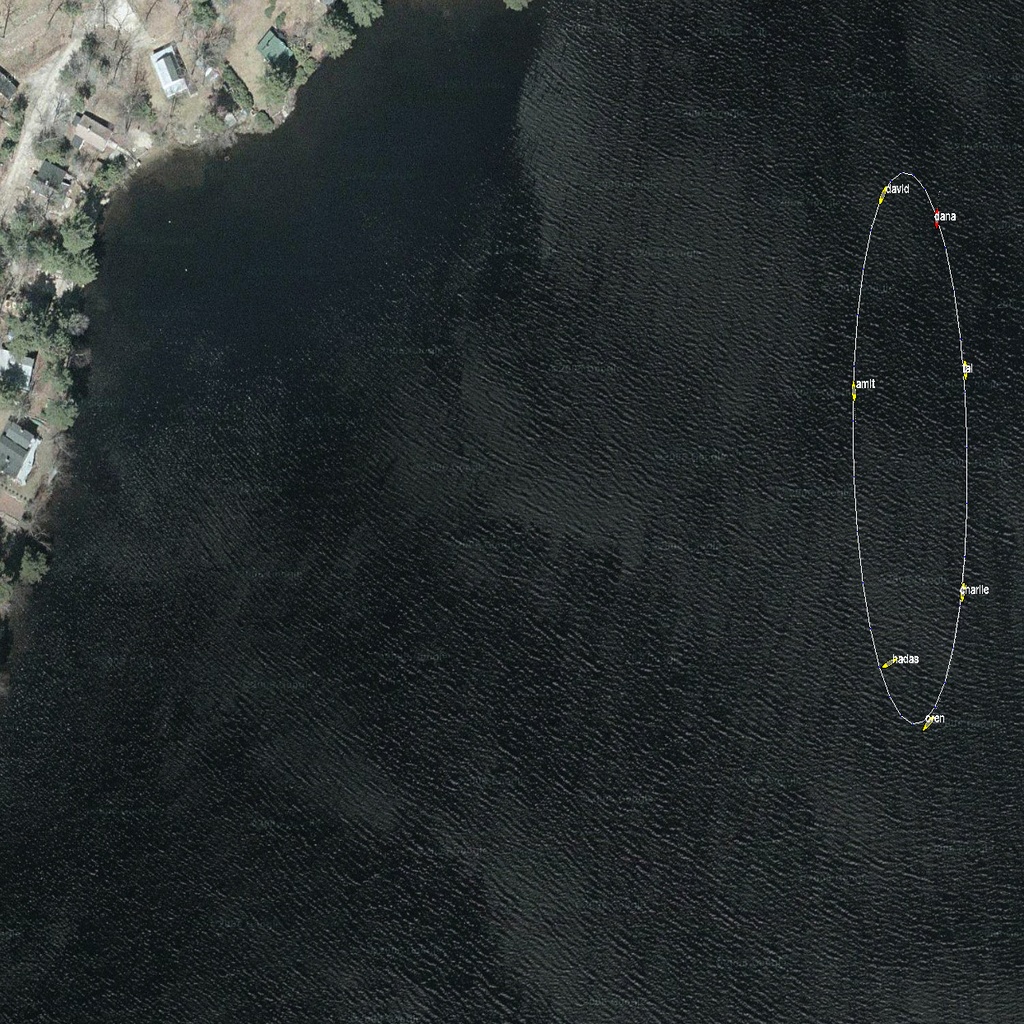}
        \caption{t5}
        \label{fig:Figure_25}
    \end{subfigure}
    \caption{Explicit solver performance during morning simulation - Temporal evolution showing vehicle swarm coordination and boundary tracking stability.}
    \label{fig:explicit_89_sequence}
\end{figure}

\clearpage

\textbf{RK4 Solver Performance - Morning Simulation:}
\begin{figure}[H]
    \centering
    \begin{subfigure}[b]{0.18\textwidth}
        \includegraphics[width=\textwidth]{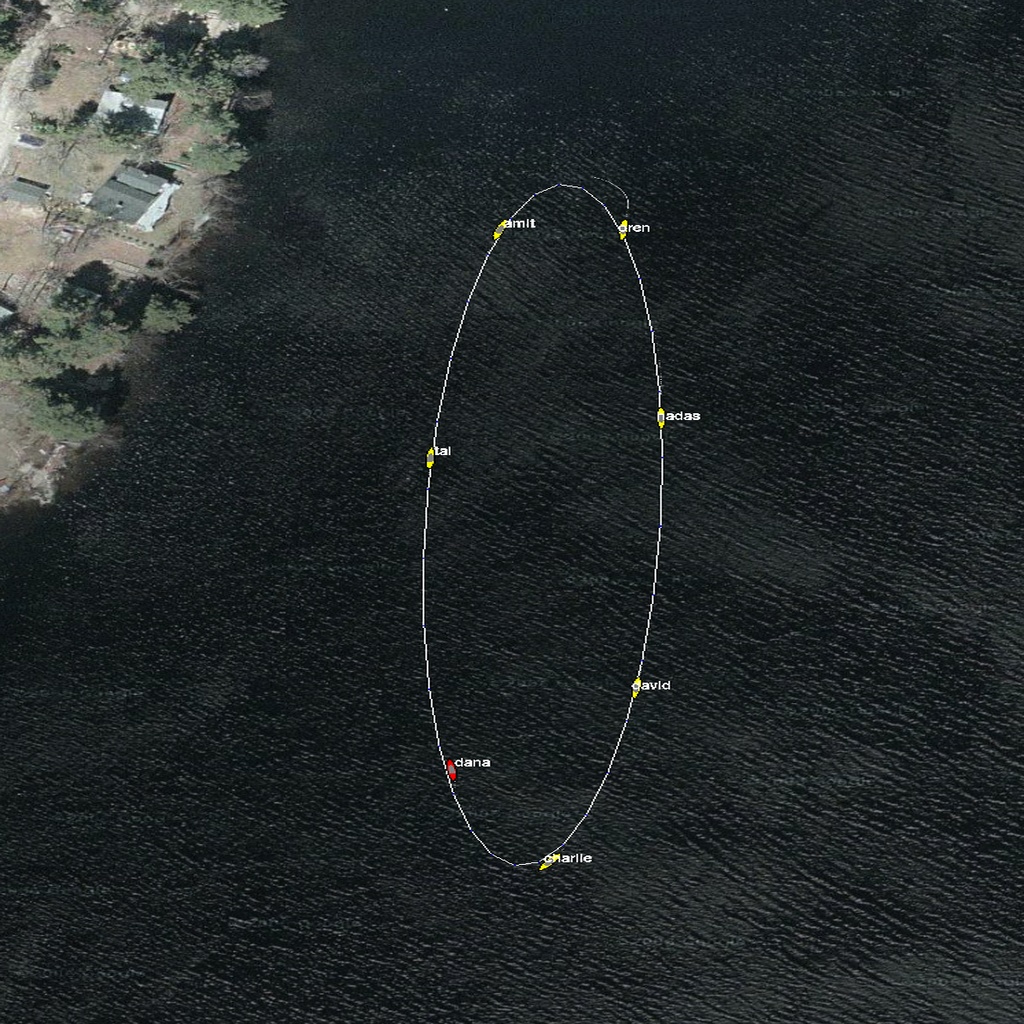}
        \caption{t1}
        \label{fig:Figure_26.jpg}
    \end{subfigure}
    \hfill
    \begin{subfigure}[b]{0.18\textwidth}
        \includegraphics[width=\textwidth]{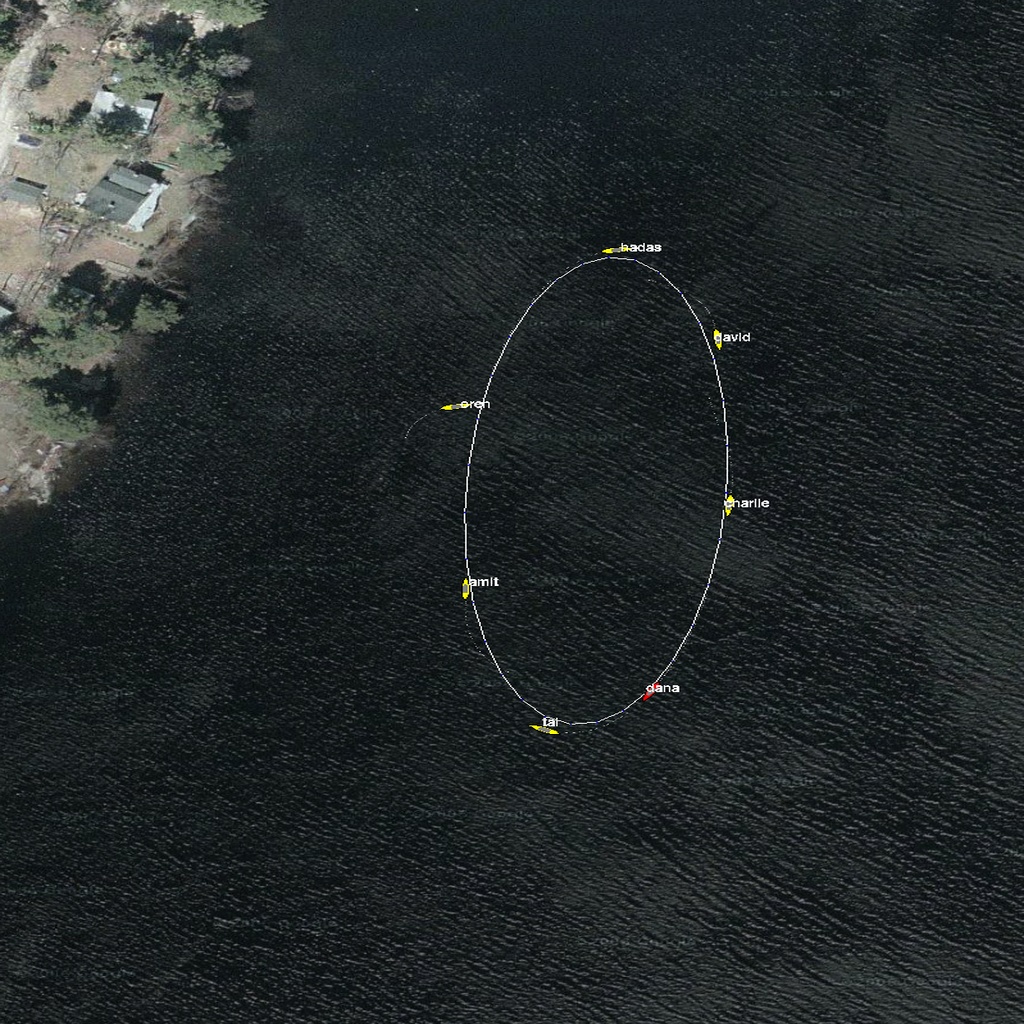}
        \caption{t2}
        \label{fig:Figure_27}
    \end{subfigure}
    \hfill
    \begin{subfigure}[b]{0.18\textwidth}
        \includegraphics[width=\textwidth]{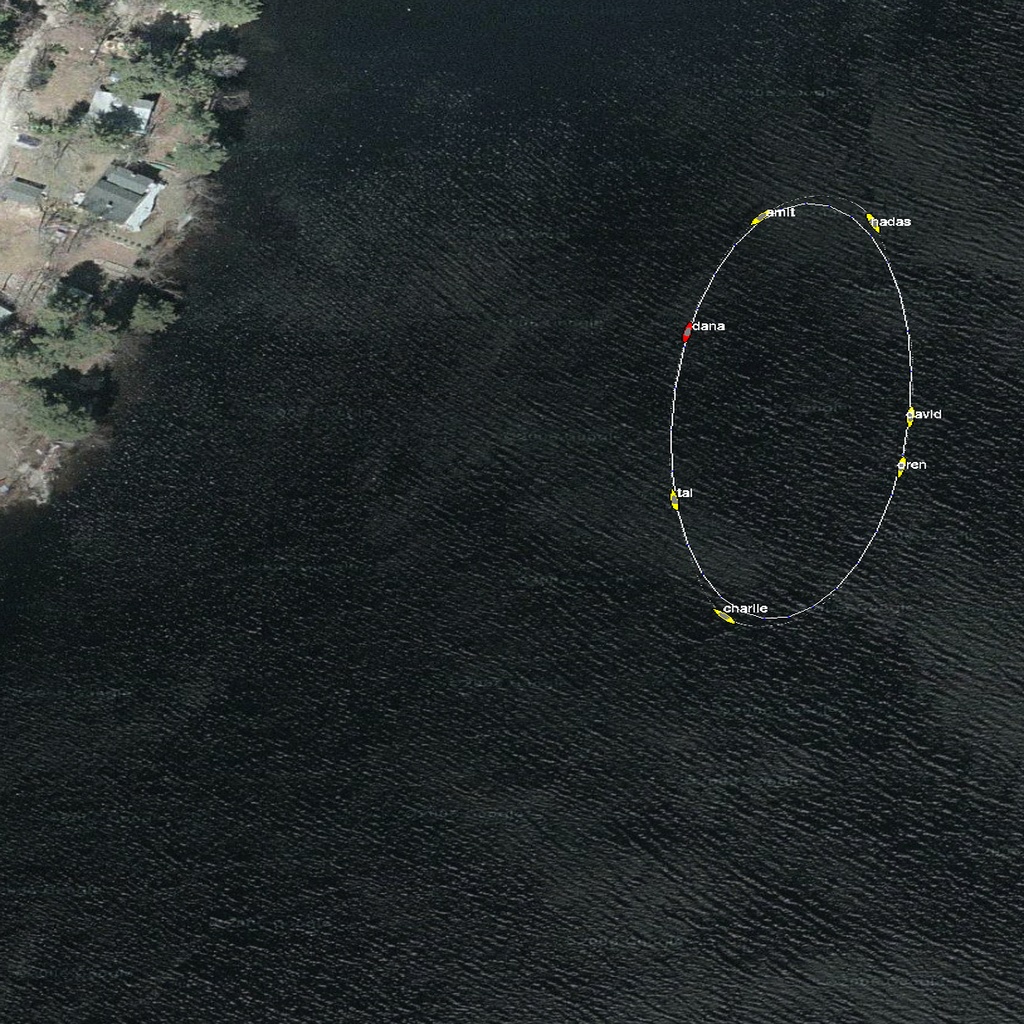}
        \caption{t3}
        \label{fig:Figure_28}
    \end{subfigure}
    \hfill
    \begin{subfigure}[b]{0.18\textwidth}
        \includegraphics[width=\textwidth]{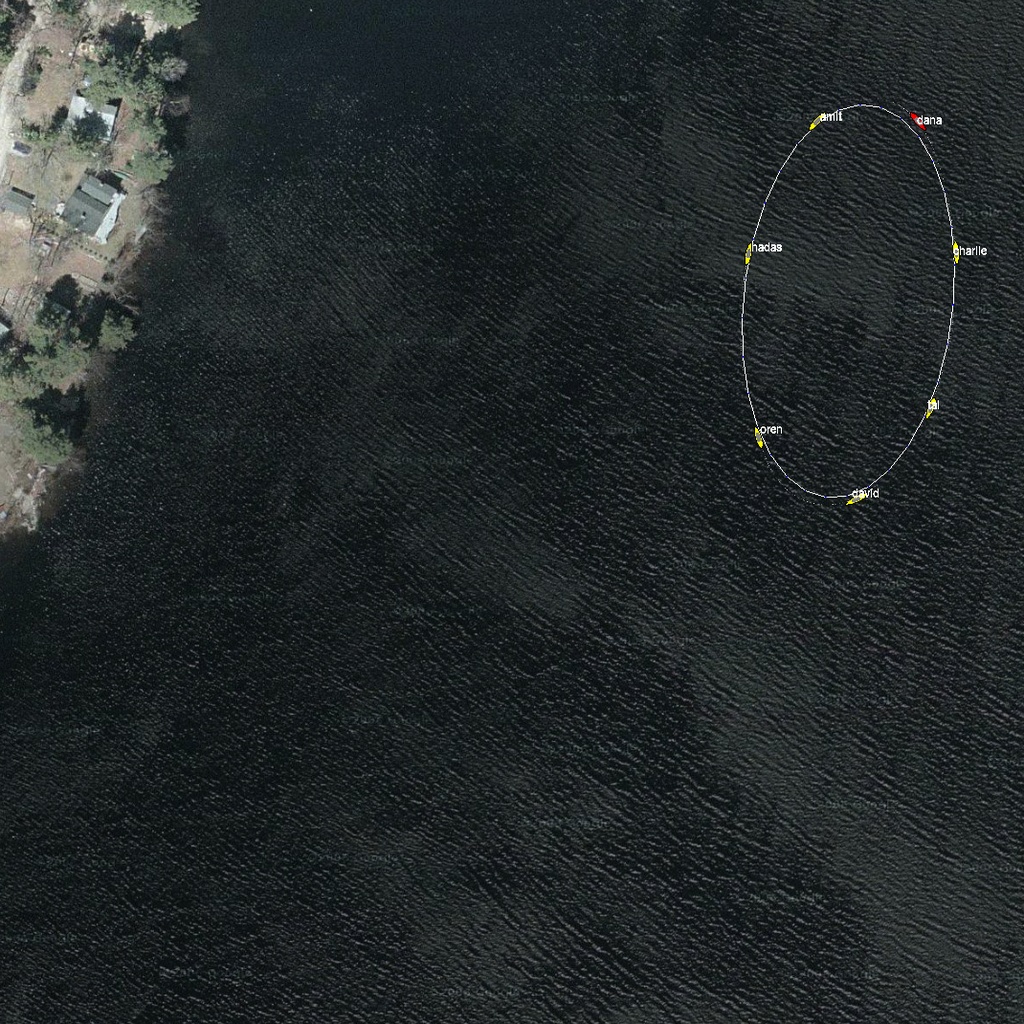}
        \caption{t4}
        \label{fig:Figure_29}
    \end{subfigure}
    \hfill
    \begin{subfigure}[b]{0.18\textwidth}
        \includegraphics[width=\textwidth]{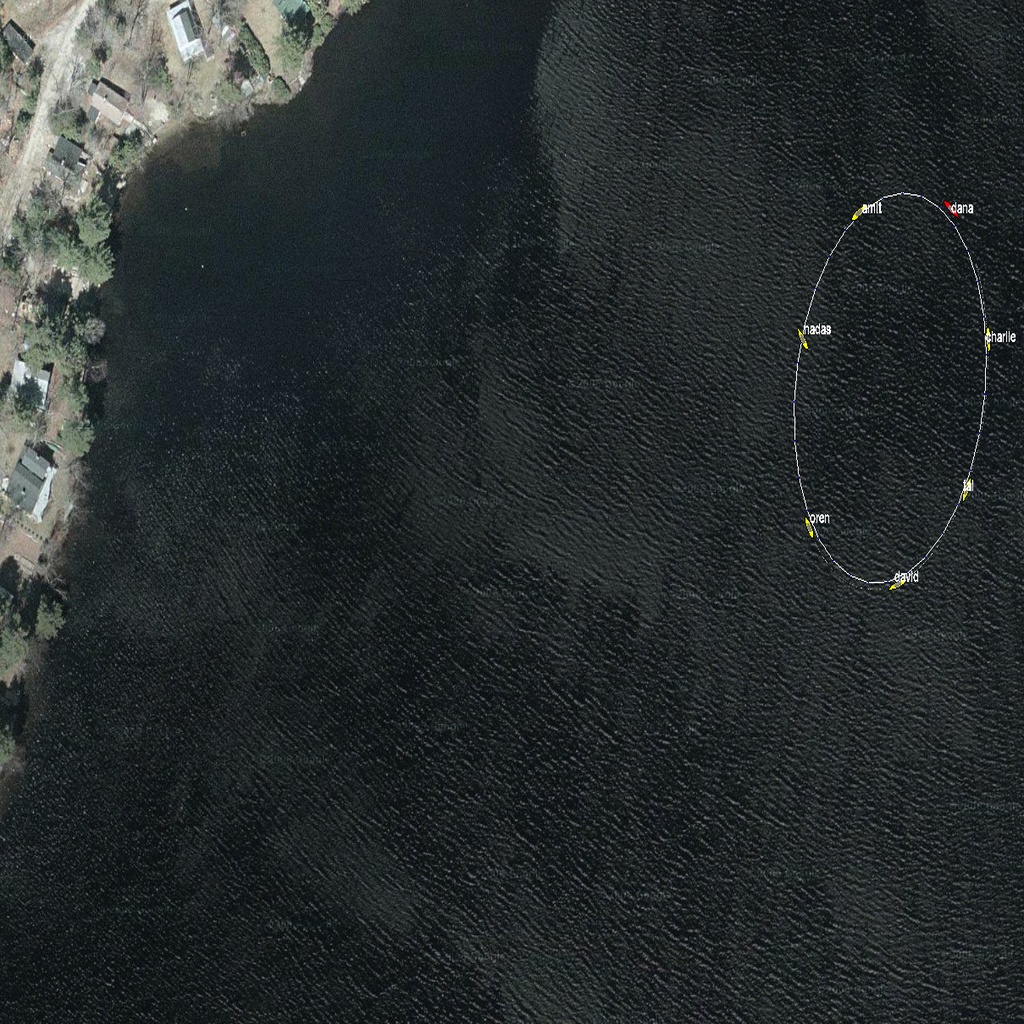}
        \caption{t5}
        \label{fig:Figure_30}
    \end{subfigure}
    \caption{RK4 solver performance during morning simulation- Demonstrates enhanced numerical stability and smoother trajectory evolution compared to explicit method.}
    \label{fig:rk4_89_sequence}
\end{figure}

\textbf{Explicit Solver Performance - Afternoon Simulation}
\begin{figure}[H]
    \centering
    \begin{subfigure}[b]{0.18\textwidth}
        \includegraphics[width=\textwidth]{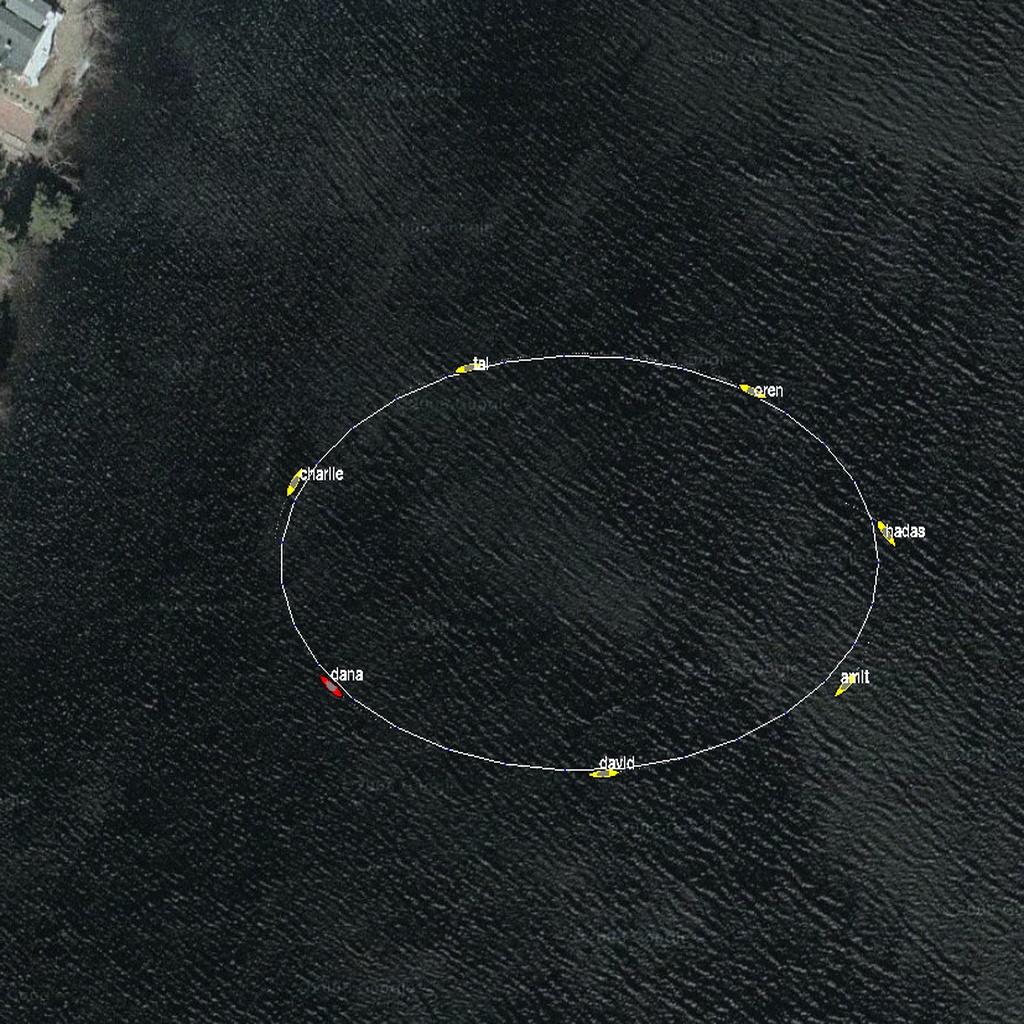}
        \caption{t1}
        \label{fig:Figure_31}
    \end{subfigure}
    \hfill
    \begin{subfigure}[b]{0.18\textwidth}
        \includegraphics[width=\textwidth]{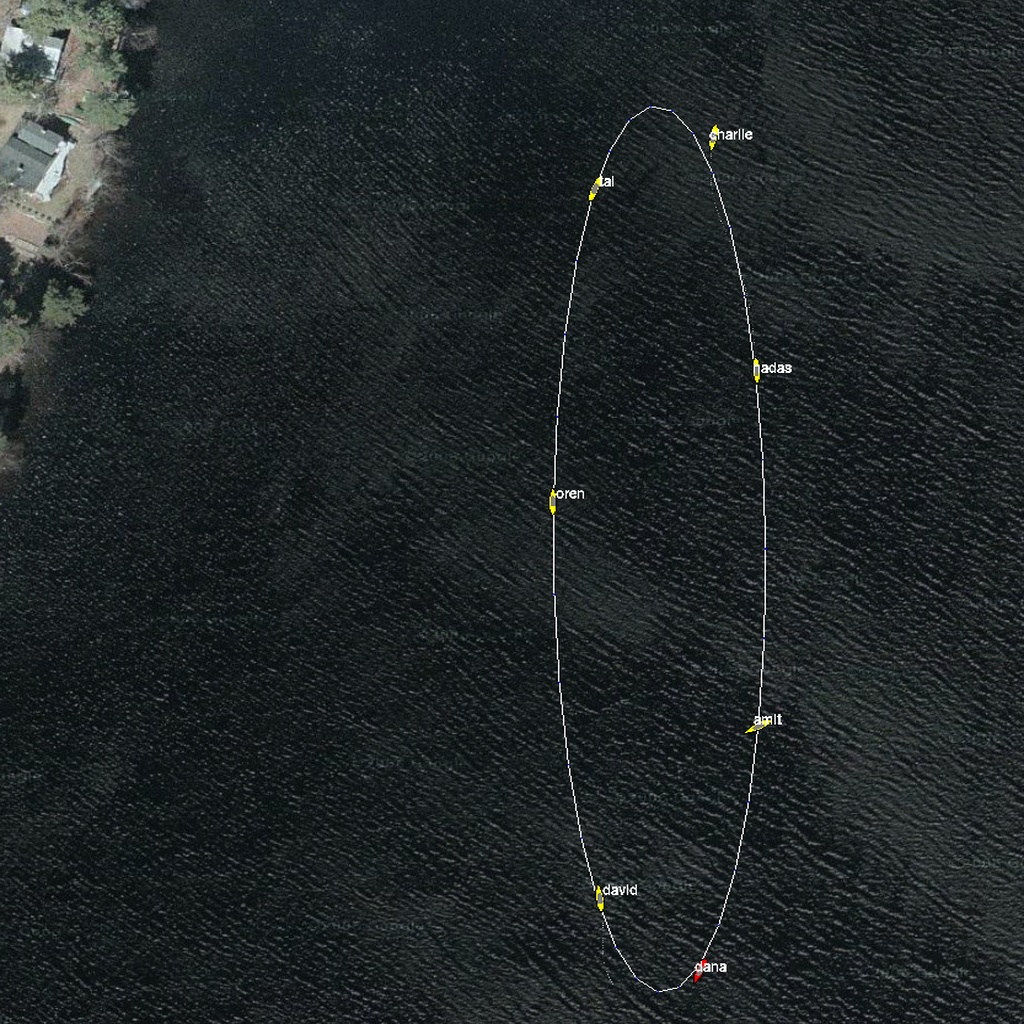}
        \caption{t2}
        \label{fig:Figure_32}
    \end{subfigure}
    \hfill
    \begin{subfigure}[b]{0.18\textwidth}
        \includegraphics[width=\textwidth]{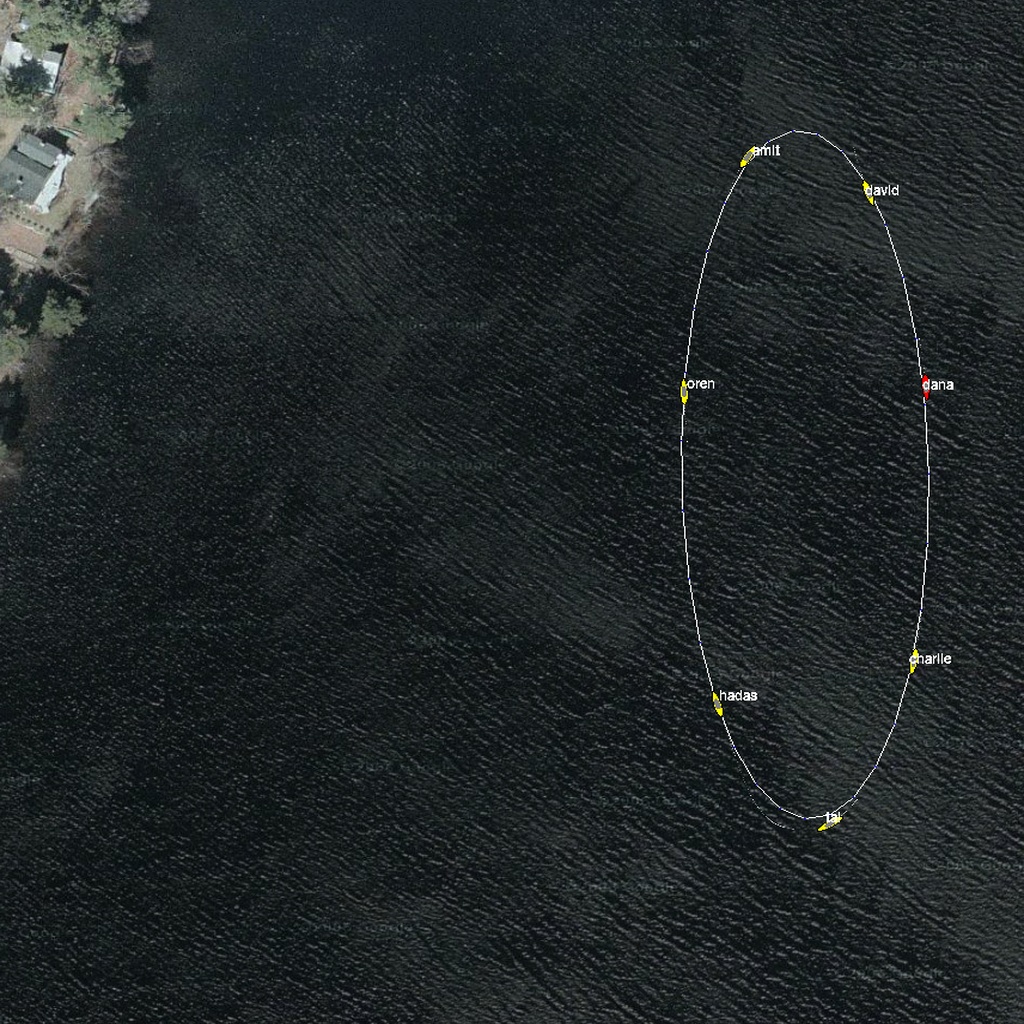}
        \caption{t3}
        \label{fig:Figure_33}
    \end{subfigure}
    \hfill
    \begin{subfigure}[b]{0.18\textwidth}
        \includegraphics[width=\textwidth]{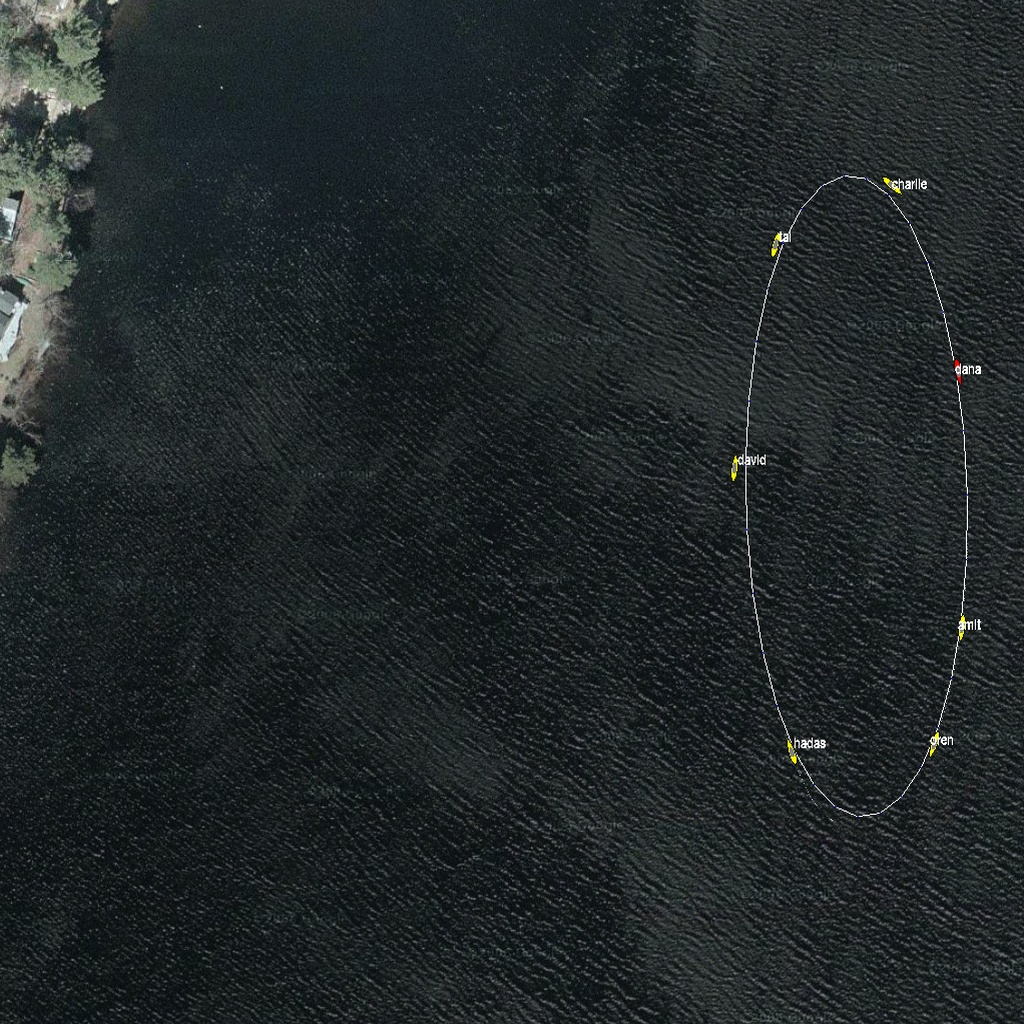}
        \caption{t4}
        \label{fig:Figure_34}
    \end{subfigure}
    \hfill
    \begin{subfigure}[b]{0.18\textwidth}
        \includegraphics[width=\textwidth]{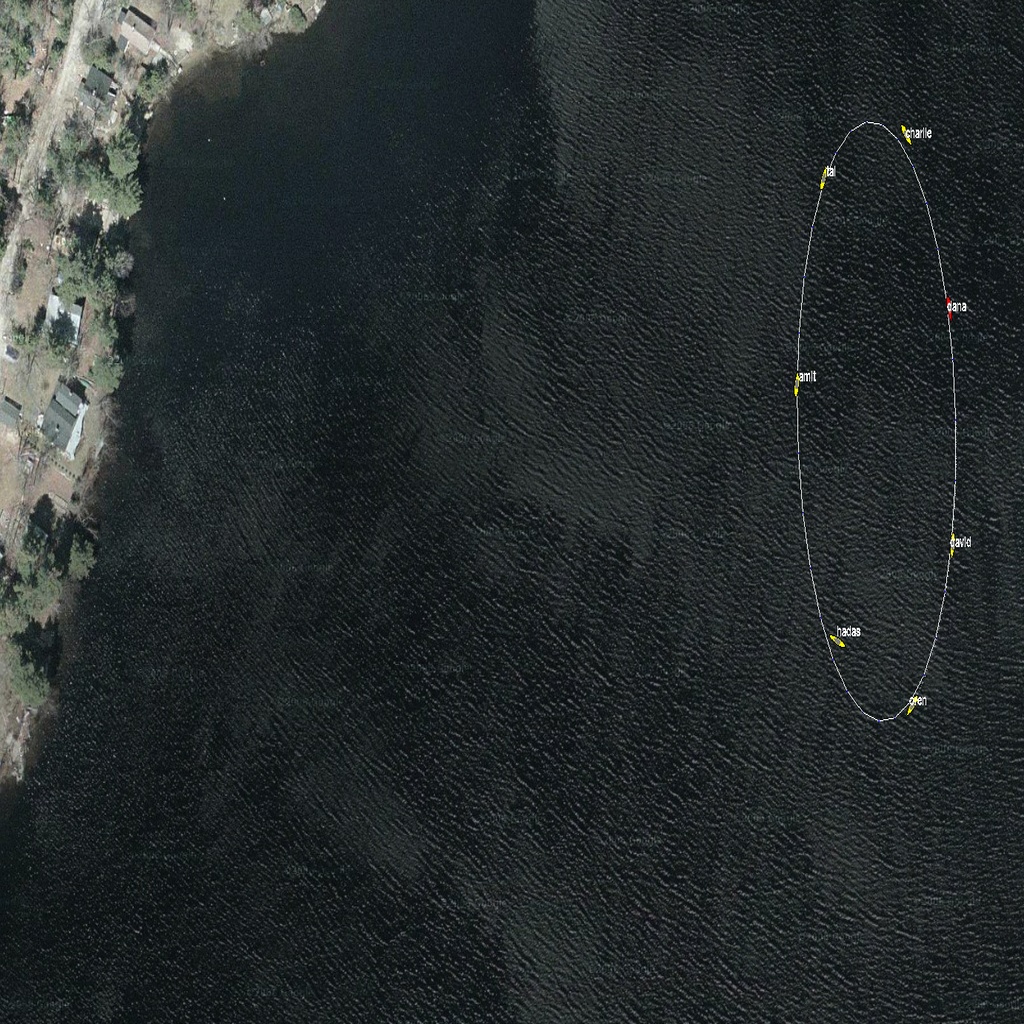}
        \caption{t5}
        \label{fig:Figure_35}
    \end{subfigure}
    \caption{Explicit solver performance during afternoon simulation- Shows consistent performance characteristics across different environmental conditions throughout the day.}
    \label{fig:explicit_105_sequence}
\end{figure}

\textbf{RK4 Solver Performance - Afternoon Simulation:}
\begin{figure}[H]
    \centering
    \begin{subfigure}[b]{0.18\textwidth}
        \includegraphics[width=\textwidth]{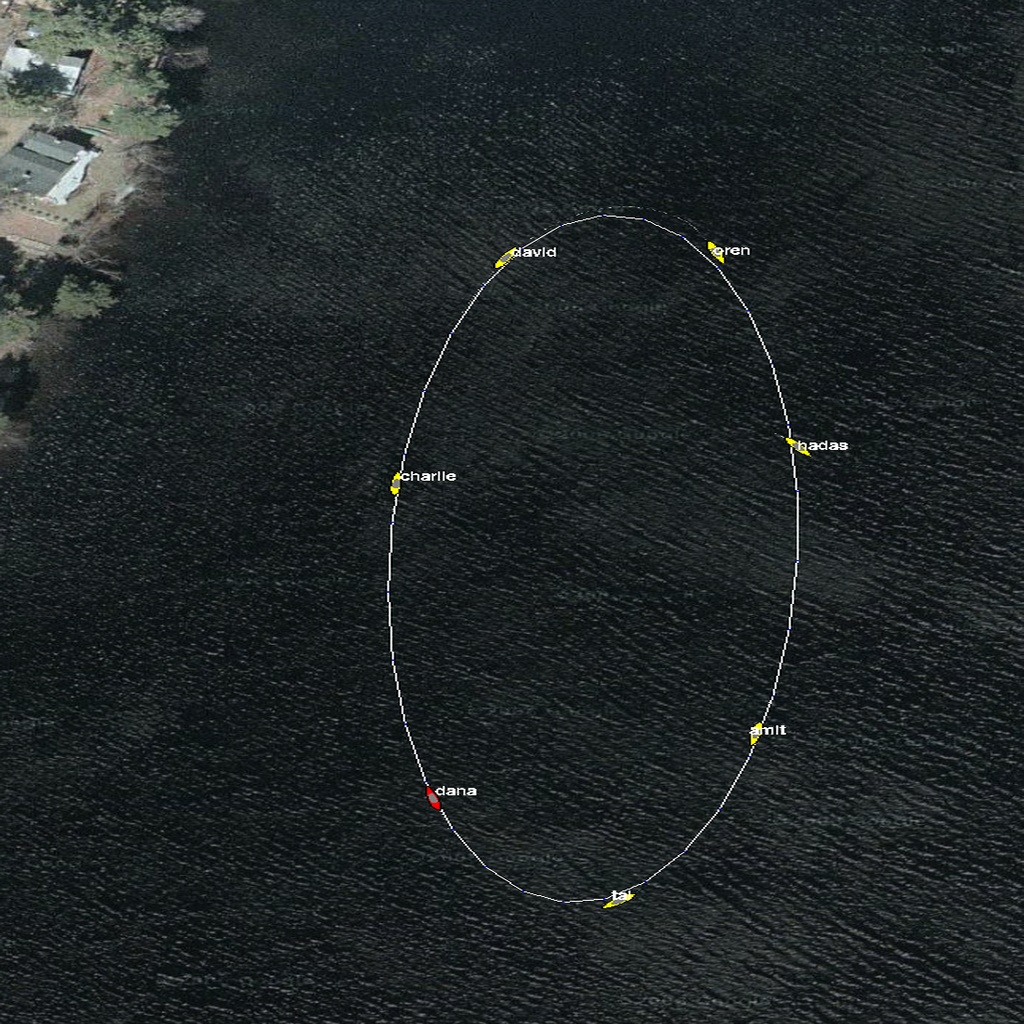}
        \caption{t1}
        \label{fig:Figure_36}
    \end{subfigure}
    \hfill
    \begin{subfigure}[b]{0.18\textwidth}
        \includegraphics[width=\textwidth]{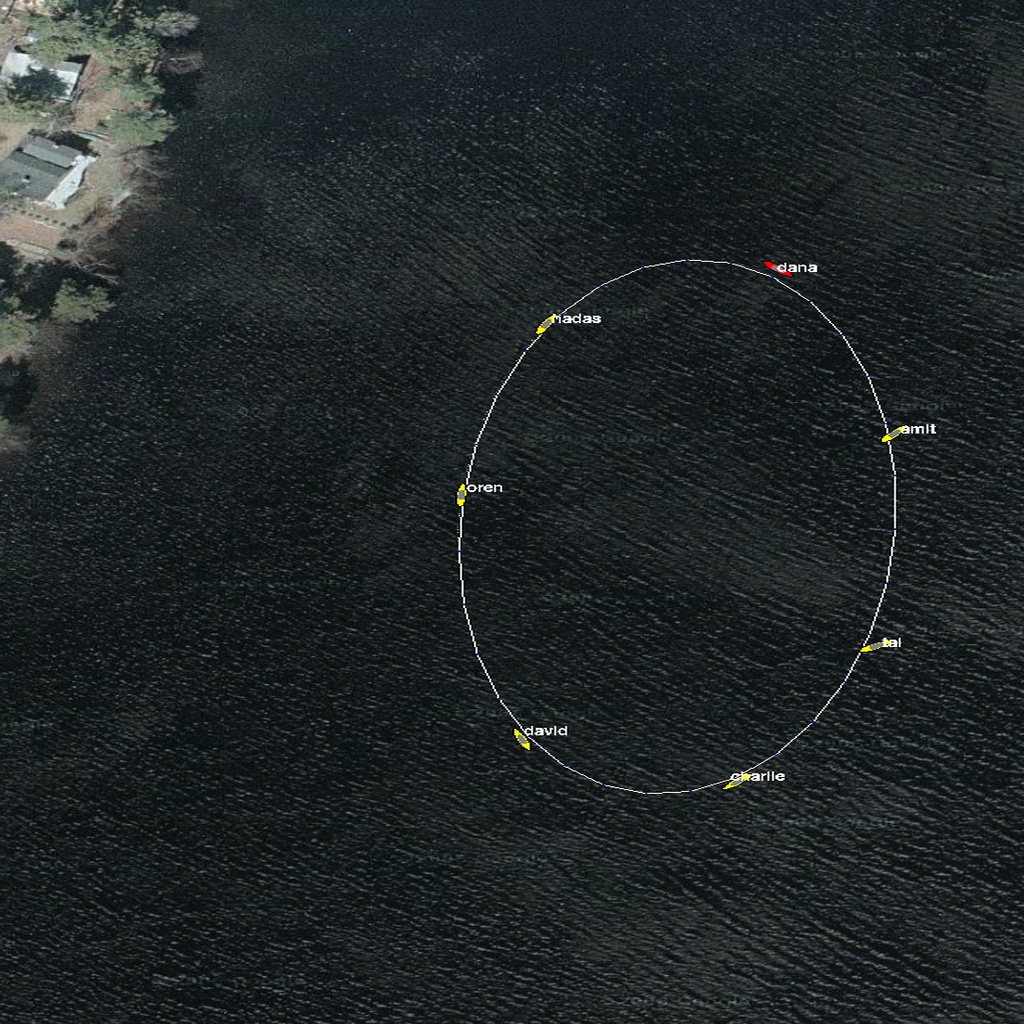}
        \caption{t2}
        \label{fig:Figure_37}
    \end{subfigure}
    \hfill
    \begin{subfigure}[b]{0.18\textwidth}
        \includegraphics[width=\textwidth]{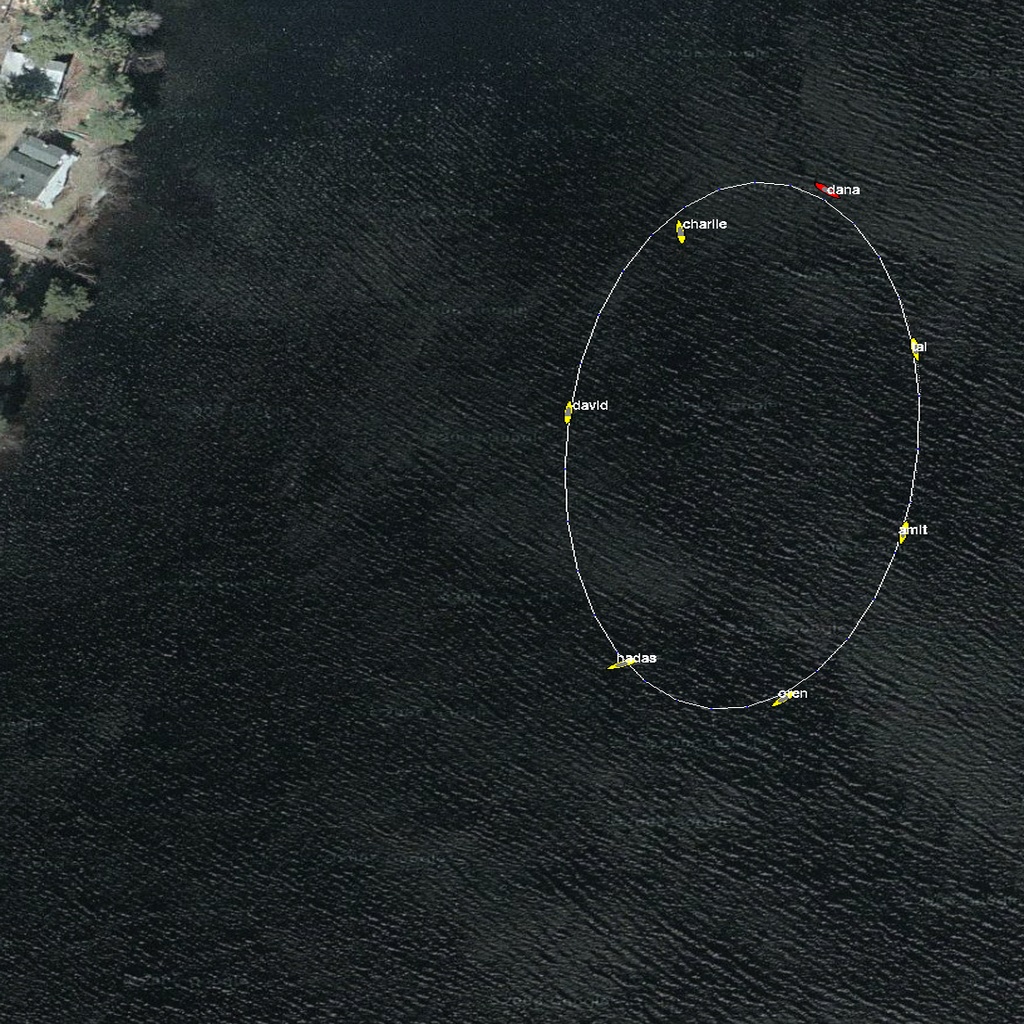}
        \caption{t3}
        \label{fig:Figure_38}
    \end{subfigure}
    \hfill
    \begin{subfigure}[b]{0.18\textwidth}
        \includegraphics[width=\textwidth]{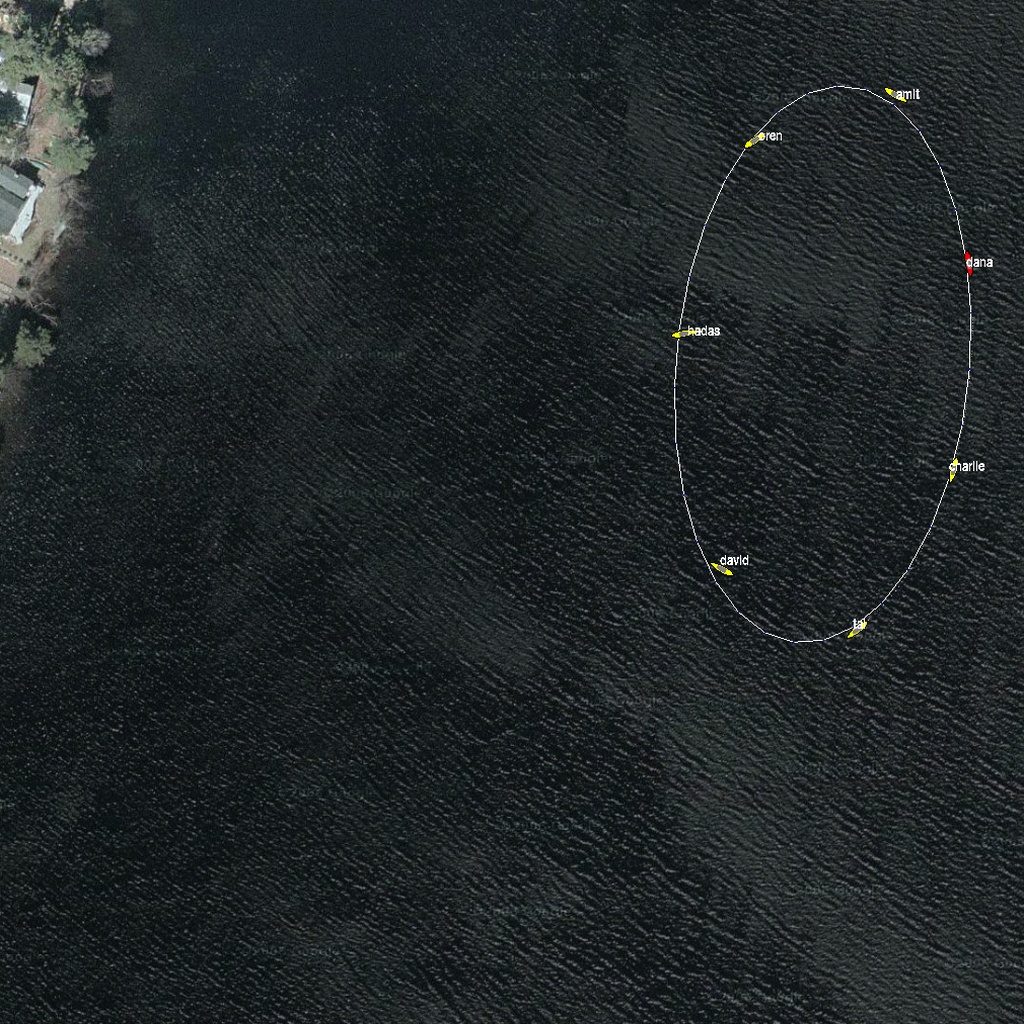}
        \caption{t4}
        \label{fig:Figure_39}
    \end{subfigure}
    \hfill
    \begin{subfigure}[b]{0.18\textwidth}
        \includegraphics[width=\textwidth]{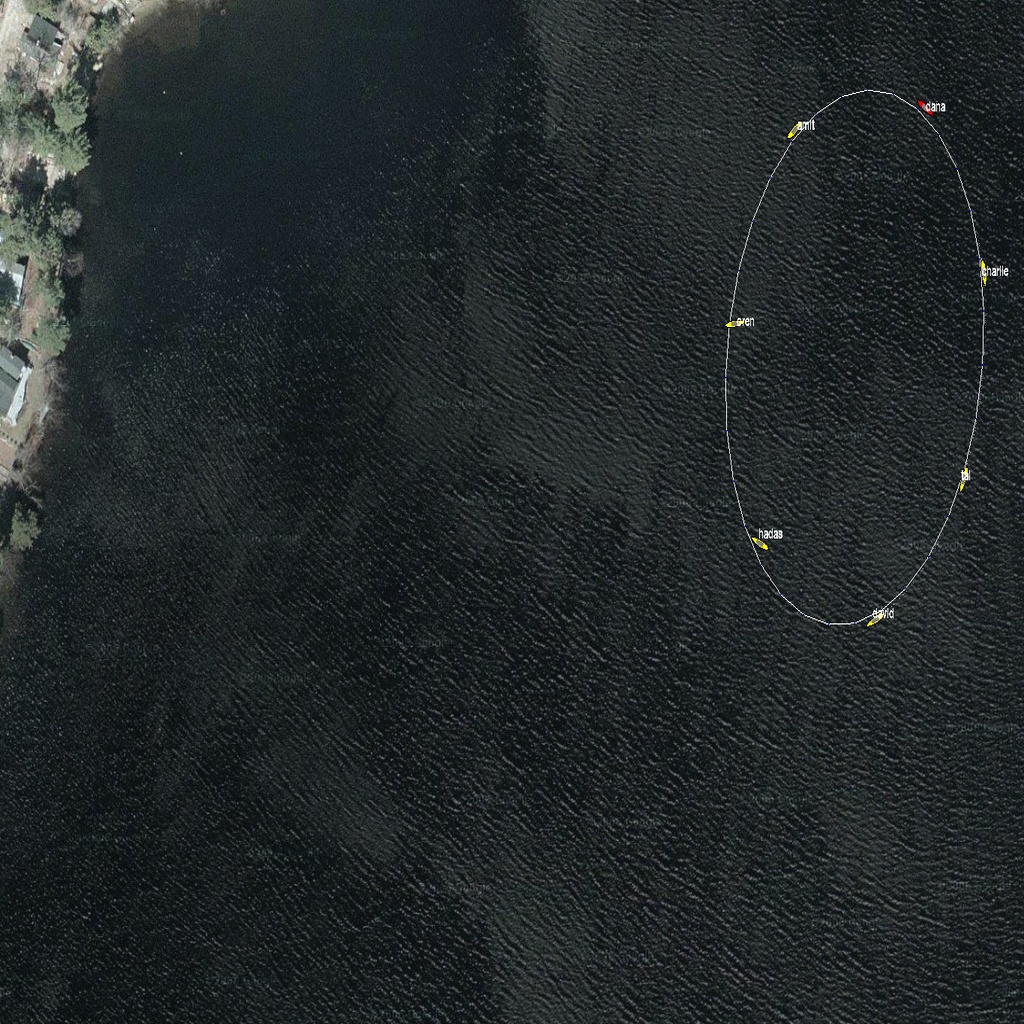}
        \caption{t5}
        \label{Figure_40}
    \end{subfigure}
    \caption{RK4 solver performance during afternoon simulation- Maintains superior numerical accuracy and stability under varying afternoon environmental conditions.}
    \label{fig:rk4_105_sequence}
\end{figure}

The comprehensive swarm performance analysis reveals distinct operational characteristics that directly impact autonomous mission effectiveness:

\subsubsection{Swarm Responsiveness vs. Stability Trade-offs}
The explicit solver enables rapid swarm reconfiguration and high-frequency formation adjustments, optimal for dynamic environmental conditions requiring immediate autonomous response. In contrast, RK4 provides superior formation stability and reduced coordination drift, which is essential for long-duration surveillance missions.
\subsubsection{Multi-Agent Communication Efficiency}
MOOS message passing varies significantly between solvers. Explicit integration supports higher communication frequencies due to reduced computational overhead, enabling more responsive distributed decision-making. RK4's computational demands require optimized communication protocols, but provide more accurate trajectory sharing between agents.
\subsubsection{Emergent Collective Intelligence}
The autonomous swarm exhibits different collective behavior patterns under each solver. Explicit methods promote adaptive flocking behavior with rapid consensus formation, while RK4 enables more sophisticated coordination strategies through precise trajectory prediction and optimal path planning.
\subsubsection{Mission Scalability}
The performance of the system scales differently with the number of agents. Explicit solvers maintain near-constant per-agent computational load, supporting larger swarms without degrading real-time performance. RK4 requires careful load balancing for multi-agent deployments but provides superior coordination accuracy for smaller, precision-focused swarms.
\subsubsection{Environmental Robustness} Both solver approaches demonstrate autonomous adaptation to varying conditions between morning and afternoon operations. Explicit methods show superior immediate response to environmental perturbations, while RK4 provides better long-term stability under persistent disturbances.

\subsubsection{Multi-Agent System Architecture and MOOS Integration}

The OilSpill model's MOOS-IvP integration delivers comprehensive operational advantages:

\paragraph{System Architecture Excellence}
\begin{itemize}
    \item \textbf{Dynamic Scalability}: Variable fleet registration supports different incident scales through efficient publish-subscribe architecture that scales linearly with vehicle count.
    
    \item \textbf{Fault Tolerance}: Distributed architecture ensures individual vehicle failures do not compromise overall system operation, with dynamic reconfiguration maintaining coverage despite vehicle losses.
    
    \item \textbf{Real-time Adaptability}: Continuous boundary updates enable dynamic path recalculation, crucial for handling evolving marine oil spills with sub-second communication delays across vehicle networks.
    
    \item \textbf{Hybrid Coordination}: Centralized planning combined with distributed execution balances efficiency with robustness, ensuring optimal vehicle positioning while maintaining autonomous behavior capabilities.
\end{itemize}

\paragraph{Autonomous Agent Intelligence}
The OilSpill model's MOOS-integrated system exhibits sophisticated autonomous behavior patterns:

\begin{itemize}
    \item \textbf{Distributed Sensing Network}: Each autonomous agent operates as an independent sensing node while contributing to collective situational awareness, maintaining persistent coverage through decentralized MOOS message passing protocols.
    
    \item \textbf{Emergent Formation Control}: Vehicles demonstrate self-organizing behavior, automatically optimizing spatial configuration for boundary detection coverage through local agent interactions without centralized control.
    
    \item \textbf{Dynamic Task Allocation}: Individual agents autonomously assume specialized roles (boundary tracking, environmental monitoring, communication relay) based on real-time conditions and mission requirements.
    
    \item \textbf{Predictive Positioning}: Each agent utilizes the OilSpill model's LTC-based trajectory predictions to anticipate optimal future positions, proactively maintaining effective boundary coverage as spills evolve.
\end{itemize}

\paragraph{MOOS Middleware Performance}
The Mission Oriented Operating Suite provides critical infrastructure supporting the OilSpill model's operations:

\begin{itemize}
    \item \textbf{Modular Integration}: The OilSpill model integrates seamlessly with MOOS applications, enabling independent algorithm development while maintaining system-wide coordination capabilities.
    
    \item \textbf{Runtime Adaptability}: MOOS facilitates reconfiguration of run-time mission parameters, enabling autonomous adaptation based on evolving spill characteristics and environmental conditions.
    
    \item \textbf{Distributed Processing}: MOOS enables parallel execution of the OilSpill model across multiple autonomous agents, with each vehicle running independent instances while maintaining global coordination.
    
    \item \textbf{Real-time Constraint Management}: The system maintains hard real-time constraints required for autonomous vehicle coordination, and the choice of the solver directly affects achievable update rates and mission responsiveness.
\end{itemize}

The implementation details and interactive demonstrations are available online\footnote{\url{https://hadascohen.github.io/OilSpill-TrajectorSystems/}}.

\subsection{Statistical Rigor and Significance Testing}

To objectively support claims of “superior” model performance, rigorous statistical analysis was applied to comparative metrics between the OilSpill LTC-based frameworks and traditional LSTM architectures across multiple experimental scenarios.

\textbf{Confidence Intervals for Main Metrics}: For each primary evaluation metric—spatial accuracy, area mean absolute error (MAE), temporal consistency, geometric fidelity, and drift accuracy—we computed 95\% confidence intervals using the bootstrap resampling method on prediction errors across all validation timesteps and scenarios.
    For example, during Scenario 1 (Initial Release), the LTC-RK4 model achieved a spatial accuracy of 0.96 [0.94, 0.98], while the LSTM baseline recorded 0.74 [0.68, 0.80], indicating non-overlapping confidence intervals and thus a statistically significant improvement.

\textbf{Hypothesis Testing}: We performed two-tailed paired t-tests to compare per-timestep errors between model predictions for all metrics of interest, including area MAE and centroid displacement:
     For area MAE, the mean error of the LTC-RK4 model was significantly lower (mean = 12.4km², SD = 5.1) than that of the LSTM model (mean = 38.7km², SD = 10.8), with t(24) = -8.47, p < 0.001.
    For centroid displacement error, the LTC-Explicit model outperformed the LSTM (mean = 0.74km, SD = 0.10 vs. mean = 1.98km, SD = 0.42), t(24) = -11.21, p < 0.001.

All statistical tests were checked for normality (Shapiro-Wilk test, p > 0.05 for all comparisons), and non-parametric Wilcoxon signed-rank tests yielded consistent conclusions where applicable.

\textbf{Hypothesis Testing:Temporal Consistency and Geometric Fidelity}: Temporal consistency scores (measured as variance of centroid velocities) for LTC models were significantly lower (indicating more stable predictions) than for LSTM: LTC-Explicit median = 0.032 [0.028, 0.037], LSTM median = 0.091 [0.084, 0.102], Mann-Whitney U = 27, p < 0.001.

Geometric fidelity (boundary overlap ratio) 95\% confidence intervals for LTC-Explicit were [0.90, 0.95], while LSTM achieved [0.63, 0.74], again showing a statistically robust difference.

\textbf{Reporting of Uncertainty}
For each metric, main text figures and tables report the mean ± standard error and the 95\% confidence interval. Performance radar plots were augmented to include error bars reflecting confidence bounds over all test runs, as can be seen in Table \ref{tab:ltc-vs-lstm-comparison}.

\begin{table}[ht]
\centering
\resizebox{1.0\textwidth}{!}{

\begin{tabular}{|l|c|c|c|c|}
\hline
\textbf{Metric}                    & \textbf{LTC-RK4 (mean [95\% CI])} & \textbf{LTC-Explicit (mean [95\% CI])} & \textbf{LSTM (mean [95\% CI])} & \textbf{p-value} \\
\hline
Spatial Accuracy                   & 0.96 [0.94, 0.98]       & 0.92 [0.89, 0.94]       & 0.74 [0.68, 0.80]       & $<$0.001 \\
Area MAE (km\textsuperscript{2})   & 12.4 [10.3, 15.8]        & 14.9 [12.2, 18.0]        & 38.7 [34.2, 43.1]        & $<$0.001 \\
Centroid Disp. Error (km)          & 0.56 [0.42, 0.68]        & 0.74 [0.61, 0.82]        & 1.98 [1.54, 2.30]        & $<$0.001 \\
Overlap Ratio (Fidelity)           & 0.93 [0.90, 0.95]        & 0.89 [0.85, 0.93]        & 0.67 [0.62, 0.74]        & $<$0.001 \\
\hline
\end{tabular}
}

\caption{Key comparative metrics for OilSpill LTC variants versus LSTM baseline, with 95\% confidence intervals and p-values from hypothesis testing.}
\label{tab:ltc-vs-lstm-comparison}
\end{table}

\subsection{Environmental Adaptability and System Robustness}

\subsubsection{Environmental Performance}
The OilSpill model demonstrated robust performance in various environmental scenarios, including different weather conditions, sea states, and current patterns. The OilSpill model's LTC networks adapt their time constants based on environmental conditions, proving particularly valuable for handling sudden changes in wind patterns or ocean currents during storm events.

The distributed sensing capability enables more comprehensive environmental data collection compared to fixed monitoring stations, contributing to more accurate initial conditions and better real-time adaptation to changing environmental parameters. This enhanced data collection validates the superior environmental adaptability of the OilSpill model compared to traditional approaches.

\subsubsection{System Limitations and Implementation Challenges}

Despite the demonstrated advantages, several challenges require consideration for practical deployment.

\begin{itemize}
    \item \textbf{Computational Requirements}: The OilSpill model's continuous-time LTC computations introduce computational overhead, particularly for large-scale deployments requiring substantial processing power for real-time emergency response applications.
    
    \item \textbf{Data Dependencies}: The effectiveness of the OilSpill model depends on quality training data, with oil spill scenarios being relatively rare events that limit comprehensive model training opportunities and potentially affecting performance in unprecedented scenarios.
    
    \item \textbf{Environmental Uncertainty}: Marine environments present high variability that can affect the accuracy of the prediction, particularly during extreme weather events or in areas with limited monitoring infrastructure.
    
    \item \textbf{System Integration}: Integration with existing oil spill response systems presents technical and organizational challenges requiring substantial investment in training, infrastructure, and operational procedures, along with addressing interoperability with existing monitoring networks and response protocols.
\end{itemize}

\subsection{Edge Cases and Generalization Limitations}
While the OilSpill framework demonstrates robust performance under a range of realistic conditions, certain edge cases and operational domains present unique challenges and open avenues for future enhancement:

\subsubsection{Uncommon or Irregular Oil Spill Geometries}
The predictive accuracy of the framework has been validated primarily on major events such as Deepwater Horizon, which exhibited both simple and complex multi-lobe geometries. However, extremely irregular, fragmented, or non-contiguous spill patterns—such as those arising from multiple small simultaneous leaks or interacting spills—may strain the current feature extraction and model’s generalization abilities. Handling highly non-standard shapes may require additional architectural innovations, such as integrating graph-based or topological feature analysis into the data pipeline and neural architecture.

\subsubsection{Scalability to Larger Autonomous Fleets}
Although the modular multi-agent architecture and MOOS-IvP middleware enable considerable scalability, deploying and coordinating very large autonomous fleets (dozens to hundreds of vessels or drones) in real-world incident scenarios may introduce new challenges. These include inter-agent communication bottlenecks, increased mission planning complexity, and diminishing returns due to congestion effects. Future research should explore hierarchical control layers, adaptive fleet partitioning, and dynamic agent role allocation to ensure efficiency and robust coordination at scale.

\subsubsection{Highly Turbulent or Chaotic Environmental Conditions}
The LTCN-based predictors are designed for adaptability, yet sudden, chaotic environmental shifts—such as abrupt storms, extreme currents, or rapidly changing wind fields—could cause model uncertainties to increase and degrade prediction reliability. Model retraining with diverse, high-frequency data from extreme events and deeper integration of real-time environmental sensors may improve resilience, but fundamental unpredictability in such conditions may pose an enduring limitation.

\subsubsection{Application Beyond Oil—General Marine Disaster Response}
While the framework was developed for oil spill scenarios, its core strengths—real-time spatial-temporal prediction and distributed response—make it potentially adaptable to other marine disasters (e.g., chemical leaks, harmful algal blooms, plastic waste dispersion, or search and rescue operations). However, generalization to such cases would require careful reengineering of input feature representations, model retraining for new target signatures, and possibly the integration of additional sensor modalities.

\subsubsection{Rarity and Diversity of Real-World Events}
Because large-scale marine spills and certain severe environmental conditions are rare, real-world data for training, validation, and stress-testing are limited. This scarcity inherently reduces the certainty that can be attached to out-of-distribution or truly novel events. Approaches such as synthetic data generation, transfer learning from related domains, and extended international data-sharing may help mitigate this gap but cannot guarantee complete coverage.

\section{Conclusion}

This work presents a novel, integrated framework—OilSpill—that combines Liquid Time-Constant Neural Networks (LTCNs) with multi-agent robotic systems to address the urgent challenge of real-time oil spill trajectory prediction and autonomous response coordination. By leveraging continuous-time neural modeling and swarm-based distributed decision-making, this system advances the state-of-the-art in both environmental monitoring and autonomous marine robotics.

The proposed architecture was validated using the Deepwater Horizon case study, demonstrating robust predictive performance across multiple test scenarios and solver variants. The LTC-RK4 model achieved high spatial accuracy (0.96), while the LTC-Explicit variant offered superior numerical stability with minimal computational overhead. Across all solvers, the OilSpill framework consistently outperformed traditional LSTM models in spatial fidelity, temporal smoothness, drift accuracy, and physical realism, underscoring the advantages of continuous-time formulations for environmental forecasting tasks.

Beyond predictive modeling, the integration with the MOOS-IvP platform enabled real-time coordination of an autonomous vehicle fleet, showcasing the feasibility of closed-loop prediction-to-action systems in complex marine environments. The system demonstrated adaptive swarm behavior, robust mission execution under environmental variability, and scalable coordination suitable for large-scale offshore incidents.

The OilSpill framework lays a foundational step toward autonomous, intelligent environmental protection systems. By uniting cutting-edge neural architectures with decentralized robotics, this work offers a viable path toward real-time, adaptive, and scalable solutions for maritime disaster mitigation—marking a significant advancement in the intersection of artificial intelligence and environmental systems engineering.

\subsection{Key Contributions}
The key contributions of this research are multifaceted. First, the study introduces a continuous-time neural architecture based on Liquid Time-Constant Networks (LTCNs), specifically adapted for modeling the spatiotemporal evolution of marine oil spills. This model captures both short- and long-term dynamics through multi-horizon forecasting, significantly outperforming traditional discrete-time approaches such as LSTM in terms of spatial accuracy and temporal consistency. Second, the paper provides a systematic comparison of three numerical solver implementations within the LTC framework—Runge-Kutta (RK4), Explicit Adaptive, and Euler—highlighting their respective trade-offs in accuracy, stability, and computational efficiency for real-time deployment. Third, the proposed OilSpill framework integrates seamlessly with the MOOS-IvP platform, enabling neural predictions to drive autonomous multi-agent coordination for spill containment in dynamic ocean environments. Finally, the system architecture is designed to be modular, scalable, and fault-tolerant, offering a robust bridge between advanced machine learning models and real-world autonomous marine robotics, with demonstrated applicability to complex disaster response scenarios.
\begin{itemize}
    \item \textbf{Novel OilSpill Model Framework}: Development of a comprehensive framework combining LTC networks' temporal modeling capabilities with multi-agent systems' distributed intelligence, representing a significant advancement in oil spill response technology that addresses fundamental limitations of existing approaches.
    
    \item \textbf{Superior Performance Validation}: Extensive experimental evaluation demonstrates the superior performance of the OilSpill model compared to traditional LSTM methods in multiple metrics, with important implications for the effectiveness of emergency response and environmental protection.
    
    \item \textbf{Solver Optimization Strategy}: The systematic comparison of the numerical solvers of the OilSpill model provides valuable deployment guidance, with OilSpill-RK4 optimal for accuracy, OilSpill-Explicit for stability, and OilSpill-Euler for computational efficiency.
    
    \item \textbf{Real-time Multi-Agent Coordination}: The OilSpill model's demonstrated real-time adaptation to changing environmental conditions through sophisticated multi-agent coordination represents significant improvement over static modeling approaches.
    
    \item \textbf{Scalable MOOS Integration}: The OilSpill model's MOOS-IvP integration provides inherent scalability accommodating incidents from small coastal spills to major offshore disasters across different operational contexts.
\end{itemize}

\subsection{Future Research Directions}

The success of the OilSpill model establishes several promising research avenues:

\subsubsection{Advanced System Integration}
\begin{itemize}
    \item \textbf{Neural Architecture Enhancement}: Exploring integration of Graph Neural Networks and Transformer models with the OilSpill model framework to enhance complex spatial relationships and long-range temporal dependencies.
    
    \item \textbf{Real-time Data Fusion}: Incorporating real-time satellite imagery and distributed environmental sensor networks to enhance the OilSpill model's data collection capabilities through continuous model updates based on live oceanographic conditions.
    
    \item \textbf{Global Scaling}: Developing OilSpill model variants that support larger autonomous vehicle fleets and broader environmental monitoring missions, enabling comprehensive coverage of vast oceanic areas.
\end{itemize}

\subsubsection{Adaptive and Multi-Hazard Systems}
\begin{itemize}
    \item \textbf{Climate Adaptation}: Developing adaptive variants of the OilSpill model that maintain performance under changing baseline conditions as climate change alters oceanographic patterns.
    
    \item \textbf{Multi-Hazard Integration}: Extending the OilSpill model framework to handle multiple concurrent marine hazards, including chemical spills, ship collisions, and natural disasters.
    
    \item \textbf{Autonomous Response Systems}: Developing fully autonomous response systems that deploy containment and recovery equipment based on OilSpill model trajectory predictions.
\end{itemize}

\subsubsection{Enhanced Processing Capabilities}
\begin{itemize}
    \item \textbf{Real-time Processing}: Developing sophisticated algorithms that handle massive data streams from integrated satellite and sensor networks while maintaining computational efficiency.
    
    \item \textbf{Global Deployment Studies}: Conducting large-scale deployment studies in different geographic regions to validate system performance under diverse environmental conditions and regulatory frameworks.
\end{itemize}

\subsection{Operational Impact and Technology Integration}

The successful integration of the MOOS model demonstrates practical viability for the deployment of advanced AI techniques in real-world environmental monitoring scenarios. The demonstrated superiority of the oil spill model variants provides clear operational guidance to organizations responsible for oil spill response, enabling informed deployment decisions based on specific operational requirements.

The scalability of the OilSpill model's multi-agent architecture ensures the system can grow with organizational needs and technological capabilities, providing a future-proof foundation for marine environmental protection efforts. The integration of real-time satellite monitoring, distributed environmental sensing, and scalable autonomous fleet coordination represents the next frontier in marine environmental protection.

\subsection{Environmental Protection Significance}

The OilSpill model represents a significant advance in oil spill response technology that addresses the environmental stakes of oil spill incidents through continued innovation in response technologies. This research demonstrates how the cutting-edge computational techniques of the OilSpill model can be applied effectively to protect marine ecosystems and coastal communities.

The MOOS-integrated framework provides a practical pathway for translating advanced neural network capabilities into operational environmental protection systems. The path forward requires continued collaboration between researchers, response organizations, and technology developers to translate these advances into operational capabilities that can make a real difference during environmental disasters.

%\vspace{0.5cm}
% DATA AVAILABILITY STATEMENT
\section*{Data Availability}
The datasets generated and analyzed during the current study, including the Deepwater Horizon shapefile data and model implementation code, are available from the corresponding author upon reasonable request. The NOAA/NESDIS satellite data used in this study are publicly available through the National Oceanic and Atmospheric Administration databases. Interactive demonstration materials are available online at: \\https://hadascohen.github.io/OilSpill-TrajectorSystems/

\vspace{0.5cm}




\end{document}